\documentclass[final,leqno,onefignum,onetabnum]{siamart190516}

\usepackage{amsmath,amssymb}
\usepackage{supertabular,multirow}
\usepackage{graphicx}
\usepackage{epsfig}
\usepackage{subfigure}
\usepackage{eufrak}
\usepackage{bm}
\usepackage{color}
\usepackage{algorithm}
\usepackage{algorithmic}
\usepackage{appendix}

\title{Variational inference formulation for a model-free simulation of a dynamical system with unknown parameters by a recurrent neural network}

\author{Kyongmin Yeo\thanks{IBM T.J. Watson Research Center, Yorktown Heights, NY, USA  (\email{kyeo@us.ibm.com}). }%Questions, comments, or corrections to this document may be directed to that email address.}
            \and Dylan E. C. Grullon\thanks{Department of Electrical Engineering and Computer Science, Massachusetts Institute of Technology, Cambridge, MA, USA}
            \and Fan-Keng Sun\footnotemark[2]
            \and Duane S. Boning\footnotemark[2]
            \and Jayant R. Kalagnanam\footnotemark[1]}

\begin{document}

\maketitle
	\slugger{sisc}{xxxx}{xx}{x}{x--x}%slugger should be set to mms, siap, sicomp, sicon, sidma, sima, simax, sinum, siopt, sisc, or sirev

\begin{abstract}
We propose a recurrent neural network for a ``model-free'' simulation of a dynamical system with unknown parameters without prior knowledge. 
The deep learning model aims to jointly learn the nonlinear time marching operator and the effects of the unknown parameters from a time series dataset. We assume that the time series data set consists of an ensemble of trajectories for a range of the parameters. The learning task is formulated as a statistical inference problem by considering the unknown parameters as random variables. A latent variable is introduced to model the effects of the unknown parameters, and a variational inference method is employed to simultaneously train probabilistic models for the time marching operator and an approximate posterior distribution for the latent variable. Unlike the classical variational inference, where a factorized distribution is used to approximate the posterior, we employ a feedforward neural network supplemented by an encoder recurrent neural network to develop a more flexible probabilistic model. The approximate posterior distribution makes an inference on a trajectory to identify the effects of the unknown parameters. The time marching operator is approximated by a recurrent neural network, which takes a latent state sampled from the approximate posterior distribution as one of the input variables, to compute the time evolution of the probability distribution conditioned on the latent variable. In the numerical experiments, it is shown that the proposed variational inference model makes a more accurate simulation compared to the standard recurrent neural networks. It is found that the proposed deep learning model is capable of correctly identifying the dimensions of the random parameters and learning a representation of complex time series data.
\end{abstract}
	
	

\pagestyle{myheadings}
\thispagestyle{plain}

\section{Introduction}

Dynamical systems have been widely used in the modeling of complex physical, biological, and engineering processes \cite{Furusawa12,Strogatz01,Tel05}. Most of the dynamical systems have a few parameters that have to be identified either by a theoretical analysis or experiments \cite{Ramsay07,Ventura06}. However, because of the high complexity of the physical (biological) systems, noise in the experimental data, and unresolved dynamics, it is very challenging to correctly identify those parameters, which introduces an uncertainty. Quantifying the effects of uncertainty in the simulation of a dynamical system has been extensively studied over the last two decades \cite{Frangos10,Najm09,Xiu02}. While most of the uncertainty quantification methods require almost complete knowledge on the system dynamics, such as the time marching operator and the source or distribution of the uncertainty, it is not uncommon, particularly in real-world applications, to encounter a very complex system, of which dynamics is not well understood.  

``Model-free'' approaches, which aim to identify the system dynamics from data, have been a long-standing research topic across many disciplines, \emph{e.g.}, statistics, physics, engineering, and so on. Classical statistical models based on autoregressive stochastic processes or state-space models \cite{Box08,harvey90} assume a linear transition of a state space to make an inference analytically tractable. Although nonlinear extensions of the classical methods have been widely used \cite{Arula02,Evensen03}, most of the nonlinear methods require at least partial knowledge on the system dynamics \cite{Hamilton15}. There has been a progress in the data-driven modeling of nonlinear dynamical systems based on a delay-coordinate embedding \cite{Hamilton16,Sugihara12}. A new class of methods, employing the dynamic mode decomposition supplemented with machine learning approaches, has been proposed to learn the governing equations directly from the data \cite{Brunton16,Tu14}. Yet, purely data-driven modeling of nonlinear dynamical systems remains as a challenging problem. 

Recently, there has been a surge of interest in adopting deep learning techniques in the modeling of physical systems \cite{Brunton20}. Due to its strength in learning nonlinear structures of the data, deep learning offers a powerful tool to build a computationally efficient and accurate surrogate model for the uncertainty quantification of physical systems \cite{Zhu18}. The physics informed neural network (PINN) proposed by \cite{Raissi19} uses the governing equations to impose a physical constraint in the training of an artificial neural network (ANN), which makes it possible to train an ANN with a sparse data. The new approach has inspired numerous follow-up studies, extending the framework to tackle the uncertainty quantification problems \cite{Geneva20,Sun20,Yang19}. Although the initial results of PINN look promising \cite{Raissi20}, it requires a set of governing equations to impose physical constraints. For data-driven identification of nonlinear dynamical systems, \cite{Lusch18} proposed a novel deep learning framework to learn a linear embedding of nonlinear dynamics based on the Koopman operator theory. \cite{Rudy19} has proposed a feedforward neural network to learn the nonlinear time marching operator and a measurement noise, which employs a Runge-Kutta time integration scheme. \cite{Yeo19}  has proposed a recurrent neural network model, which is capable of approximating the probability distribution of a stochastic process without any distributional assumptions. 

In a data-driven modeling, it is typically assumed that the data is acquired from one source, or under an identical condition. However, in practice, many of the dataset consist of an ensemble over a range of parameters. For examples, an experimental data consists of many sets of experiments with varying parameters, and a real-world observation data may be affected by changes in the environmental conditions. In general, complex nonlinear interactions make it challenging to separate the effects of each parameters on the dynamics without knowing a functional relation even for a noise-free data. Data-driven identification of the effects of parameters can be thought as a representation learning problem, which has been of great interest in the data mining community \cite{Bengio13}. In representation learning, we aim to find a low-dimensional representation from a complex data. In the context of the dynamical systems, if we can successfully find such a ``representation'' of the effects of the parameters from a dataset, the learned representation can be used to constrain the data-driven simulation to follow the correct dynamics. Built upon the recent advancement in a variational inference technique \cite{Kingma14}, a promising approach have been proposed \cite{Hsu17,Yingzhen18}, where two neural networks are jointly trained to ``disentangle'' a time-invariant factor from a time-dependent feature. 

In this study, we are interested in learning the time marching operator of a dynamical system from noisy observations, when the dataset consists of an ensemble of trajectories generated from a wide range of parameters. We formulate the learning task as a variational inference problem. It is shown that the proposed deep learning approach can identify the dimensionality of the parameters and make a more accurate simulation by exploiting the learned representation. This paper is organized as follows. A formal description of the problem setup is given in section \ref{sec:Prob}. Section \ref{sec:method} describes the variational inference techniques and a probabilistic model is proposed to learn the nonlinear dynamics with unknown parameters. In section \ref{sec:results}, the proposed model is tested against two nonlinear time series. Finally, the conclusions are given in section \ref{sec:summary}.
	
\section{Problem formulation}\label{sec:Prob}
Let $\bm{\phi}\in\mathbb{R}^d$ be the state of a physical system. The dynamics of $\bm{\phi}$ is governed by a system of differential equations;
\begin{equation} \label{eqn:general_dyn}
\frac{d \bm{\phi}}{dt} = \mathcal{F}(\bm{\phi},\bm{u};\bm{\alpha}),
\end{equation}
in which $\bm{u}\in\mathbb{R}^{N_u}$ is an exogenous forcing and $\bm{\alpha}\in\mathbb{R}^{N_{\alpha}}$ denotes the parameters of the governing equations. The exogenous forcing may come from an ambient condition, such humidity, temperature, and pressure, or a driving force, for example, electro-magnetic field or pressure gradient. We do not assume a prior knowledge on the time marching operator, $\mathcal{F}(\bm{\cdot};\bm{\alpha})$. In general, the ground truth, $\bm{\phi}(t)$, is not accessible. We only observe a corrupted measurement, $\bm{y}_t$, at a discrete time interval, i.e.,
\begin{equation}
\bm{y}_t = \bm{\phi}_t + \bm{\epsilon}_t,
\end{equation}
where $\bm{\phi}_t = \bm{\phi}(t \delta t)$, $\delta t$ is a sampling interval, and $\bm{\epsilon}_t$ is a stochastic noise process. The time series data set consists of $K$ trajectories of the noisy observations and exogenous forcing; $\mathcal{D} = \{ ( \bm{Y}^{k}_{0:T},\bm{U}^{k}_{0:T});~k=1,\cdots,K\}$, where $\bm{Y}^{k}_{0:T}=(\bm{y}^k_0,\cdots,\bm{y}^k_T)$ and $\bm{U}^{k}_{0:T}=(\bm{u}^k_0,\cdots,\bm{u}^k_T)$. We assume that $\mathcal{D}$ is generated by gathering data from the systems under a range of environmental conditions.

Here, we aim to build a data-driven model from an ensemble of the systems in (\ref{eqn:general_dyn}), where the physical parameter, $\bm{\alpha}$, changes from one trajectory to another. To account for the variability of $\bm{\alpha}$, we consider $\bm{\alpha}$ as a random variable with a probability distribution, $p(\bm{\alpha})$. A trajectory in $\mathcal{D}$ can be thought as a realization from $p(\bm{\alpha})$, i.e.,
\begin{equation}
\frac{d \bm{\phi}^k}{dt} = \mathcal{F}(\bm{\phi}^k,\bm{u}^k;\bm{\alpha}^k),~~\bm{\alpha}^k \sim p(\bm{\alpha}),~~\text{for}~k = 1,\cdots,K,
\end{equation}
and $\bm{y}^k_t = \bm{\phi}^k_t+\bm{\epsilon}^k_t$. 
Note that both $\bm{u}$ and $\bm{\alpha}$ varies from one trajectory to another, while the random parameter, $\bm{\alpha}$, does not change over time in a trajectory. Define a \emph{complete} data set as
\[
\mathcal{D}^* = \{ ( \bm{Y}^{k}_{0:T},\bm{U}^{k}_{0:T},\bm{\alpha}^k);~k=1,\cdots,K\}.
\]
The underlying data generating distribution of $\mathcal{D}^*$ can be written as
\begin{equation} \label{eqn:data_distribution}
\mathcal{D}^* \sim p(\bm{Y}_{0:T},\bm{\alpha}|\bm{U}_{0:T}) = p(\bm{Y}_{0:T}|\bm{\alpha},\bm{U}_{0:T})p(\bm{\alpha}).
\end{equation}
The first term, $p(\bm{Y}_{0:T}|\bm{\alpha},\bm{U}_{0:T})$, represents the temporal evolution of the dynamical system under aleatoric uncertainties, \emph{e.g.}, the noise process. The exogenous forcing, $\bm{U}_{0:T}$, is assumed to be known and, thus, is not a target of the inference. When $\bm{U}_{0:T}$ itself is a stochastic process of interest, it can be appended to $\bm{Y}_{0:T}$ as a part of the target variable. 

Now, the goal of the data-driven model is to approximate the data generating distribution (\ref{eqn:data_distribution}). However, it is assumed that we do not have an access to  $\mathcal{D}^*$. Only the incomplete data set, $\mathcal{D}$, where $\bm{\alpha}$ is omitted, is available. Without a prior knowledge on $\bm{\alpha}$ or $\mathcal{F}(\bm{\cdot};\bm{\alpha})$, it is challenging to directly model (\ref{eqn:data_distribution}). To circumvent the difficulties, we introduce a latent variable, $\bm{z} \in \mathbb{R}^{N_z}$, and aim to learn a probabilistic model,
\begin{equation} \label{eqn:approx_data_dist}
p(\bm{Y}_{0:T},\bm{z}|\bm{U}_{0:T}) = p(\bm{Y}_{0:T}|\bm{U}_{0:T},\bm{z})p(\bm{z}),
\end{equation}
from $\mathcal{D}$. The probabilistic model (\ref{eqn:approx_data_dist}) is defined to have the same structure with the data generating distribution (\ref{eqn:data_distribution}). Using the product rule, (\ref{eqn:approx_data_dist}) can be written as
\begin{equation} \label{eqn:approx_data_dist-1}
p(\bm{Y}_{0:T}|\bm{U}_{0:T},\bm{z})p(\bm{z}) = p(\bm{y}_0) p(\bm{z}) \prod_{t=0}^{T-1}p(\bm{y}_{t+1}|\bm{Y}_{0:t},\bm{U}_{0:t},\bm{z}). 
\end{equation}
Here, we use the fact that $\bm{y}_t$ is independent from the future exogenous forcing, $\bm{U}_{t+1:T}$ as well as $\bm{u}_t$, i.e.,
\[
p(\bm{y}_t|\bm{Y}_{0:t-1},\bm{U}_{0:T}) = p(\bm{y}_t|\bm{Y}_{0:t-1},\bm{U}_{0:t-1}).
\]
Since $\bm{u}_t$ denotes the exogenous forcing applied at the time of the observation of $\bm{y}_t$, $\bm{u}_t$ only affects future observations, $\bm{y}_{t+i}$ for $i \in \mathbb{N}^+$. In the probabilistic model (\ref{eqn:approx_data_dist-1}), it is not of central interest to find the marginal distributions, $p(\bm{z})$ and $p(\bm{y}_0)$. We are interested in approximating the time marching operator for $\bm{y}$, $p(\bm{y}_{t+1}|\bm{Y}_{0:t},\bm{U}_{0:t},\bm{z})$, given a proper conditioning variable $\bm{z}$.
Hence, there are two major tasks. The first task is to train an inference model for $\bm{z}$ that provides a good summarization on the characteristics of a trajectory,  $(\bm{Y}_{0:T},\bm{U}_{0:T})\in\mathcal{D}$, \emph{i.e.}, $p(\bm{z}|\bm{Y}_{0:T},\bm{U}_{0:T})$. Then, a dynamics model is trained with respect to the identified latent variable, \emph{i.e.}, $p(\bm{Y}_{0:T}|\bm{U}_{0:T},\bm{z})$, in which $\bm{z} \sim p(\bm{z}|\bm{Y}_{0:T},\bm{U}_{0:T})$. 

\section{Methodology}\label{sec:method}
In this section, we describe a variational inference formulation to jointly train the inference and dynamics models in section \ref{sec:VI}, a recurrent neural network to evaluate a likelihood function in section \ref{sec:rnn}, and a regularized optimization formulation for the variational inference problem in section \ref{sec:reg_opt}. The architecture of the artificial neural networks and the data-driven simulation method are described in section \ref{sec:ann} and section \ref{sec:simulation}, respectively.

\subsection{Variational Inference} \label{sec:VI}
Variational inference (VI) has been widely used in the Machine Learning community due to its strength in learning an inference model of a latent variable \cite{Blei17}. The original formulation of VI imposes some restrictions on the probabilistic model to make the inference problem analytically tractable \cite{Bishop06}. Recently, \cite{Kingma14} proposed a VI formulation based on deep learning approaches. The new VI method allows more flexible probabilisitic models by using artificial neural networks. In this study, we employ the VI formulation by \cite{Kingma14}. In the variational inference, we aim to find an approximate posterior distribution of the latent variable, $q(\bm{z}|\bm{Y}_{0:T},\bm{U}_{0:T})$, because the true posterior distribution, $p(\bm{z}|\bm{Y}_{0:T},\bm{U}_{0:T})$, is intractable. Hereafter, we omit the obvious dependence on the exogenous forcing, $\bm{U}_{0:T}$, in the notation for simplicity. 

A probabilistic model is usually trained by maximizing a likelihood function. However, we cannot directly maximize the complete likelihood function, $p(\bm{Y}_{0:T},\bm{z}|\bm{U}_{0:T})$, because the latent variable, $\bm{z}$ is not observed. Hence, we consider the marginal log likelihood function of $\mathcal{D}$,
\begin{equation} \label{eqn:likelihood}
\sum_{i=1}^K \log p(\bm{Y}^i_{0:T}) = \sum_{i=1}^K D_{KL}(q(\bm{z}|\bm{Y}^i_{0:T})||p(\bm{z}|\bm{Y}^i_{0:T})) + \text{ELBO}.
\end{equation}
Here, $D_{KL}(q || p)$ denotes the Kullback-Leibler divergence,
\begin{equation}
D_{KL}(q(\bm{z})||p(\bm{z})) = \int q(\bm{z}) \log \frac{q(\bm{z})}{p(\bm{z})} d\bm{z}.
\end{equation}
The evidence lower bound (ELBO) is
\begin{equation} \label{eqn:elbo}
\text{ELBO} = - \sum_{i=1}^K \left\{ D_{KL}(q(\bm{z}|\bm{Y}^i_{0:T})||p(\bm{z})) - E_{\bm{z} \sim q(\bm{z}|\bm{Y}^i_{0:T})} \left[ \log p(\bm{Y}^i_{0:T}|\bm{z}) \right] \right\},
\end{equation}
in which $p(\bm{z})$ is a prior distribution. It is straightforward to prove (\ref{eqn:likelihood}) by expanding the terms on the right-hand side of (\ref{eqn:likelihood}) and applying Bayes' theorem,
\[
p(\bm{Y}_{0:T}) = \frac{p(\bm{Y}_{0:T}|\bm{z})p(\bm{z})}{p(\bm{z}|\bm{Y}_{0:T})}.
\] 
The Kullback-Leibler divergence defines a distance between two probability distributions and is always non-negative. Hence, ELBO on the right-hand side of (\ref{eqn:likelihood}) provides a ``lower bound'' of the marginal log likelihood. Note that it is challenging to find a model that maximizes the marginal likelihood function (\ref{eqn:likelihood}), because the true posterior distribution, $p(\bm{z}|\bm{Y}_{0:T})$, is not tractable. In a VI formulation, we aim to maximize ELBO, instead of the likelihood function itself. It is obvious from (\ref{eqn:likelihood}) that maximizing ELBO corresponds to minimizing the difference between the true and the approximate posterior distributions \cite{Bishop06}.

\subsection{Recurrent Neural Network} \label{sec:rnn}
The second term of ELBO (\ref{eqn:elbo}) represents the temporal evolution of the random variable $\bm{y}$, as shown in (\ref{eqn:approx_data_dist-1}). Here, we use a recurrent neural network to model the time evolution of $\bm{y}$. A recurrent neural network is a nonlinear state-space model to tackle a sequential inference problem \cite{GoodfellowBengio16}. A typical recurrent neural network consists of two steps,
\begin{align}
\bm{h}_{t+1} &= \bm{\Psi}_h(\bm{h}_t,\bm{y}_t,\bm{z}), \label{eqn:RNN} \\
p(\bm{y}_{t+1}|\bm{h}_{t+1}) &= \bm{\Psi}_y(\bm{h}_{t+1}),
\end{align}
in which $\bm{h}_t$ is a hidden state, and $\bm{\Psi}_h$ and $\bm{\Psi}_y$ are the nonlinear functions of the RNN. It is clearly shown that $\bm{y}_{t+1}$ becomes conditionally independent from the past, $\bm{Y}_{0:t}$, given the hidden state, $\bm{h}_{t+1}$. From the conditional independence, the marginal likelihood function of a trajectory becomes
\begin{equation} 
p(\bm{Y}_{0:T}|\bm{z}) = \idotsint \left[ \prod_{t=1}^T p(\bm{y}_t|\bm{h}_t)p(\bm{h}_t|\bm{h}_{t-1},\bm{y}_{t-1},\bm{z}) \right] p(\bm{y}_0)p(\bm{h}_0) d\bm{h}_0 \cdots d\bm{h}_T. \label{eqn:Y-1}
 \end{equation}
Due to the deterministic nature, the transition probability of RNN is given by a Dirac delta function,
\begin{equation}
p(\bm{h}_{t+1}|\bm{h}_t,\bm{y}_t,\bm{z})  = \delta(\bm{h}_{t+1} - \bm{\Psi}_h(\bm{h}_t,\bm{y}_t,\bm{z})).
\end{equation}
If we further assume the prior distribution, $p(\bm{h}_0) = \delta(\bm{h}_0)$, the marginal likelihood function (\ref{eqn:Y-1}) becomes
\begin{equation}
p(\bm{Y}_{0:T}|\bm{z}) = \left[ \prod_{t=1}^T p(\bm{y}_t|\bm{h}_{t})\right] p(\bm{y}_0),
\end{equation}
and, hence, the log likelihood function is simply
\begin{equation} \label{eqn:rnn_loglikeli}
\log p(\bm{Y}_{0:T}|\bm{z}) = \sum_{t=1}^T \log p(\bm{y}_t|\bm{h}_t) + \log p(\bm{y}_0).
\end{equation}
Note that $\bm{h}_t$ is a function of the entire trajectory up to $t-1$, i.e., $\bm{Y}_{0:t-1}$, as well as the latent variable, $\bm{z}$. Here, we use a Gaussian distribution with a diagonal covariance for a probabilistic model for RNN,
\begin{equation}
 p(\bm{y}_t|\bm{h}_t) = \mathcal{N}(\bm{y}_t;\bm{\mu}_t,diag(\bm{\sigma}^2_t)).
\end{equation}
Here, $\bm{\mu}_t \in \mathbb{R}^d$ and $\bm{\sigma}_t \in \mathbb{R}^d$ are, respectively, the mean and standard deviation of the Gaussian distribution, and $diag(\bm{\cdot})$ denotes a diagonal matrix. The recurrent neural network, $\bm{\Psi} = (\bm{\Psi}_y\circ\bm{\Psi}_h)$, takes ($\bm{y}_{t-1},\bm{h}_{t-1},\bm{z}$) as an input, and the output is a  $\mathbb{R}^{d\times 2}$ matrix, which provides $\bm{\mu}_t$ and $\bm{\sigma}_t$;
\[
(\bm{\mu}_t,\log \bm{\sigma}_t) = \bm{\Psi}(\bm{y}_{t-1},\bm{h}_{t-1},\bm{z}).
\]
Note that the artificial neural network computes $\log \sigma_t$, instead of $\sigma_t$, which makes it easier to satisfy the positivity constraint, $\sigma_t > 0$.
Then, the log likelihood function (\ref{eqn:rnn_loglikeli}) becomes
\begin{equation} \label{eqn:rnn_mse}
\log p(\bm{Y}_{1:T}|\bm{z}) = -\sum_{t=1}^T \sum_{i=1}^d \left\{ \frac{1}{2} \left(\frac{{y_t}_i - {\mu_t}_i}{{\sigma_t}_i}\right)^2 + \log {\sigma_t}_i \right\} + C.
\end{equation}
Here, the constant terms are lumped together in $C$. For example, it is typical to use a non-informative prior for $\bm{y}_0$, \emph{e.g.} $p(\bm{y}_0) = \l^{-d}$, where $l$ is the length of an interval. Then, $\log p(\bm{y}_0) = -d \log l$ is just a constant, which does not contribute to the optimization. Thus, it is absorbed into $C$. Note that, if the probabilistic model for RNN is given as a Gaussian with a constant diagonal covariance, \emph{i.e.}, $\bm{\sigma}_t = \sigma$, and $\sigma$ is not estimated, (\ref{eqn:rnn_mse}) reduces to the standard mean-square loss function, which is typically used in a standard RNN. Since (\ref{eqn:rnn_mse}) provides a measure about how well the time series data is reconstructed by RNN, it is called a ``reconstruction error''.

\subsection{Regularized Optimization Formulation}\label{sec:reg_opt}
Following the convention of a standard optimization problem, we aim to minimize negative ELBO, instead of maximizing ELBO. Then, the loss function of the variational inference problem is
\begin{equation}
\mathcal{L} = \sum_{i=1}^K  D_{KL}(q(\bm{z}|\bm{Y}^i_{0:T})||p(\bm{z})) - E_{\bm{z} \sim q(\bm{z}|\bm{Y}^i_{0:T})} \left[ \sum_{t=1}^T \log p(\bm{y}^i_t|\bm{h}_{t}) \right].
\end{equation}
An isotropic Gaussian distribution is used as the prior distribution, 
\[
p(\bm{z}) = \mathcal{N}(\bm{z};\bm{0},\sigma_z^2\bm{I}).
\]
Similar to the generative RNN model, we use a Gaussian distribution with a diagonal covariance for the posterior distribution; 
\[
q(\bm{z}|\bm{Y}_{0:T}) = \mathcal{N}(\bm{z};\bm{m}_q,diag(\bm{\sigma_q}^2)),
\]
where $\bm{m}_q \in \mathbb{R}^{N_z}$ and $\bm{\sigma}_q \in \mathbb{R}^{N_z}$ denote the mean and standard deviation, respectively. We use a feed-forward artificial neural network, $\bm{\eta}$, to approximate the posterior distribution, such that 
\begin{equation}
(\bm{m}_q,\log \bm{\sigma}_q) = \bm{\eta}(\bm{Y}_{0:T}).
\end{equation}
Again, the output of $\bm{\eta}$ is a  $\mathbb{R}^{N_z\times 2}$ matrix, which provides $\bm{m}_q$ and $\bm{\sigma}_q$. Then, the Kullback-Leibler divergence in ELBO can be computed analytically,
\begin{equation}
D_{KL}(q(\bm{z}|\bm{Y}_{0:T})||p(\bm{z})) = \sum_{i=1}^{N_z} \left\{  \frac{1}{2} \frac{\sigma^2_{q_i}+m_{q_i}^2}{\sigma^2_z}  - \log\left( \frac{\sigma_{q_i}}{\sigma_z} \right) \right\} - \frac{N_z}{2}.
\end{equation}
After \cite{chung15}, it has become popular to use an artificial neural network for the prior distribution, $p(\bm{z})$, which is jointly trained with $q(\bm{z}|\bm{Y}_{0:T})$. However, as shown in Appendix \ref{sec:prior_network}, training the prior and posterior distributions jointly leads to an ill-posed problem and, thus, should be avoided.
Finally, the loss function is 
\begin{align}
  \mathcal{L}^k_q &= \sum_{i=1}^{N_z} \left\{  \frac{1}{2} \frac{\sigma^2_{q_i}+m_{q_i}^2}{\sigma^2_z}  - \log\left( \frac{\sigma_{q_i}}{\sigma_z} \right) \right\} \Big|_{\bm{Y}^k_{0:T}} \\
    \mathcal{L}^k_y &= E_{\bm{z} \sim q(\bm{z}|\bm{Y}^k_{0:T})} \left[ \sum_{t=1}^T \sum_{i=1}^{d} \frac{1}{2}\frac{(y^k_{t_i} - \mu_{t_i})^2}{\sigma^2_{t_i}} + \log \sigma_{t_i}\right], \\
  \mathcal{L} =& \sum_{k=1}^K (\mathcal{L}^k_q + \mathcal{L}^k_y ) + C.\label{eqn:VI_loss}
\end{align}
Again, all the constant terms are lumped together in $C$. 

The first term in the loss function (\ref{eqn:VI_loss}) plays a role of a regularization to keep $q(\bm{z}|\bm{Y}_{0:T})$ around $p(\bm{z})$, while the second term tries to move $q(\bm{z}|\bm{Y}_{0:T})$ away from $p(\bm{z})$ in the direction of minimizing the reconstruction. The loss function has a similar structure with a standard regularized optimization problem, but without an explicit regularization coefficient. Note, however, that $\mathcal{L}_y$ depends on the length of the time series, $T$. If we limit our interest to a stationary, or ergodic, system, the distribution of the reconstruction error per one time step is also stationary, 
\[
\langle \log p(\bm{y}_t|\bm{h}_t) \rangle = \cdots = \langle \log p(\bm{y}_{t+T}|\bm{h}_{t+T}) \rangle,
\]
where $\langle \cdot \rangle$ denotes an ensemble average, e.g., average over different trajectories and time intervals. Then, the loss function can be written as
\[
\mathcal{L}_y = -E_{\bm{z} \sim q(\bm{z}|\bm{Y}_{0:T})} \left[ \log p(\bm{Y}_{0:T}|\bm{z}) \right]  \simeq -T \cdot E_{\bm{z} \sim q(\bm{z}|\bm{Y}_{0:T})} \left[ \langle \log p(\bm{y}_t|\bm{h}_t) \rangle \right].
\]
Now, it is obvious that $\mathcal{L}_y$ increases linearly with $T$, while $\mathcal{L}_q$ is fixed. So, for a larger $T$, the relative contribution of $\mathcal{L}_q$ to $\mathcal{L}$ becomes smaller, which makes the effects of the regularization (Kullback-Leibler divergence) weaker. Hence, the length of a time series, $T$, implicitly plays a role of a regularization coefficient, when $T$ is a hyperparameter that one has to choose when training a RNN.

Let $T_{ref}$ be the length of a time series for an optimal regularization, and $T$ be the length of the time series chosen for training of a RNN. Because $T_{ref}$ is typically not known, one can train multiple models with different $T$'s and choose the best performing model. However, there are some restrictions in choosing $T$. For a RNN to correctly learn the dynamics, $T$ should be larger than the largest characteristic timescale of the process. And, at the same time, $T$ should be chosen small enough to make the training computationally tractable. Note that the computational time to train a RNN is proportional to $T$. Instead of tuning $T$, we can rescale the loss function, considering the stationarity of the reconstruction error, such that
\begin{equation}
\mathcal{L} = \sum_{k=1}^K \left( \mathcal{L}^k_q + \frac{T_{ref}}{T}\mathcal{L}^k_y \right) + C =  \frac{T_{ref}}{T} \sum_{k=1}^K \left( \lambda \mathcal{L}^k_q +\mathcal{L}^k_y \right) + C,
\end{equation}
in which $\lambda = T/T_{ref}$. Since the scaling factor and the constant term are irrelevant in optimization, we can define a new loss function as
\begin{equation} \label{eqn:lambda_loss}
\mathcal{L} = \sum_{k=1}^K  \lambda \mathcal{L}^k_q + \mathcal{L}^k_y.
\end{equation}
Note that (\ref{eqn:lambda_loss}) has a similar structure with $\beta$-VAE (variational auto-encoder) proposed by \cite{Higgins17}. While they derived $\beta$-VAE from a constrained optimization formulation, here we show that the same formulation naturally arises from a variational inference method for a time series problem.

In most of the current deep learning frameworks, such as TensorFlow and PyTorch, once the network and loss functions are defined, the computational graph is automatically constructed and the gradients of the loss function with respect to the network parameters can be automatically computed, which makes it easy to train a deep learning model without understanding the mathematical details. Nevertheless, we provide a more detailed explanation about the model training in Appendix \ref{sec:training}.

\subsection{Artificial Neural Network} \label{sec:ann}

The recurrent neural network used in this study consists of three layers. The first layer transforms an input variable,
\begin{equation}\label{eqn:rnn-1}
\bm{\Psi}_{x} = (\mathcal{R}\circ\bm{L}^{N_c}_x)(\bm{x}_t).
\end{equation}
Here, $\bm{x}_t \in \mathbb{R}^{N_x}$ is the input variable, $N_c$ is the number of neurons in the RNN, and $\bm{L}^{N_c}_x$ denotes a linear transformation of the input variable $\bm{x}_t$. The linear transformation operator is defined as
\[
\bm{L}^a_i(\bm{x}) = \bm{W}\bm{x} + \bm{b},
\]
in which the superscript $a$ is the dimension of the output vector, the subscript $i$ denotes an identifier, and $\bm{W}\in\mathbb{R}^{a\times N_x}$ and $\bm{b}\in\mathbb{R}^{a}$ are the weight matrix and bias vector, respectively. And, $\mathcal{R}(\cdot)$ is a linear rectifier,
\[
\mathcal{R}(x) = \max(0,x).
\]
In the variational inference problem, the input variable is $\bm{x}^T_t = (\bm{y}^T_t,\bm{z}^T)^T$ and, for a standard RNN, $\bm{x}_t = \bm{y}_t$.
A two-level Gated Recurrent Unit (GRU) is used to compute the time evolution of the state variables \cite{Cho14},
\begin{align}
\bm{h}^{(1)}_{t+1} &= \text{GRU}_1(\bm{h}^{(1)}_{t},\bm{\Psi}_{x} ), \label{eqn:rnn-2}\\
\bm{h}^{(2)}_{t+1} &= \text{GRU}_2(\bm{h}^{(2)}_{t},\bm{h}^{(1)}_{t+1}). \label{eqn:rnn-3}
\end{align}
Here, $\bm{h}^{(1)}_t$ and $\bm{h}^{(2)}_t$ denote the internal states of $\text{GRU}_1$ and $\text{GRU}_2$, respectively.
Finally, the last layer computes the mapping between the state vector and the generative distribution,
\begin{align}
&\bm{g} = (\mathcal{R}\circ\bm{L}^{N_c}_g)(\bm{h}^{(2)}_{t+1})\\
&\bm{\mu}_{t+1} = \bm{L}^d_{\mu}(\bm{g}),~~\log(\bm{\sigma}_{t+1}) = \bm{L}^d_{\sigma}(\bm{g}).
\end{align}
Here, we used the two-level GRU, because it shows a better performance than a single-layer GRU for some prediction problems. However, it can be replaced by any RNN architectures.

The approximate posterior distribution, $q(\bm{z}|\bm{Y}_{0:T})$, has an explicit dependence on the length of the sequence used in the training, \emph{i.e.}, $\bm{\eta}(\bm{Y}_{0:T})$. Note, however, that the purpose of $q(\bm{z}|\bm{Y}_{0:T})$ is to identify the latent state that can explain the dynamics observed in the conditioning sequence, $\bm{Y}_{0:T}$. Hence, it is desirable to have a flexible model, of which inference result becomes invariant as long as the length of the conditioning sequence is longer than the characteristic timescale of the process. Here, we propose to use a RNN to encode the conditioning variable $\bm{Y}_{0:T}$ to the approximate posterior distribution, $q(\bm{z}|\bm{Y}_{0:T})$. Let $\bm{\Psi}_h^{enc}$ be the \emph{encoder} RNN, \emph{e.g.}, the first two layers of the RNN defined above, (\ref{eqn:rnn-1} -- \ref{eqn:rnn-3}). From the state-space model description of RNN, it is clear that \cite{GoodfellowBengio16},
 \begin{align}
 \bm{h}^{enc}_{t+1} &= \bm{\Psi}^{enc}_h(\bm{y}_t,\bm{\Psi}^{enc}_h(\bm{y}_{t-1},\bm{\Psi}^{enc}_h(\cdots\bm{\Psi}^{enc}_h(\bm{y}_1,\bm{\Psi}^{enc}_h(\bm{y}_0,\bm{h}_0))\cdots) \nonumber \\
 &=f(\bm{y}_t,\cdots,\bm{y}_0,\bm{h}_0),
 \end{align}
where $\bm{h}^{enc}_t = (\bm{h}^{(1)}_t,\bm{h}^{(2)}_t)$.
It shows that the hidden state, $\bm{h}^{enc}_{t+1}$, provides a representation of $\bm{Y}_{0:t}$. Because the hidden states of a RNN approximates a relaxation process \cite{Yeo19}, the dependence on $\bm{h}_0$ will vanish when $t$ is larger than a relaxation timescale. $\bm{\Psi}_h^{enc}$ may be trained as a standard RNN by using the entire data set, $\mathcal{D}$. The use of the encoder RNN makes it possible to compute the approximate posterior distribution, $q(\bm{z}|\bm{Y}_{0:t})$ for any sequence length $t$, instead of being fixed to the training sequence length, $T$. The effects of the sequence length, $t$, in the inference is discussed in figure \ref{fig:MG_time_latent}.

Now, define an artificial neural network for the approximate posterior as
\begin{equation}
\bm{\eta} = (\widehat{\bm{\eta}}\circ{\bm{\Psi}_{h,t}^{enc}})(\bm{Y}_{0:t}).
\end{equation}
Here, $\bm{\Psi}^{enc}_{h,t}$ indicates computing the encoder RNN over an input sequence, $\bm{Y}_{0:t}$, to compute $\bm{h}^{enc}_{t+1}$. The artificial neural network, $\widehat{\bm{\eta}}$, consists of the following operations,
\begin{align}
&\bm{v}_i = (\mathcal{R}\circ\bm{L}^{N_{v_i}}_{v_i})(\bm{v}_{i-1}),~\text{for}~i = 1,\cdots,N_\eta, \\
&\bm{m}_q = \bm{L}^{N_z}_m(\bm{v}_{N_\eta}),~\log(\bm{\sigma}_q) = \bm{L}^{N_z}_{\sigma_q}(\bm{v}_{N_\eta}),
\end{align}
in which $\bm{v}_0 = \bm{h}^{enc}_{t+1}$, $N_{\eta}$ is the number of the layers, and $N_{v_i}$ is the number of neurons in each layer. In other words, $\widehat{\bm{\eta}}$ is a multi-layer feedforward network.

In summary, the proposed variational inference model requires training two recurrent neural networks and one feedforward neural network;
\begin{align}
(\bm{\mu}^{enc}_t,\bm{\sigma}^{enc}_t) &= (\bm{\Psi}^{enc}_y \circ \bm{\Psi}^{enc}_h)(\bm{y}_{t-1},\bm{h}^{enc}_{t-1}) ~\text{for}~t=1,\cdots,T, \label{eqn:vi-rnn-1} \\
(\bm{m}_q,\bm{\sigma}_q) &= (\widehat{\bm{\eta}}\circ\bm{\Psi}^{enc}_{h,T})(\bm{Y}_{0:T}), \label{eqn:vi-rnn-2} \\
(\bm{\mu}_t,\bm{\sigma}_t) &= (\bm{\Psi}^{VI}_y \circ \bm{\Psi}^{VI}_h)(\bm{y}_{t-1},\bm{h}_{t-1},\bm{z})  ~\text{for}~t=1,\cdots,T. \label{eqn:vi-rnn-3}
\end{align}
Here, the outputs of the artificial neural networks define the following probability distributions,
\begin{align}
p(\bm{y}_t|\bm{Y}_{0:t-1}) &= \mathcal{N}(\bm{y}_t;\bm{\mu}^{enc}_t,diag({\bm{\sigma}^{enc}_t}^2)), \nonumber \\
q(\bm{z}|\bm{Y}_{0:T}) &= \mathcal{N}(\bm{z};\bm{m}_q,diag(\bm{\sigma}^2_q)), \nonumber \\
p(\bm{y}_t|\bm{Y}_{0:t-1},\bm{z}) &= \mathcal{N}(\bm{y}_t;\bm{\mu}_t,diag({\bm{\sigma}^2_t})). \nonumber
\end{align}
The encoder RNN (\ref{eqn:vi-rnn-1}) is trained independently by using the standard back-propagation through time method. In the encoder RNN, $\bm{\Psi}^{enc}_y$ is necessary only for the training. Once the encoder is trained, it is no longer used. After $\bm{\Psi}^{enc}_h$ is trained as a stand-alone standard RNN, the artificial neural networks of the VI model (\ref{eqn:vi-rnn-2}--\ref{eqn:vi-rnn-3}) are jointly trained as described in Appendix \ref{sec:training}. Following a convention in the deep learning community, hereafter, we use \emph{decoder} RNN to denote the dynamics model, $\bm{\Psi}^{VI}  = (\bm{\Psi}^{VI}_y \circ \bm{\Psi}^{VI}_h)$. A sketch of VI-RNN is also shown in Appendix \ref{sec:training} (figure \ref{fig:sketch}).

\subsection{Simulation Method}\label{sec:simulation}

Once the inference and dynamics models are trained, a stochastic simulation can be performed to compute the time evolution of the probability density function \cite{Yeo19}. In the simulations, we assume that there is a sufficiently long time series data before the simulation start time, $t=0$, of which length is denoted by $\tau$. For a one-step-ahead prediction, \emph{i.e.}, computing the probability distribution at $T+1$ given the data up to $T$, we need to compute
\begin{equation}
p(\bm{y}_{T+1}|\bm{Y}_{-\tau:T},\bm{U}_{-\tau:T}) = \int p(\bm{y}_{T+1}|\bm{Y}_{-\tau:T},\bm{U}_{-\tau:T},\bm{z})  q(\bm{z}|\bm{Y}_{-\tau:0},\bm{U}_{-\tau:0}) d\bm{z}.
\end{equation}
Note that the history data, $t \in [-\tau,0]$, is used to spin-up the simulation. The spin-up period is required not only to identify the effects of the unknown parameters by $q(\bm{z})$, but also to remove the spurious effects from the misspecification of the initial condition of a RNN. Due to the lack of a physical meaning of the hidden state of a RNN, $\bm{h}_t$, it is typical to start a simulation from an arbitrary value, usually zero, for $\bm{h}_0$. Similar to the analysis in \cite{Yeo19}, the update rule of $\bm{h}_t$ of a GRU is essentially a relaxation process, where the effects of the initial condition vanishes after a relaxation timescale \cite{Hart20,Yeo19b}. Instead of the ``cold start'' approach, it is possible to infer the correct $\bm{h}_0$ from the time series data for a ``hot start''. However, it requires to solve a high-dimensional nonlinear inverse problem and, thus, it is not computationally practical. The effects of the ``cold start'' are discussed in figure \ref{fig:MG_time_latent}.

The integration over the approximate posterior distribution is computed by using a Monte Carlo method. We first compute, $q(\bm{z}|\bm{Y}_{-\tau:0},\bm{U}_{-\tau,0})$, and draw $M$ samples;
\[
\bm{z}^{(m)} \sim q(\bm{z}|\bm{Y}_{-\tau:0},\bm{U}_{-\tau,0})~~\text{for}~m = 1,\cdots,M.
\]
Then, $M$ identical recurrent neural networks ($\bm{\Psi}^{VI}$) are run through the dataset with the input sequence, $x^{(m)}_{t-1} = (y_{t-1},\bm{z}^{(m)})$ for $t= -\tau,\cdots,T$ and $m=1,\cdots,M$ to evaluate $p(y_{T+1}|\bm{Y}_{-\tau:T},\bm{U}_{-\tau:T},\bm{z}^{(m)})$. In practice, it can be done by setting $(x_t^{(1)},\cdots,x_t^{(M)})$ in the batch dimension of the input to $\bm{\Psi}^{VI}$. Finally, the predictive distribution is approximated by a mixture distribution,
\begin{align}
\mu_{T+1} &= \frac{1}{M}\sum_{m=1}^M \mu^{(m)}_{T+1}, \\
\sigma^2_{T+1} &= \frac{1}{M} \sum_{m=1}^M {\mu_{T+1}^{(m)}}^2 + {\sigma_{T+1}^{(m)}}^2 - \mu_{T+1}^2, \\
p(y_{T+1}|\bm{Y}_{-\tau:T},\bm{U}_{-\tau:T}) &= \mathcal{N}(\mu_{T+1},\sigma^2_{T+1}),
\end{align}
in which $\mu^{(m)}_{T+1}$ and $\sigma^{(m)}_{T+1}$ denote the outputs of $\bm{\Psi}^{VI}$ for $x_T^{(m)}$. 

A multiple-step prediction corresponds to computing the time evolution of the probability density function due to the nonlinear marching operator. Unlike the one-step-ahead prediction, a multiple-step prediction requires to marginalize over $\bm{z}$ as well as all the intermediate time steps, \emph{i.e.};
\begin{align}\label{eqn:forecast}
&p(\bm{y}_{T+1}|\bm{Y}_{-\tau:0},\bm{U}_{-\tau:T}) = \\ &E_{\bm{z}\sim q(\bm{z}|\bm{Y}_{-\tau:0},\bm{U}_{-\tau:0})} \left[ \idotsint  p(\bm{y}_1|\bm{Y}_{-\tau:0},,\bm{U}_{-\tau:0},\bm{z})  \prod_{t=1}^{T}  p(\bm{y}_{t+1}|\bm{Y}_{-\tau:t},\bm{U}_{-\tau:t},\bm{z}) d\bm{y}_{t} \right]. \nonumber  
\end{align}
Here, $T \ge 1$ denotes a prediction horizon. Note that $\bm{U}$ is assumed to be given for the simulation. In Appendix \ref{sec:MC_algorithm}, a Monte Carlo simulation procedure is outlined. 

\section{Numerical experiments} \label{sec:results}

In the numerical experiments, the dimension of the internal state of the RNN is fixed at $N_c = 128$ both for the encoder RNN ($\bm{\Psi}^{enc}$) and the decoder RNN ($\bm{\Psi}^{VI}$). Hence, the only difference between $\bm{\Psi}^{enc}$ and $\bm{\Psi}^{VI}$ is the number of input variables in (\ref{eqn:rnn-1}). For $\bm{\Psi}^{enc}$, $\bm{x}_t = (\bm{y}_t,\bm{u}_t)$, while the input variable to $\bm{\Psi}^{VI}$ is $\bm{x}_t = (\bm{y}_t,\bm{u}_t,\bm{z})$. The posterior network ($\widehat{\bm{\eta}}$) is a three-layer feedforward network and the number of neurons in each layer is equal to the dimension of $\bm{h}^{enc}$, i.e., $N_\eta = 3$ and $N_{v_i} = 2\times N_c = 256$ for $i=1, 2,3$. The dimension of the latent variable is $N_z = 10$.

Both the encoder RNN and the variational inference model ($\bm{\Psi}^{VI} + \widehat{\bm{\eta}}$) are trained by using a minibatch stochastic gradient descent method, called ADAM \cite{Kingma15}. The initial learning rate is set to $\xi_{max} = 10^{-3}$ and decreased by the following cosine function \cite{Loshchilov17},
\begin{equation}
\xi_l = \xi_{min} + \frac{\xi_{max}-\xi_{min}}{2}\left\{  1+\cos \left( \frac{l}{L}\pi \right)\right\},
\end{equation}
in which $l$ is the iteration count. The minimum learning rate and the maximum number of iterations are set to $\xi_{min} = 10^{-4}$ and $L = 3\times10^4$, respectively. The momentum coefficients of ADAM are set to the default values recommended in \cite{Kingma15}. The size of the minibatch is 20. At each SGD iteration, 20 training data are randomly selected from $\mathcal{D}$ by first randomly sampling from the $K$ trajectories \emph{with replacement} and then randomly selecting the starting point of those trajectories. The length of the training sequence is set to $200$ time steps. The size of the Monte Carlo samples to evaluate the loss function $\mathcal{L}_y$ is $25$. The encoder RNN is trained as a standard regression RNN, which does not require a Monte Carlo sampling. The optimization parameters of the encoder RNN $\bm{\Psi}^{enc}$ are the same with VI-RNN.

The variational inference model requires a sequential training of two RNNs, first the encoder RNN, $\bm{\Psi}^{enc}$, and then the decoder RNN, $\bm{\Psi}^{VI}$. It takes about 4.1 hours to train $\bm{\Psi}^{enc}$ and additional 6.2 hours to train $\bm{\Psi}^{VI}$ on a single NVIDIA TESLA K80 GPU, which makes the total training time of the proposed variational inference model around 10.3 hours. So, training the variational inference model takes about 2.5 times more wall-clock time than a standard RNN. Instead of the sequential training, it is possible to train all of the artificial neural networks directly from the formulation (\ref{eqn:vi-rnn-2}--\ref{eqn:vi-rnn-3}), which is a usual practice in the deep learning community. However, in our experiments, training all the networks together makes the model performance worse. One possible reason of the suboptimal performance is because, while the standard method of training an RNN requires to compute the objective function, \emph{e.g.}, mean-square loss, and back-propagates the error at every time step, in the training of the encoder RNN, $\bm{\Psi}^{enc}$, of the variational inference formulation, the error back-propagates only after $\bm{\Psi}^{enc}$ rolls over the entire sequence, $\bm{Y}_{0:T}$. Hence, it becomes very difficult to correctly train $\bm{\Psi}^{enc}$.

In all of the experiments, the dataset consists of 500 trajectories, each with a length of 1,000 time steps, \emph{i.e.}, $K = 500$ and $T=1,000$. The variables are normalized by the respective maximum and minimum values, \emph{e.g.},
\[
y^*_i = \frac{y_i - \min(y_i)}{\max(y_i)-\min(y_i)}-0.5,~\text{for}~i = 1,\cdots,d.
\]
The dataset ($\mathcal{D}$) is separated into two disjoint datasets, one for the training and the other for the validation. The training dataset ($\mathcal{D}_T$) consists of the first 400 trajectories in $\mathcal{D}$ and the last 100 trajectories are in the validation dataset ($\mathcal{D}_V$).

\subsection{Mackey-Glass Time Series} \label{subsec:MG}
We first consider the Mackey-Glass time series with random parameters. The Mackey-Glass equation is a nonlinear time-delay dynamical system \cite{Mackey77},
\begin{equation} \label{eqn:MG}
\frac{d \phi(t)}{dt} = \frac{\alpha \phi(t-\tau)}{1+\phi^{10}(t-\tau)} - \gamma \phi(t).
\end{equation}
The dynamics of the Mackey-Glass system switches between periodic and chaotic, depending on the set of parameters. The Mackey-Glass equation has been extensively used as a benchmark system to investigate time-delay dynamics \cite{Farmer82,Sprott,Erneux17}. 

\begin{figure}
  \centering
  \includegraphics[width=0.48\textwidth]{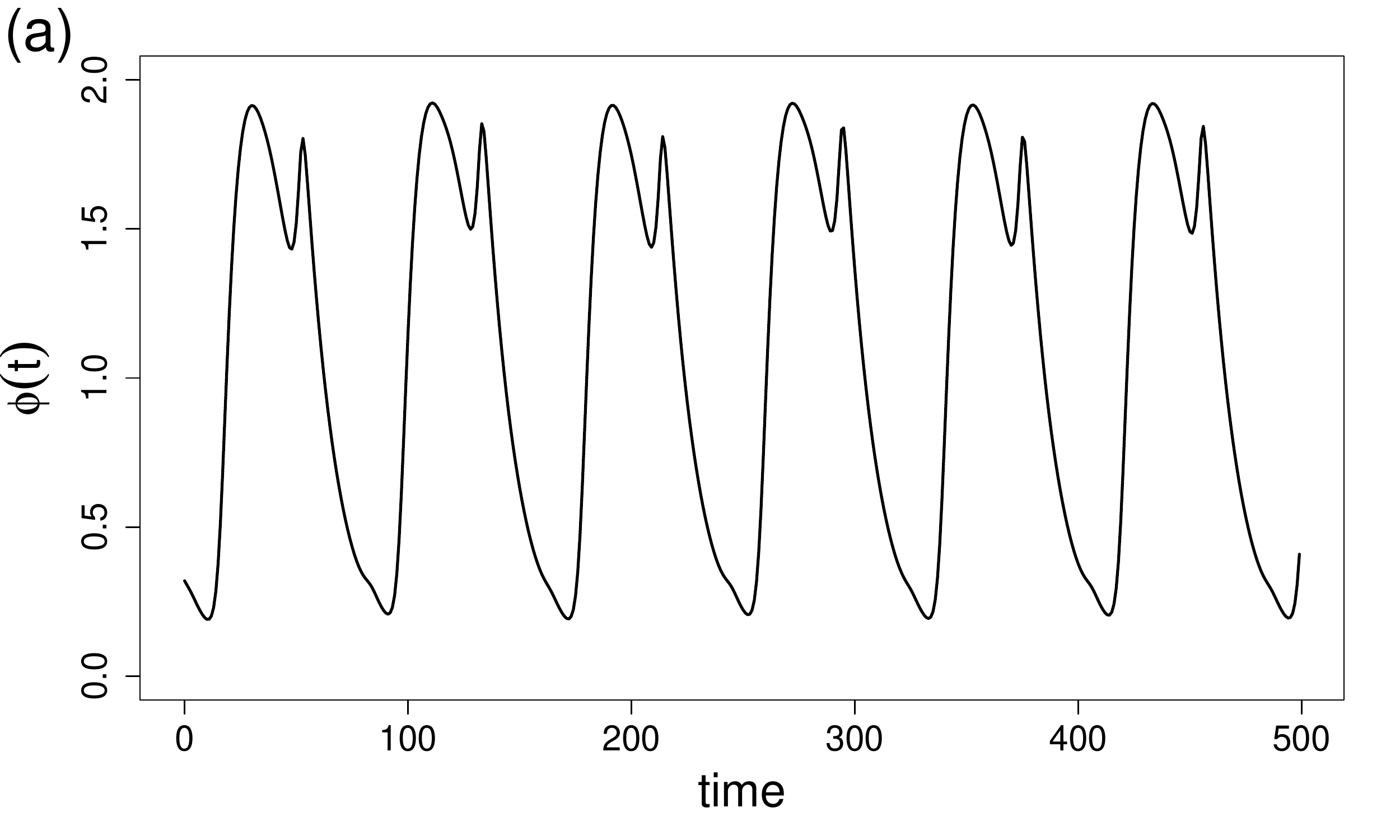}
  \includegraphics[width=0.48\textwidth]{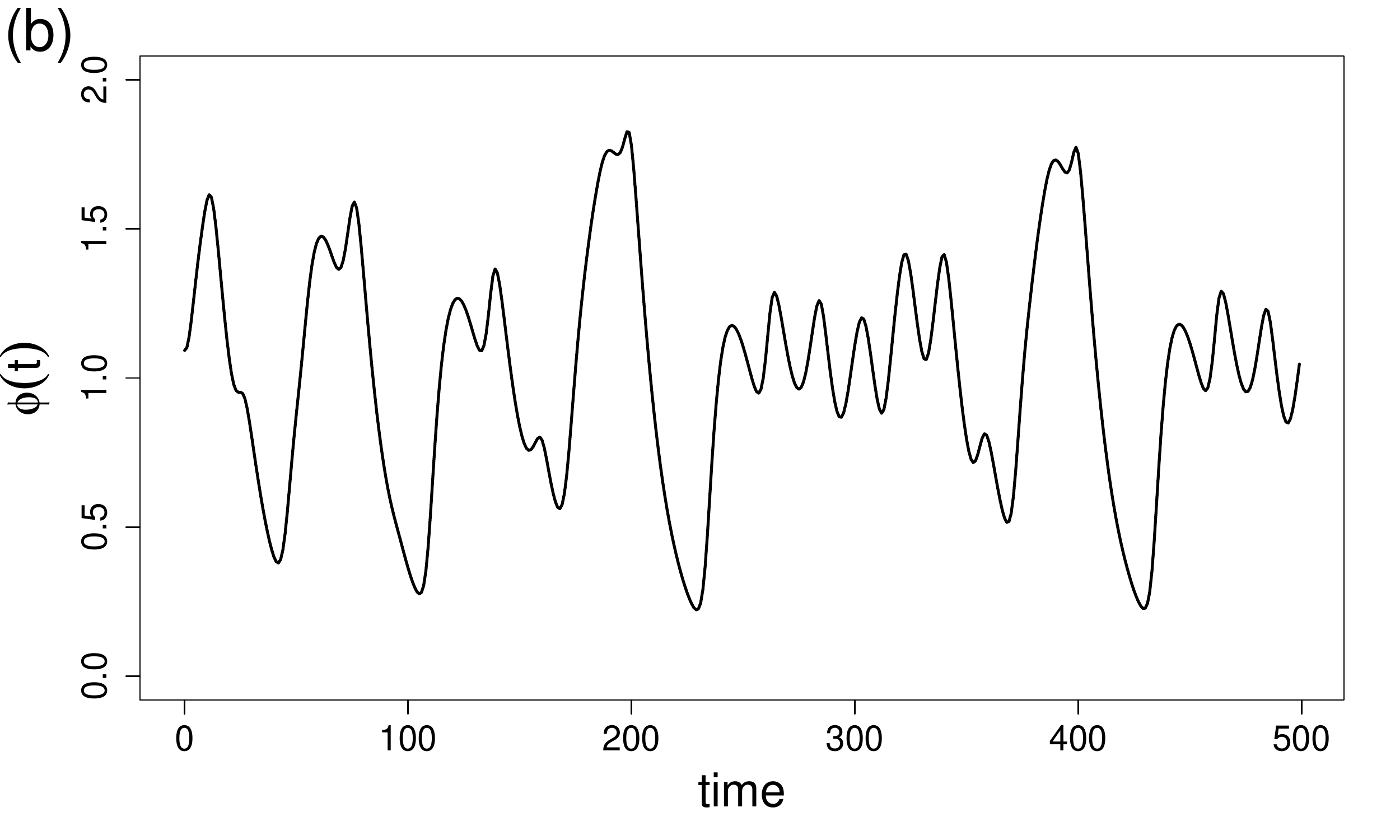}\\
  \includegraphics[width=0.48\textwidth]{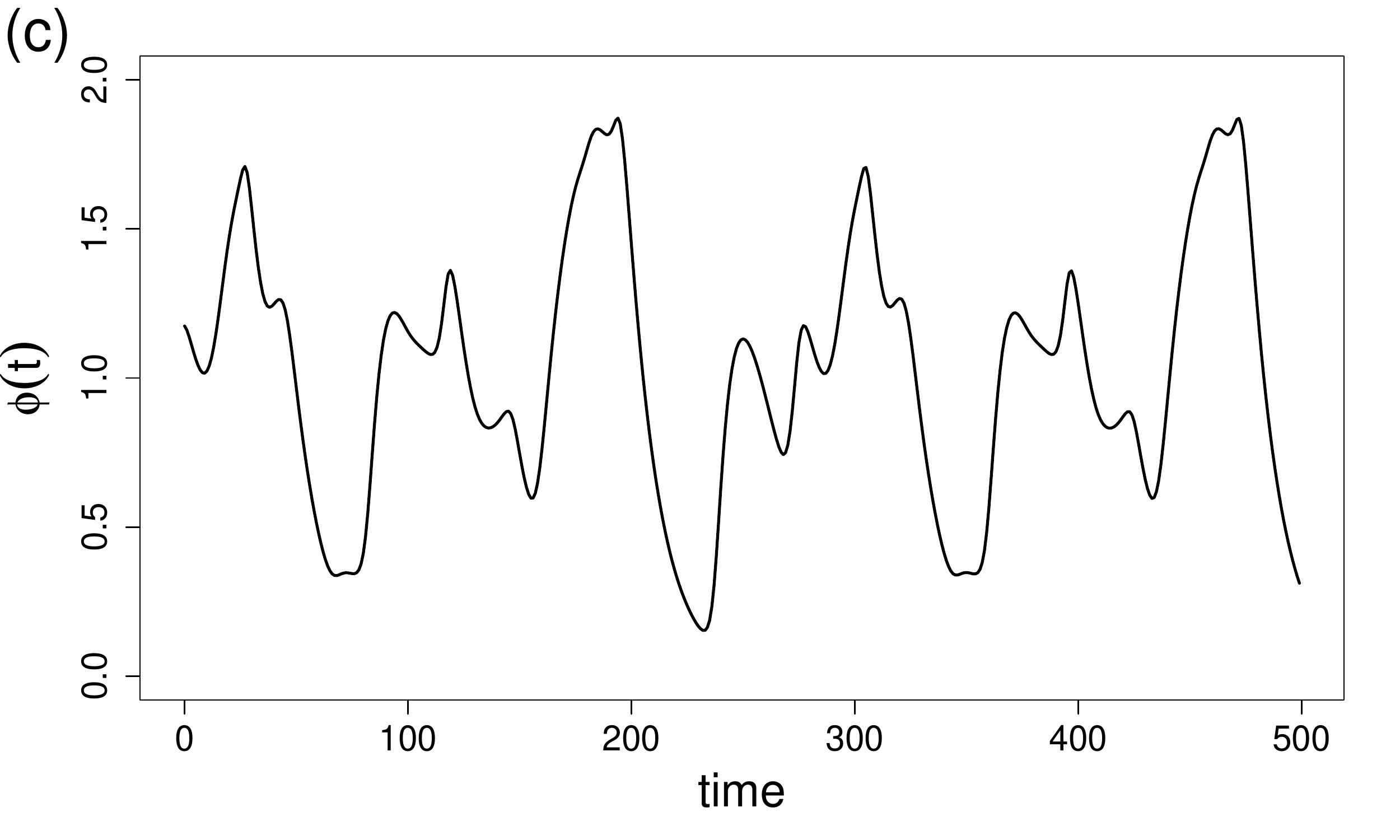}
  \includegraphics[width=0.48\textwidth]{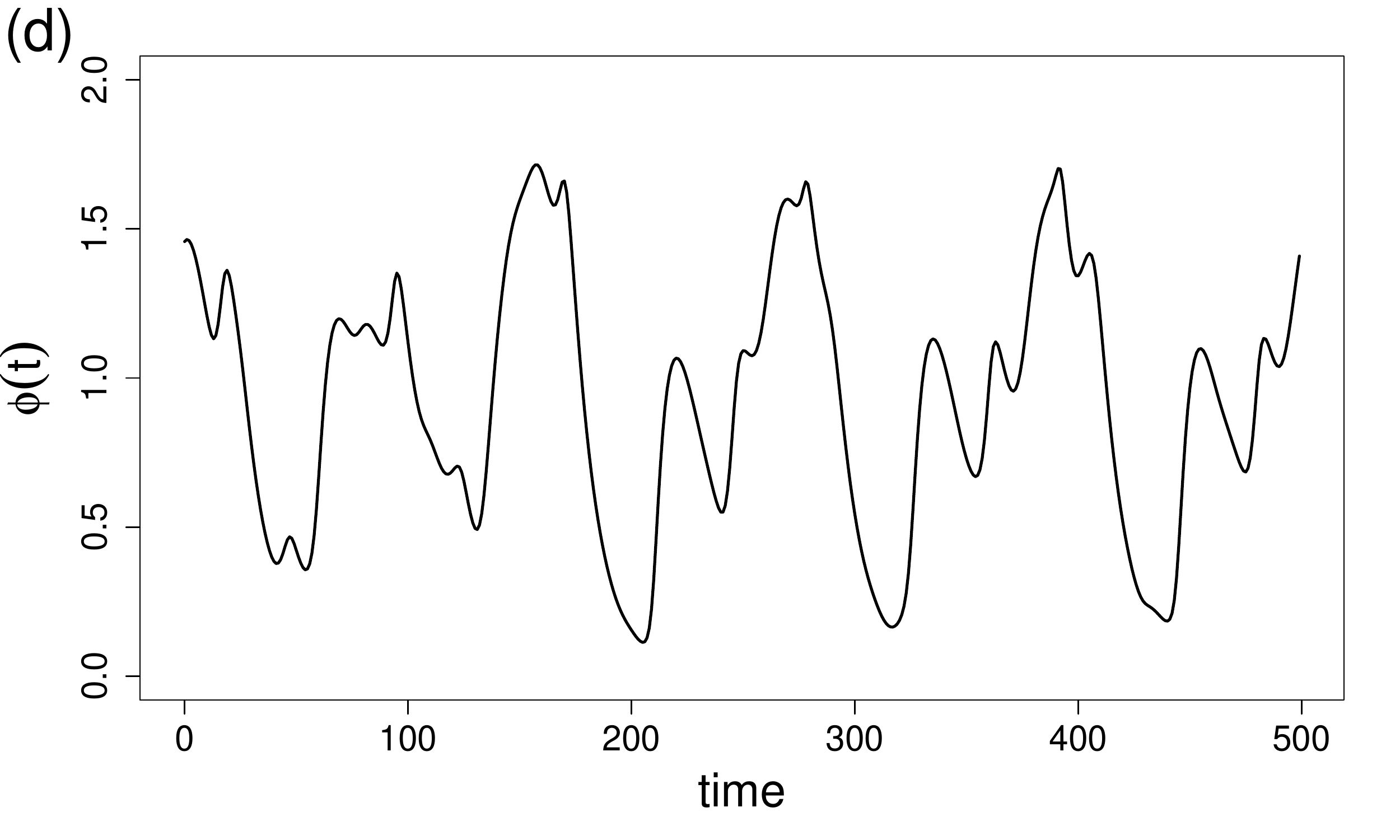}
  \caption{Sample ground-truth trajectories of the Mackey-Glass time series for different sets of the parameters, $(\alpha,\gamma,\tau)$; (a) (0.350,0.070,33.72), (b) (0.242,0.080,24.44), (c) (0.222,0.076,31.79), (d) (0.257,0.094,31.85).} \label{fig:MG_data}
\end{figure}

To create the dataset, $\mathcal{D}=\{(y^k_0,\cdots,y^k_T); k = 1,\cdots,500, T = 1000\}$, first, ground-truth trajectories are generated by randomly sampling the three parameters in (\ref{eqn:MG}) from uniform distributions,
\begin{equation} \label{eqn:MG_param}
\alpha \sim \mathcal{U}(0.2,0.4), ~
\gamma \sim \mathcal{U}(0.05,0.1), ~
\tau \sim \mathcal{U}(20,40).
\end{equation}
The parameters are chosen around an onset of a chaos regime \cite{Farmer82}. Figure \ref{fig:MG_data} shows four sample trajectories in the data set. The Mackey-Glass equation is integrated by using a third-order Adams-Bashforth method with the time-step size of 0.01. Then, the ground-truth trajectories are downsampled to make time series data with a sampling interval of $\delta t = 1$. Finally, the time series data is perturbed by an uncorrelated random noise, 
\begin{equation} \label{eqn:noisy_obs}
y^k_t = \phi^k(t\delta t) + \epsilon_t^k,~\text{for}~k=1,\cdots,500~\text{and}~t=1,\cdots,1000.
\end{equation}
Here, the noise process is $\epsilon^k_t \sim \mathcal{N}(0,\sigma^2_\epsilon)$. The standard deviation of $\epsilon$ is $\sigma_\epsilon = 0.03$, which is about 5\% of the standard deviation of $\mathcal{D}$.

We first compare the performances of the proposed variational-inference problem (VI-RNN) and the standard RNN for a one-step-ahead prediction task, \emph{i.e.}, sequentially making a prediction of $(\mu_t,\sigma_t)$ by using $(y_0,\cdots,y_{t-1})$. Hereafter, we use RNN to refer to the results from the standard RNN model, unless stated otherwise. The one-step-ahead prediction of VI-RNN is computed with $(M=200)$ samples as outlined in section \ref{sec:simulation}.

\begin{table}
\center{
\caption{Normalized root mean-square errors of the mean ($e_\mu$) and the standard deviation ($e_\sigma$), and normalized log likelihood (NLL) of RNN and VI-RNN.} \label{tbl:MG_one_step_err}
\begin{tabular}{c|ccccc}
\hline \hline
& RNN & \multicolumn{4}{c}{VI-RNN}\\
$\lambda$ & - & 0.01 & 0.1 & 1.0 & 10.0\\
\hline
$e_\mu$ & 0.030 & 0.028 & 0.028 & 0.026 & 0.031 \\
$e_\sigma$ & 0.139 & 0.095 & 0.103 & 0.087 & 0.123 \\
NLL & 0.975 & 0.976 & 0.977 & 0.981 & 0.972\\
\hline \hline
\end{tabular}
}
\end{table}

The model performance is evaluated against the 100 trajectories in $\mathcal{D}_V$. The length of the testing trajectory is $T=600$. After the simulations are performed for $t=1,\cdots,600$, the first 200 time steps are discarded and the error metrics are computed for $t=201,\cdots,600$ to remove the data used for the conditioning of the latent variable, $q(\bm{z}|\bm{Y}_{0:200})$, from the evaluation. Three error metrics are considered. First, the normalized root mean-square error of the mean is defined with respect to the ground-truth;
\begin{equation}
e_\mu = \left [\frac{1}{|\mathcal{D}_V|}\sum_{k=1}^{|\mathcal{D}_V|} \overline{\frac{\left(\mu^k - \phi^k \right)^2}{ Var(\phi^k)}} \, \right]^{1/2},
\end{equation} 
in which the overline denotes a time average over the length of the evaluation period, $Var(\phi^k)$ is the variance of the $k$-th ground-truth times series, and $|\mathcal{D}_V|$ is the size of the validation dataset. Similarly, the normalized root mean-square error of the standard deviation is defined as
\begin{equation}
e_\sigma = \left [\frac{1}{|\mathcal{D}_V|}\sum_{k=1}^{|\mathcal{D}_V|} \overline{\frac{Var(\mu^k)}{ Var(\phi^k)}} \, \right]^{1/2} - 1.
\end{equation} 
Because the ground-truth distribution is a Gaussian distribution, $e_\mu$ and $e_\sigma$ are sufficient to assess the accuracy of the inference. Additionally, we compute a normalized log likelihood with respect to the data. The log likelihood (LL) is computed without the constant term,
\begin{equation}
\text{LL} = \frac{1}{|\mathcal{D}_V|}\sum_{k=1}^{|\mathcal{D}_V|}\overline{ -\frac{1}{2}\frac{(\mu^k - y^k)^2}{\sigma^k} - \log(\sigma^k)}.
\end{equation}
Then, LL is normalized by that of a perfect model,
\begin{equation}
\text{NLL} = \frac{LL}{-0.5 - \log(\sigma_\epsilon)}.
\end{equation}
The normalized log likelihood (NLL) provides a relative performance of the model with respect to a perfect inference model and is always less than one.

The error metrics are shown in table \ref{tbl:MG_one_step_err}. VI-RNNs are trained with four different penalty parameters, $\lambda = 0.01,0.1,1,10$. It shows that, in general, VI-RNN makes better inferences compared to RNN. But, the improvement is only marginal. For VI-RNN, the model performance becomes better as $\lambda$ is increased from 0.01 to 1. When $\lambda$ becomes too large, the error of VI-RNN starts to increase. It is shown that at $\lambda = 10$ the errors of VI-RNN become larger than those of RNN. This is a typical behavior of a penalized maximum log likelihood, or a regularized optimization, method, in which the model performance starts to degrade when the regularization term becomes too large. For the one-step-ahead prediction task, although the error metrics of VI-RNN are smaller than RNN for a proper range of $\lambda$, the difference is within a range of statistical noise. This is due to the nature of a sequential inference problem; simply, when $y_t$ is provided as a input, the prediction of $y_{t+1}$ will not deviate too much. 

We now consider a multiple-step forecast of the Mackey-Glass time series. A multiple-step forecast corresponds to computing the time evolution of the probability distribution of $y$ without an observation, which requires an evaluation of the high-dimensional integral in (\ref{eqn:forecast}). Unless the time evolution is given by a linear Gaussian model, a multiple-step forecast generally requires a Monte Carlo simulation (see Appendix  \ref{sec:MC_algorithm}) \cite{Yeo19}. The wall-clock time of the Monte Carlo simulation is shown in Table \ref{tbl:MC_time} as a function of the sample size, $N_s$. The computation is performed on a single NVIDIA TESLA K80 GPU. Note that, for a multiple-step forecast of the probability density function, RNN also requires a Monte Carlo simulation, and the computational cost of the Monte Carlo simulation of RNN is virtually the same with that of VI-RNN.  

\begin{table}
\center{
\caption{Wall-clock time of the Monte Carlo simulation of VI-RNN per time step with respect to the sample size, $N_s$.} \label{tbl:MC_time}
\begin{tabular}{cccccc}
\hline \hline
$N_s$ & 100 & 500 & 1000 & 2000 & 4000\\
\hline
second & $4.6\times 10^{-3}$ & $2.0\times10^{-2}$ & $4.0\times10^{-2}$ & $8.1\times10^{-2}$ & $1.6\times10^{-1}$\\ 
\hline \hline
\end{tabular}
}
\end{table}

\begin{figure}
  \centering
  \includegraphics[width=0.48\textwidth]{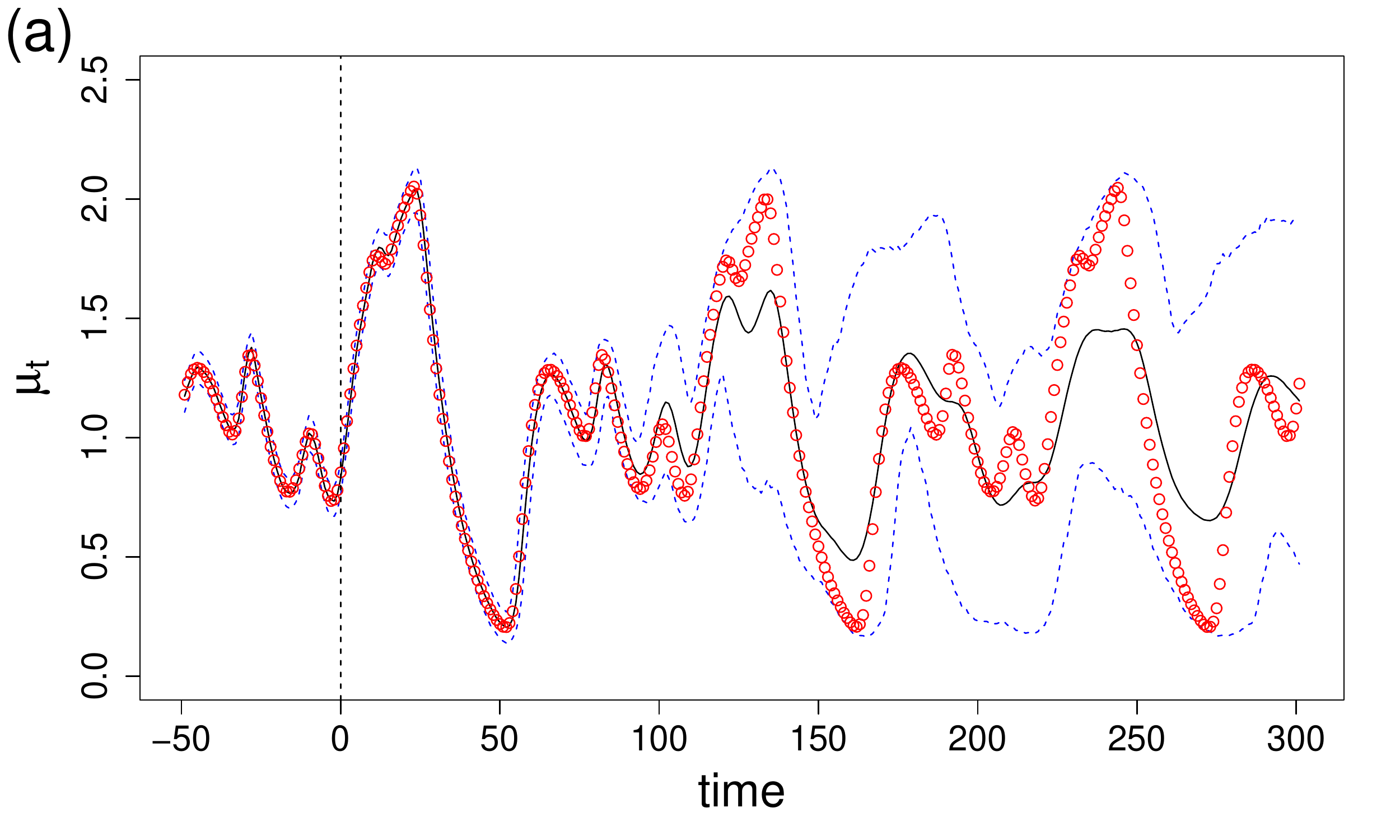}
  \includegraphics[width=0.48\textwidth]{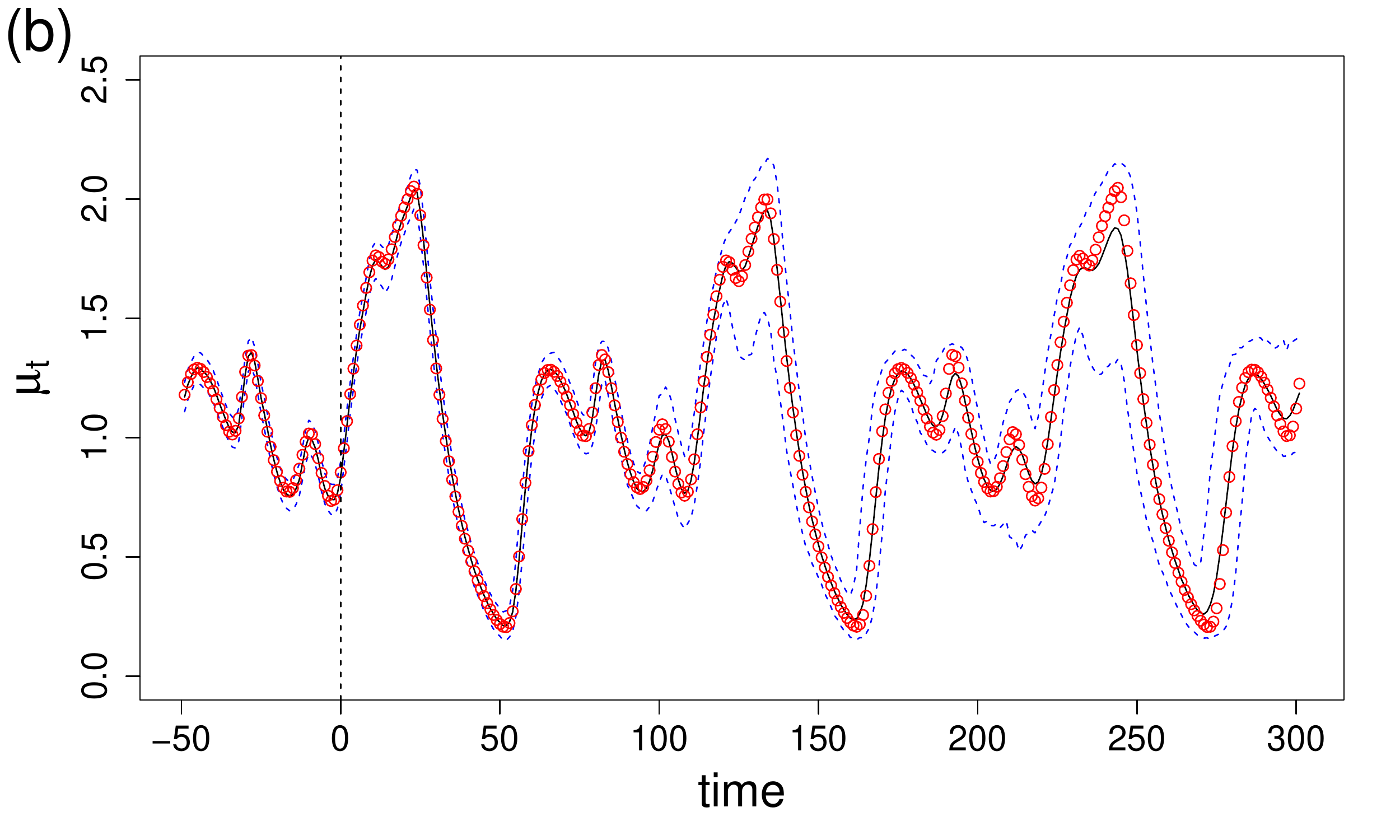}\\
  \includegraphics[width=0.48\textwidth]{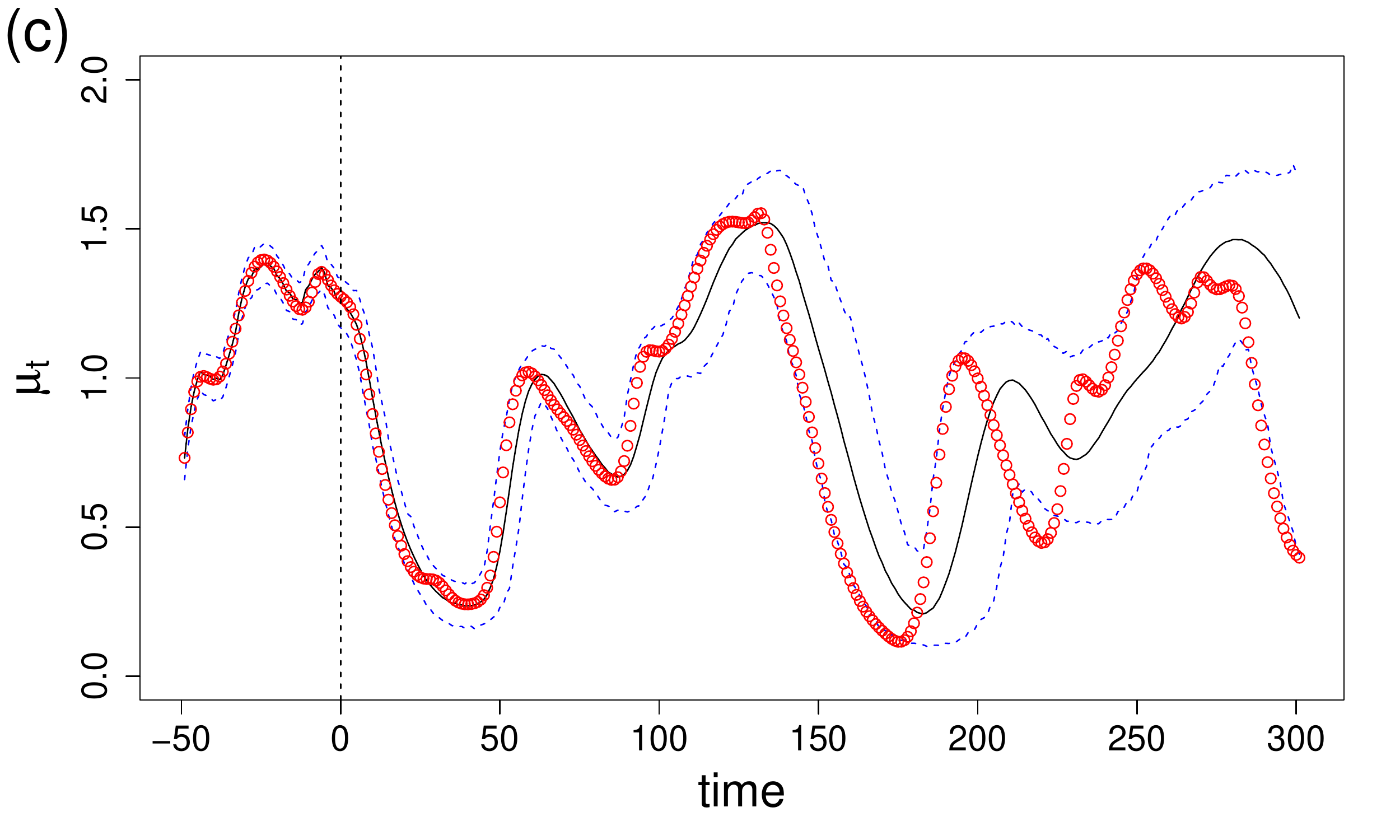}
  \includegraphics[width=0.48\textwidth]{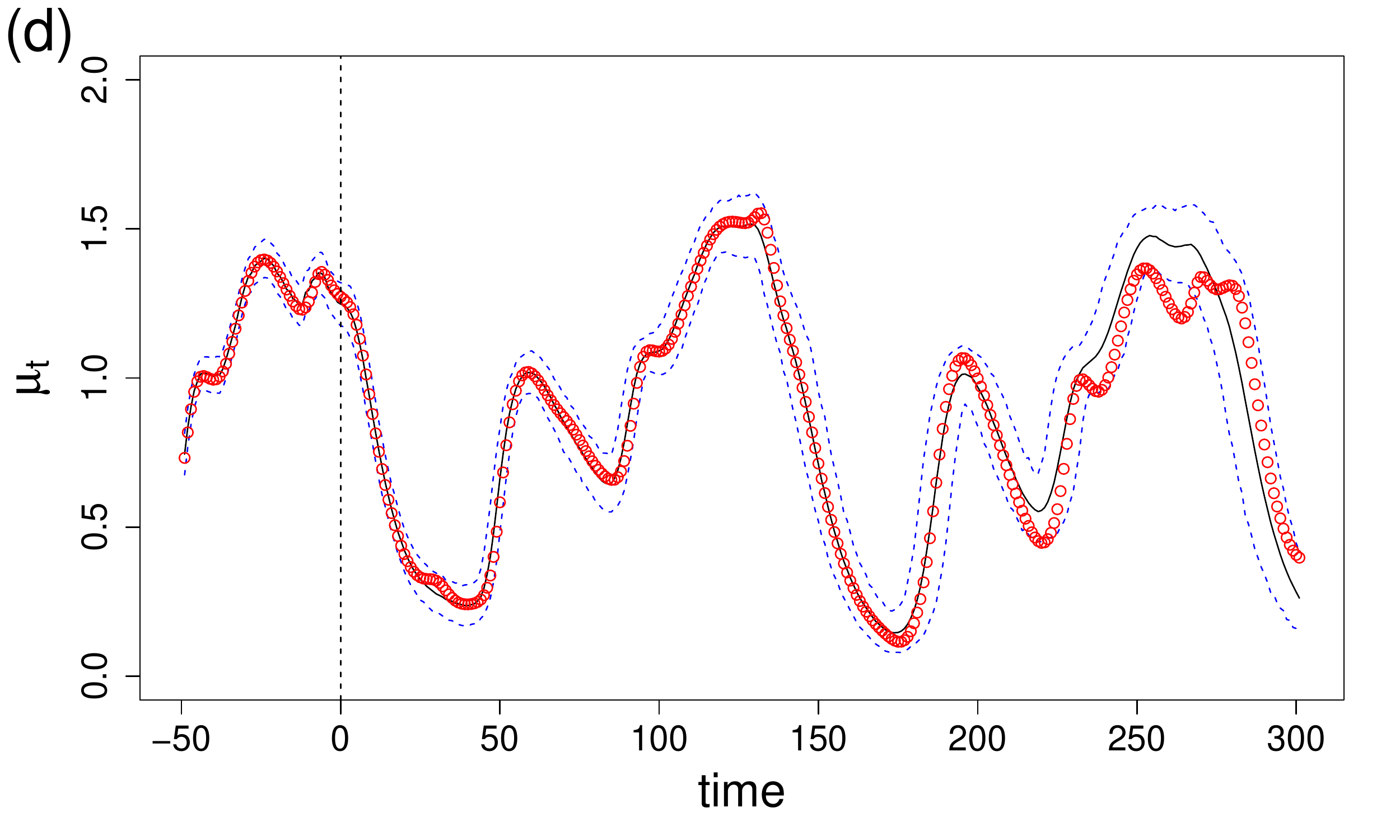}
  \caption{Multiple-step forecast of the Mackey-Glass time series for two sets of the parameters, $(\alpha,\gamma,\tau)$; (a,b) (0.305,0.092,22.19), and (c,d) (0.227,0.098,39.83). (a,c) are computed from RNN, and (b,d) are VI-RNN with $\lambda = 1$. The solid and dashed lines, respectively, denote the expectation and 95\% prediction interval, and the circles (${\color{red}\circ}$) are the ground truth. The vertical dashed line denotes the starting point of the simulation.} \label{fig:MG_multi_step}
\end{figure}

Figure \ref{fig:MG_multi_step} shows the multiple-step forecasts from RNN and VI-RNN for two trajectories. The Monte Carlo simulations are performed with $N_s = 1,000$ samples. To initiate both RNN and VI-RNN, $\bm{Y}_{-200:0}$ is used. The only difference between the Monte Carlo simulations of RNN and VI-RNN is the marginalization over $\bm{z}$. The advantage of the variational inference model is clearly shown in the multiple-step forecast task. Although RNN makes good predictions at a short forecast horizon, it quickly starts to deviate from the ground truth for $t > 100$. On the other hand, the multiple-step prediction of VI-RNN stays very close to the ground truth for the simulation horizon shown in figure \ref{fig:MG_multi_step}. It is also shown that the model uncertainty, represented by the 95\% prediction interval, of VI-RNN stays much tighter than that of RNN over the forecast horizon. In VI-RNN, we first make an inference about the dynamical system using a history data, \emph{e.g.}, $q(\bm{z}|\bm{Y}_{-200:0})$, and the outcome of the inference, $\bm{z}$, is supplied to VI-RNN as an additional input data. The results suggest that $\bm{z}$ acts to constrain VI-RNN to follow the dynamics identified from the history data.

\begin{table}
\center{
\caption{Empirical coverage probability, $\text{CP}_p$} \label{tbl:MG_cover}
\begin{tabular}{c|ccccc}
\hline \hline
 $p$ & 0.6 & 0.7 & 0.8 & 0.9 & 0.95 \\
 \hline
 RNN & 0.61 & 0.71 & 0.81 & 0.91 & 0.96 \\
 VI-RNN & 0.60 & 0.70 & 0.80 & 0.90 & 0.95 \\
 \hline \hline 
 \end{tabular}
}
\end{table}

For a more quantitative evaluation of the predictive distribution, the empirical coverage probability is shown in Table \ref{tbl:MG_cover}. The empirical coverage probability is computed as
\begin{equation}
\text{CP}_p = \frac{1}{T \, N_{test}} \sum_{k=1}^{N_{test}} \sum_{t=1}^T \chi_{p}(y^k_t).
\end{equation}
Here, $p$ is the coverage level of the prediction interval ($\mathcal{I}_p$), $N_{test}$ is the number of testing trajectories, $T$ is the prediction horizon, $\chi_p(y)$ is an indicator function, which is one if $y \in \mathcal{I}_p$ and zero otherwise. The prediction interval, $\mathcal{I}_p = [L_p,U_p]$, is computed from the inverse empirical cumulative distribution,
\[
L_p = \Phi^{-1}\left(\frac{1}{2}(1-p)\right), ~\text{and}~U_p = \Phi^{-1}\left(\frac{1}{2}(1+p)\right),
\]
where $\Phi$ denote the empirical cumulative distribution. Multiple-step forecasts are performed from five different initial conditions, $t_0 = (300,350,400,450,500)$, for the entire 100 trajectories in $\mathcal{D}_V$, which makes $N_{test} = 500$, and the empirical coverage probability is computed for the prediction horizon $T=500$. It is shown that the empirical coverage probabilities of both RNN and VI-RNN are very close to the $p$-value of the predictive interval, indicating that, in terms of the coverage, both models seem to provide reliable predictions of the time evolution of the probability distributions.

\begin{figure}
  \centering
  \includegraphics[width=0.48\textwidth]{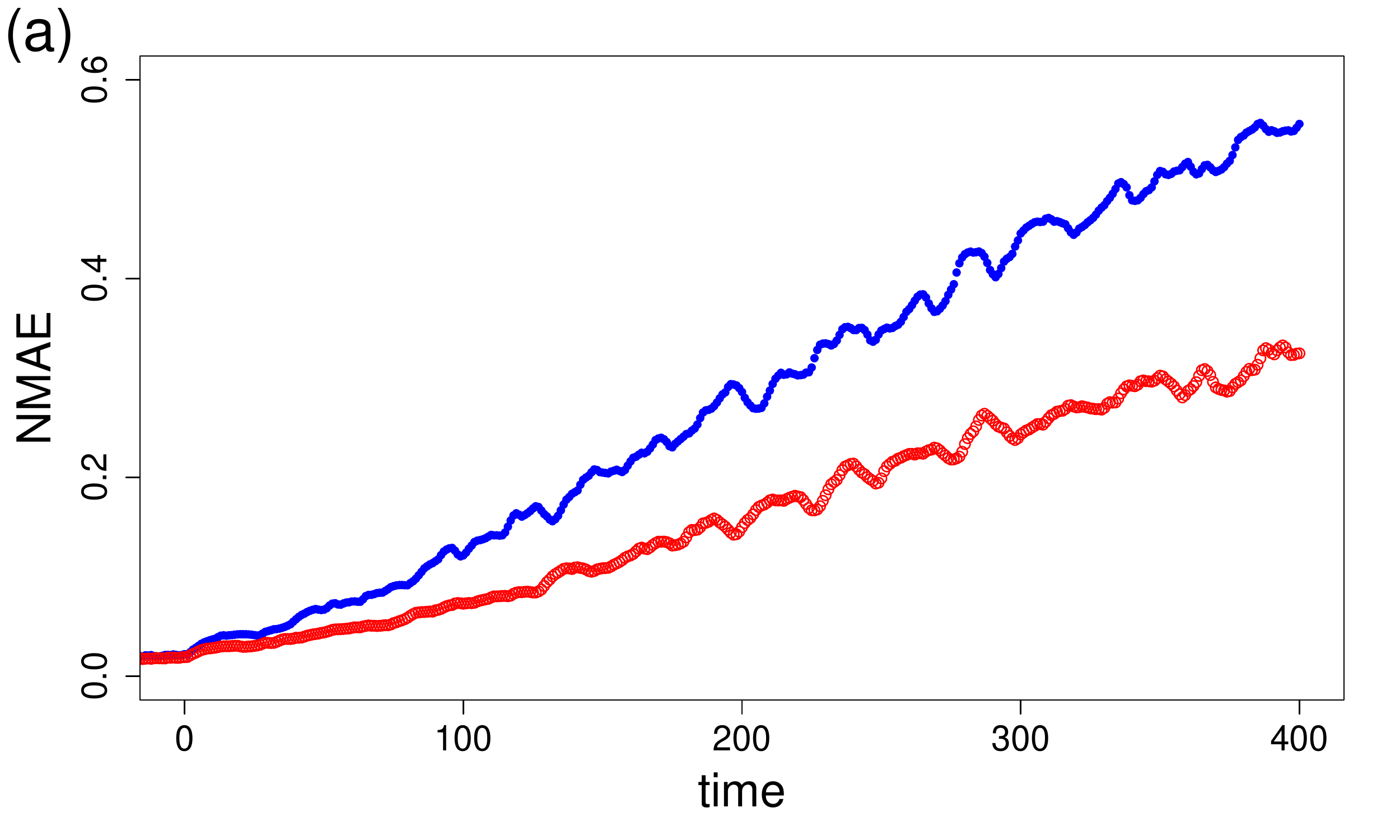}
  \includegraphics[width=0.48\textwidth]{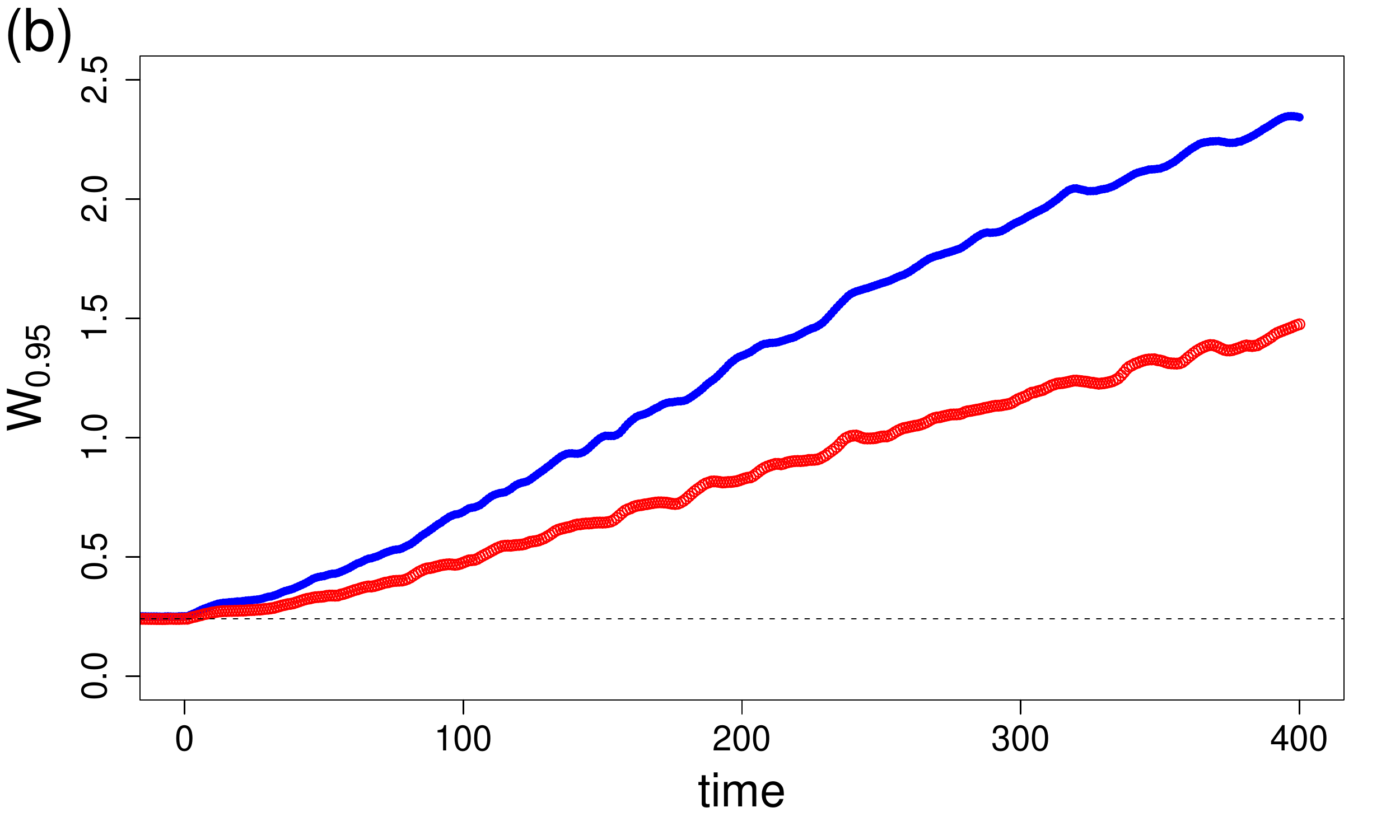}
  \caption{Temporal growth of (a) normalized mean absolute errors and (b) widths of 95\% prediction interval of RNN ({\color{blue}$\bullet$}) and VI-RNN trained with $\lambda=1$ (${\color{red}\circ}$) for the multiple-step forecast. The dashed line in (b) denotes the noise level.} \label{fig:MG_multistep_error}
\end{figure}

For a more quantitative comparison, the temporal growth of a normalized mean absolute error (NMAE) and a normalized width of the 95\% prediction interval ($W_{0.95}$) are shown in figure \ref{fig:MG_multistep_error}. Because the ground truth probability distribution is a Gaussian distribution, of which mean is the noiseless trajectory, $\phi(t \delta t)$, NMAE and $W_{0.95}$ provide more direct information about the accuracy of the predictive distribution.
The metrics are defined as
\begin{align}
\text{NMAE}(t) &= \frac{1}{N_{test}}\sum_{k=1}^{N_{test}} \sum_{t=1}^T \frac{|E[y^k_t]-\phi^k(t\delta t)|}{std(\phi^k)},\\
W_{0.95}(t) &= \frac{1}{N_{test}}\sum_{k=1}^{N_{test}} \sum_{t=1}^T  \frac{U_{0.95}(y^k_t)-L_{0.95}(y^k_t)}{std(\phi^k)},
\end{align}
in which $std(\phi)$ denotes the standard deviation of a trajectory $\phi$. It is clearly shown that the multiple-step forecast of RNN shows much more rapid growth of NMAE and $W_{95}$ than those of VI-RNN. At $t=400$, NMAE of VI-RNN is only about 58\% of RNN. 

\begin{figure}
  \centering
  \includegraphics[width=0.48\textwidth]{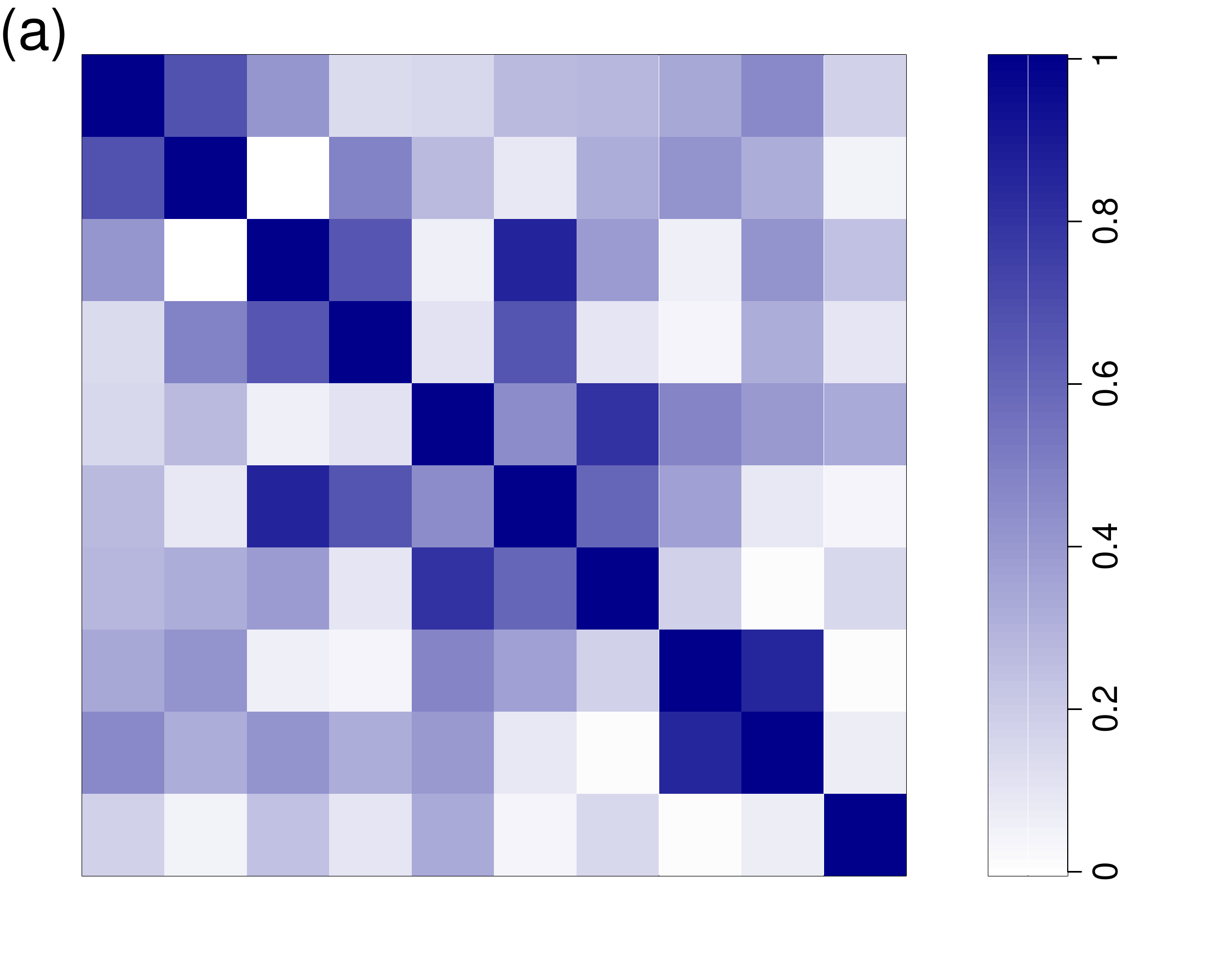}
  \includegraphics[width=0.48\textwidth]{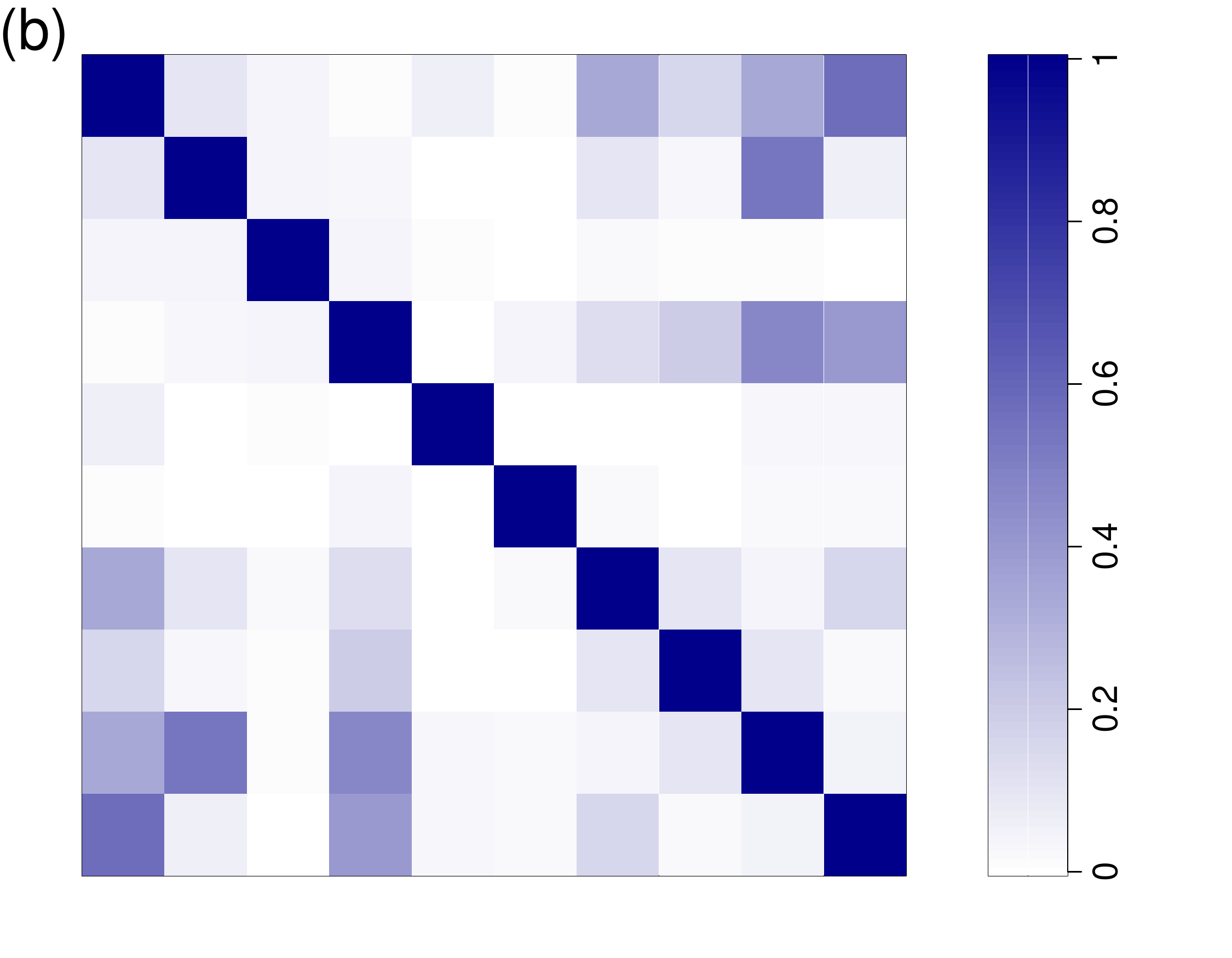}\\
  \includegraphics[width=0.48\textwidth]{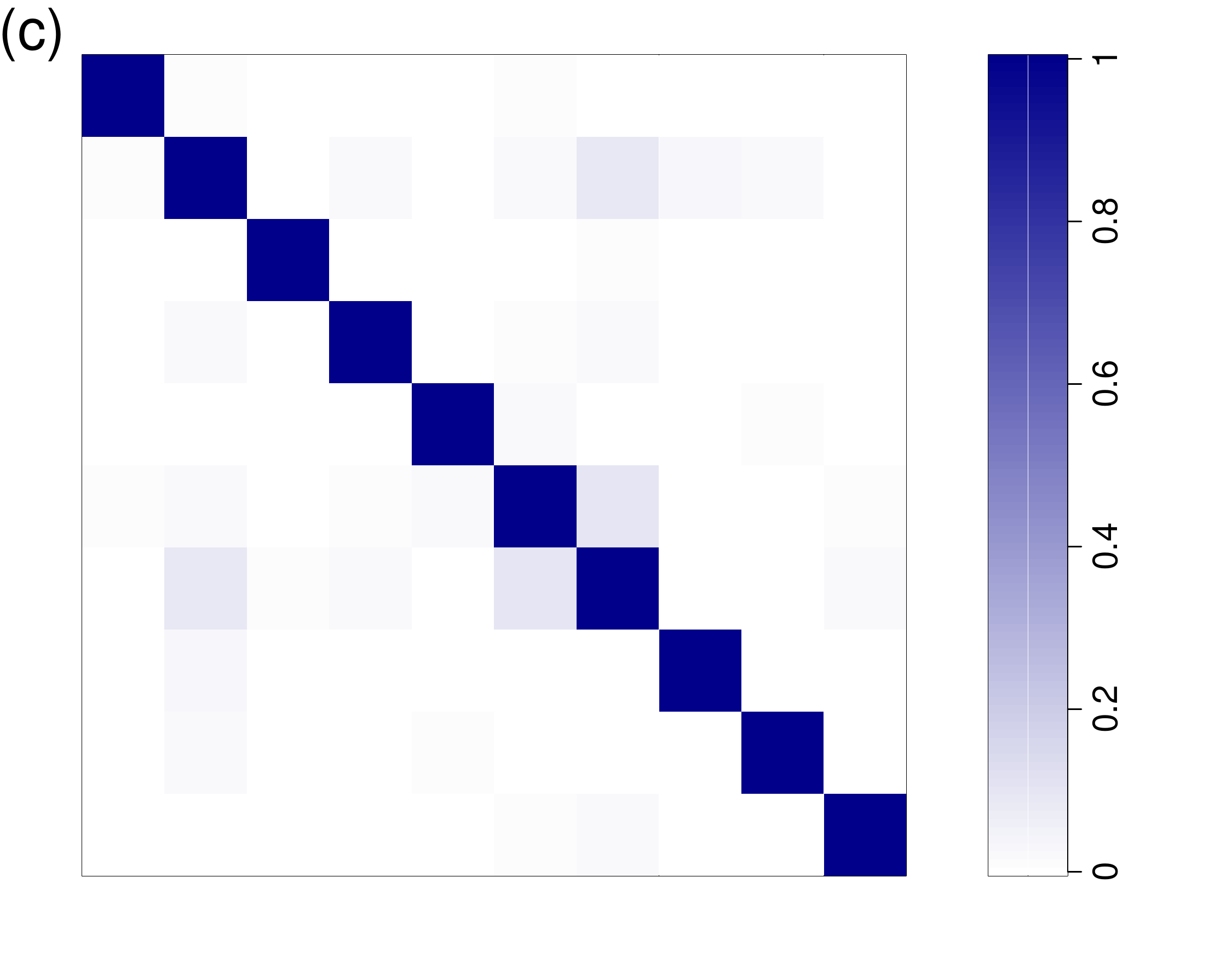}
  \includegraphics[width=0.48\textwidth]{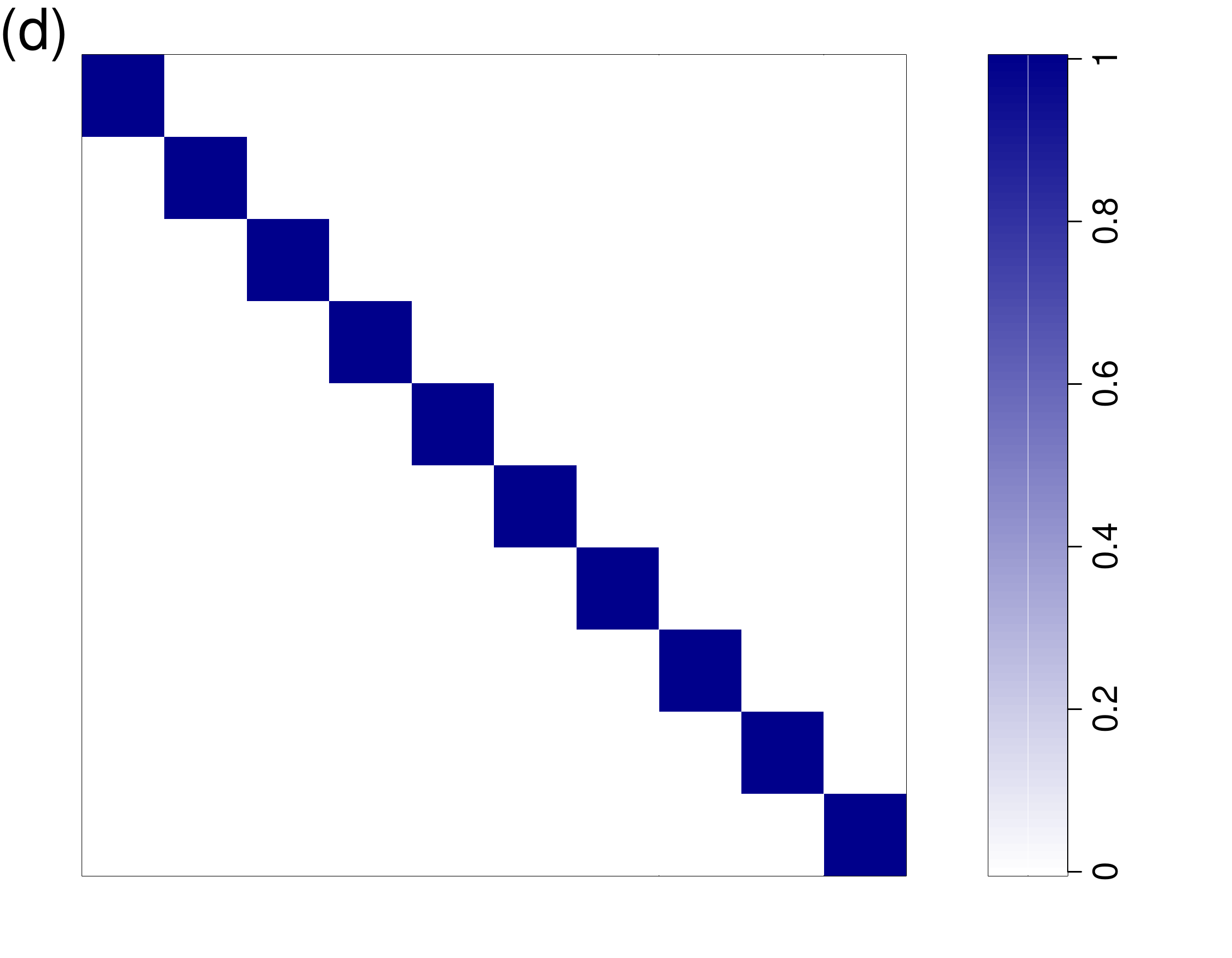}
  \caption{Absolute value of correlation between $\bm{z}$ for VI-RNN trained with (a) $\lambda = 0.01$, (b) $\lambda = 0.1$, and (c) $\lambda = 1.0$, and (d) $\lambda = 10.0$.} \label{fig:MG_q_cor}
\end{figure}

To have a better understanding on VI-RNN, the approximate posterior distributions of the latent variable, $q(\bm{z}|\bm{Y})$, are computed from the training data, $\mathcal{D}_T$. The approximate posterior distributions are computed at 5 different time stamps for the entire 400 trajectories in $\mathcal{D}_T$, which makes the total number of samples is 2,000. Note that $\bm{z}$ is sampled across different trajectories, i.e., the results are marginalized over the data distribution. Figure \ref{fig:MG_q_cor} shows absolute value of the correlation coefficients between the latent variables, \emph{i.e.}, $|Cor(\bm{z},\bm{z})|$, for four different penalty parameters, $\lambda$. The contribution of the Kullback-Leibler divergence to the loss function (\ref{eqn:lambda_loss}) increases linearly with $\lambda$, which pushes $q(\bm{z}|\bm{Y})$ towards the prior distribution, $p(\bm{z})$. Because $p(\bm{z})$ is given by an independent Gaussian distribution, it is shown that $\bm{z}$ also becomes linearly independent from each other as $\lambda$ increases. 

\begin{figure}
  \centering
  \includegraphics[width=0.48\textwidth]{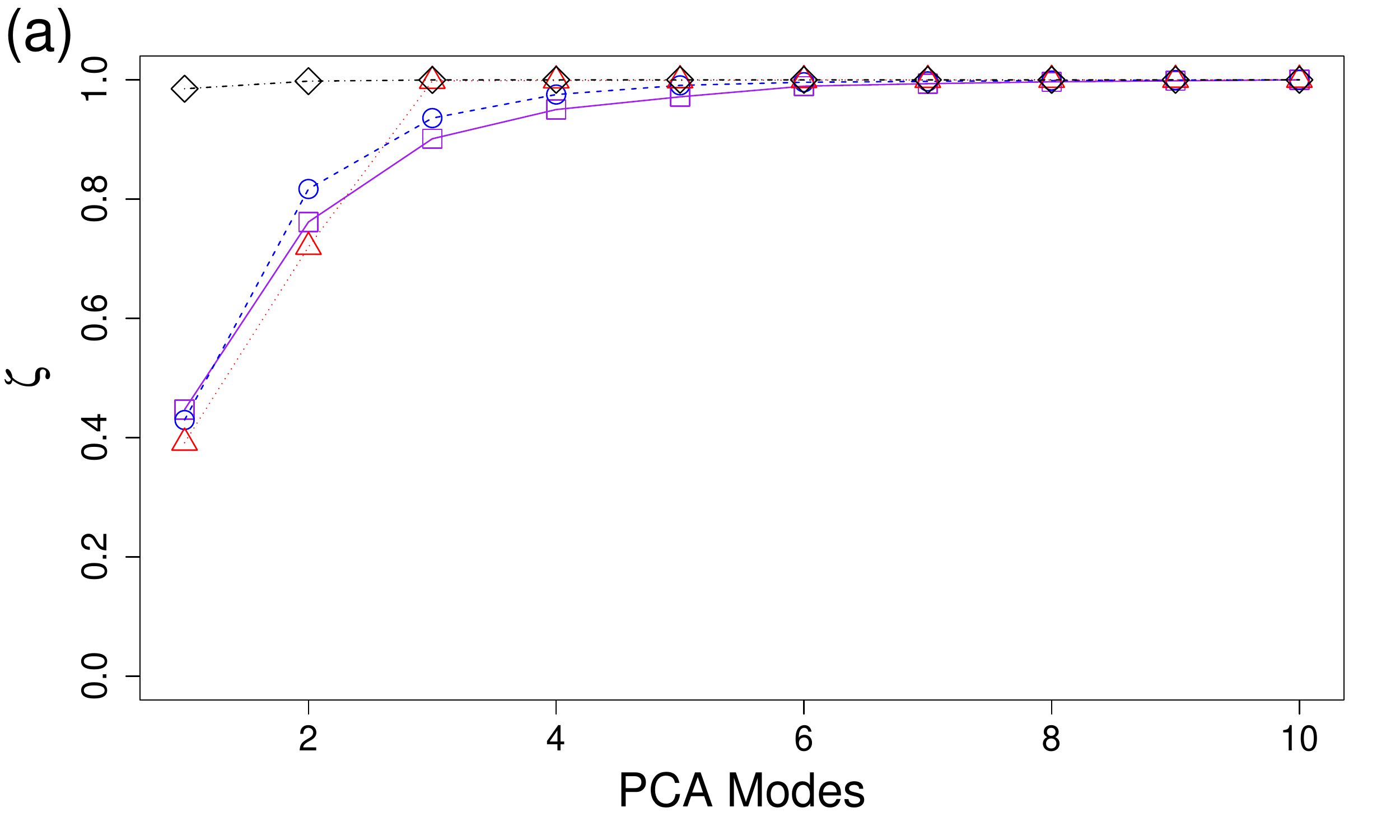}
  \includegraphics[width=0.48\textwidth]{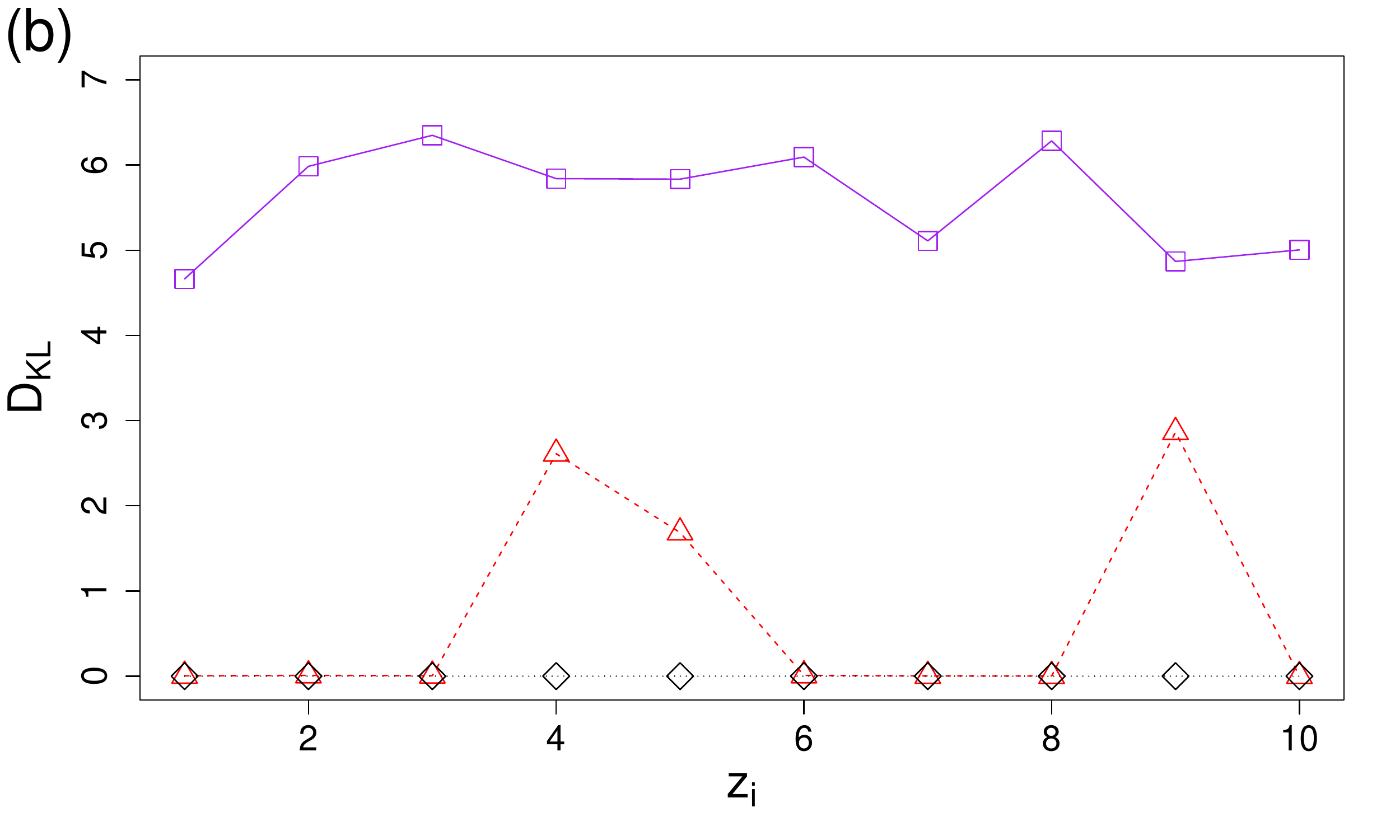}
  \caption{(a) Cumulative eigenvalues of PCA modes of $\bm{m}_q$ for $\lambda = 0.01$ (${\color{magenta}\Box}$), 0.1 (${\color{blue}\circ}$), 1.0 (${\color{red}\triangle}$), and 10.0 ($\diamond$). (b) Element-wise Kullback-Leibler divergence for $\lambda = 0.01$ (${\color{magenta}\Box}$), 1.0 (${\color{red}\triangle}$), and 10.0 ($\diamond$). } \label{fig:MG_PCA_KL}
\end{figure}

In the posterior distribution, the mean components ($\bm{m}_q$) contain information about the physical parameters, \emph{e.g.}, $(\alpha,\gamma,\tau)$, while the standard deviation ($\bm{\sigma}_q$) represents uncertainty around the estimation. To investigate the behavior of $\bm{z}$, a principal component analysis (PCA) is performed for $\bm{m}_q$ from the same $\bm{z}$'s used to compute the correlation (Figure \ref{fig:MG_q_cor}). Figure \ref{fig:MG_PCA_KL} (a) shows the cumulative eigenvalues of the PCA components of $\bm{m}_q$;
\begin{equation}
\zeta_i = \frac{\sum_{j=1}^i \nu_j}{\sum_{j=1}^{N_q} \nu_j},
\end{equation}
in which $\nu_j$ is the $j$-th eigenvalue of PCA. The cumulative eigenvalue, $\zeta_i$, indicates the fraction of variations in the data captured by the first $i$-th PCA modes. It is shown that, for the range of $\lambda$ in this study, the first thee modes can represent more than 90\% of the variations in the data, \emph{i.e.}, $\zeta_3 > 0.9$ for $0.01 \le \lambda \le 10$. Note that the Mackey-Glass time series has three random parameters. In particular, for $\lambda = 1$, $\zeta_3$ becomes larger than 0.998, indicating the variations in the data can be almost completely explained by only the first three modes. When $\lambda = 10$, the posterior distribution becomes almost isotropic, which makes $\zeta_1 > 0.98$.

Figure \ref{fig:MG_PCA_KL} (b) shows an element-wise Kullback-Leibler divergence,
\begin{equation}
D_{{KL}_i} = \frac{1}{N_{test}} \sum_{k=1}^{N_{test}}\frac{1}{2} \left( \frac{{m^{(k)}_{q_i}}^2+{\sigma^{(k)}_{q_i}}^2}{\sigma^2_z} -1 \right)- \log \frac{\sigma^{(k)}_{q_i}}{\sigma_z}.
\end{equation}
Here, the standard deviation of the prior $p(\bm{z})$ is $\sigma_z = 1$.  When $\lambda = 1$, the distributions of only three elements of $\bm{z}$ ($z_4,z_5,z_9$) become significantly different from $p(\bm{z})$. When $\lambda = 10$, $q(\bm{z}|\bm{Y})$ is almost equal to $p(\bm{z})$, \emph{i.e.}, the latent variable becomes just a random noise. 

\begin{figure}
  \centering
  \includegraphics[width=0.49\textwidth]{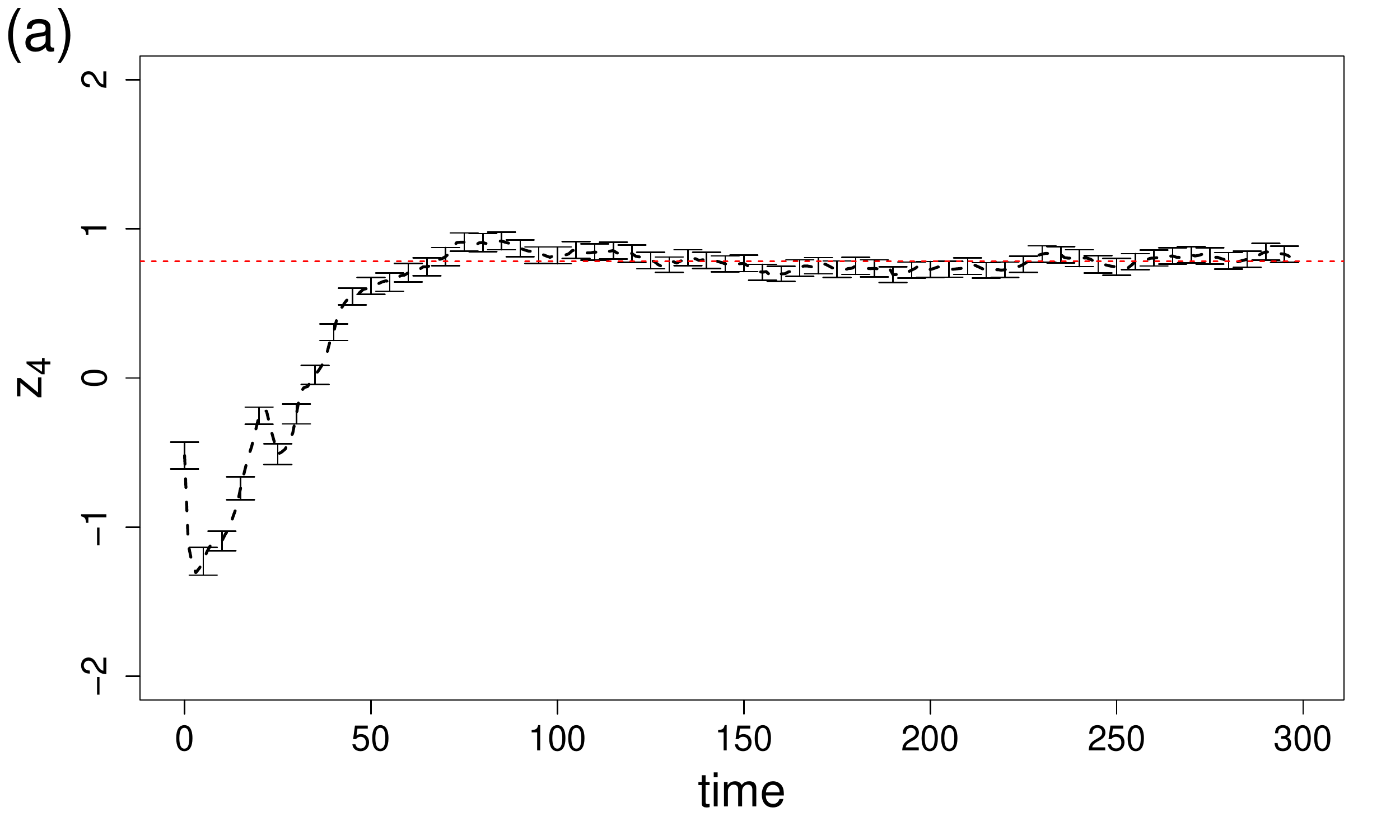}
  \includegraphics[width=0.49\textwidth]{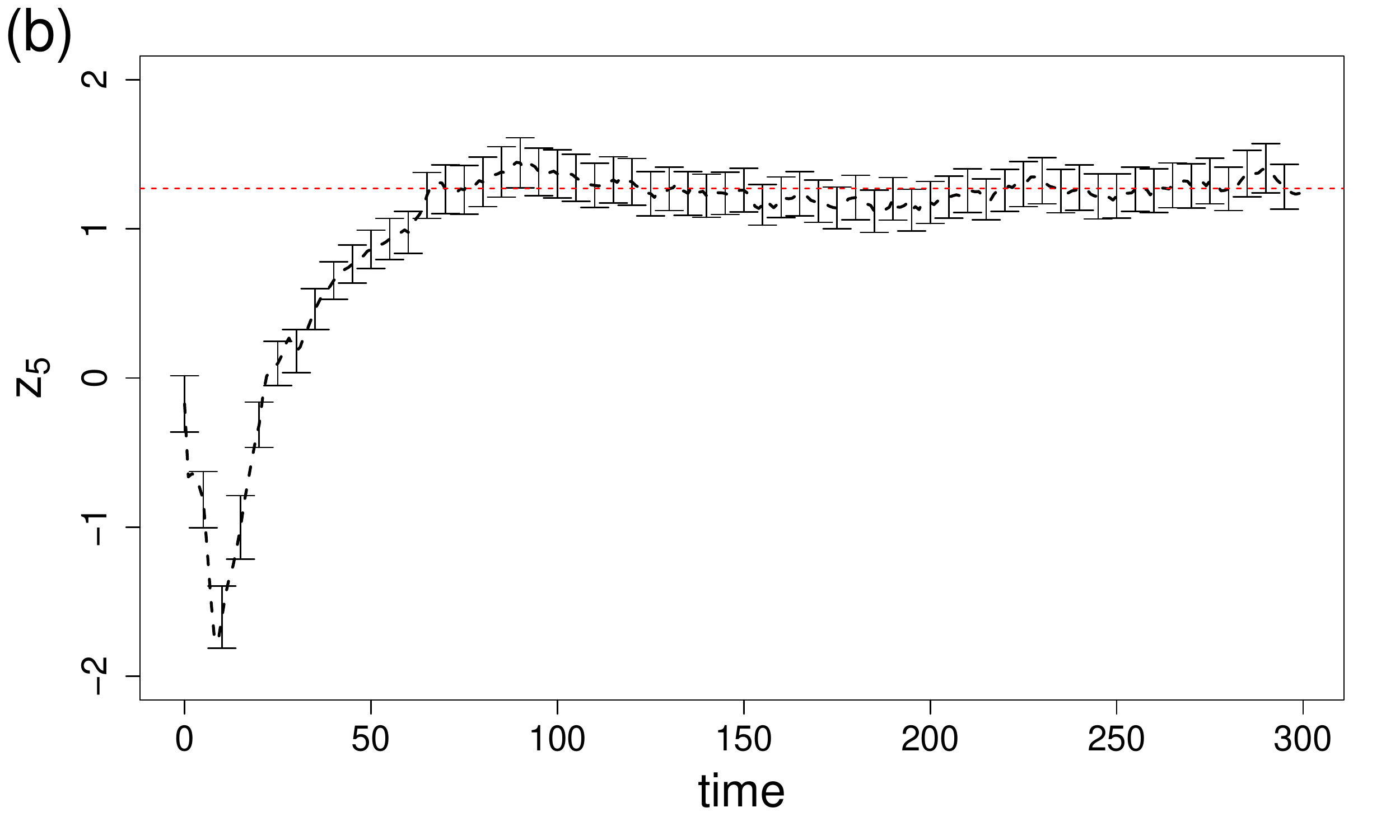}\\
  \includegraphics[width=0.49\textwidth]{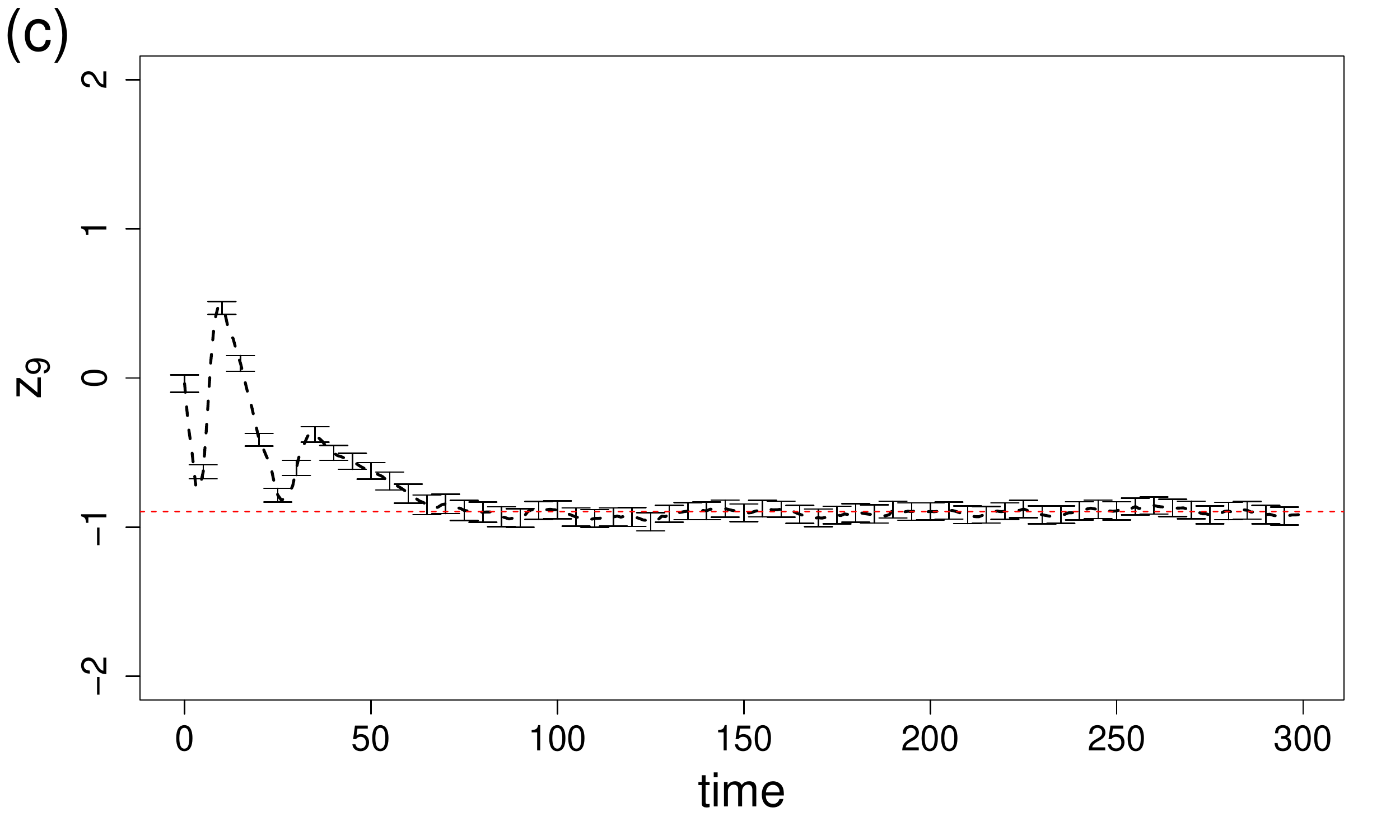}
  \includegraphics[width=0.49\textwidth]{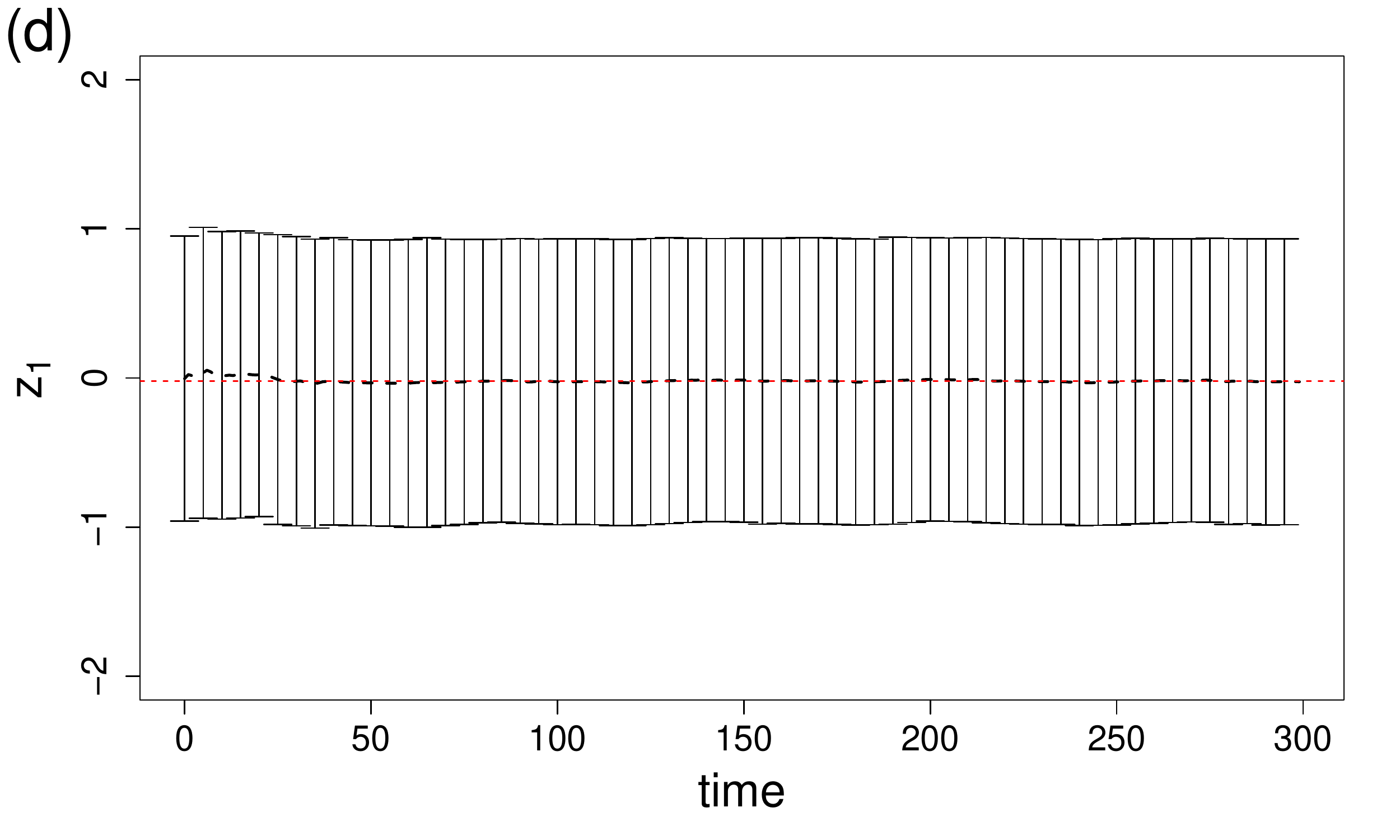}
  \caption{The posterior distribution, $q(z_i|\bm{Y}_{0:t})$, as a function of $t$ for a sample trajectory in $\mathcal{D}_V$. VI-RNN is trained with $\lambda = 1$; (a) $z_4$, (b) $z_5$, (c) $z_9$, and (d) $z_1$. The dashed line is $m_q$ and $\sigma_q$ is denoted as the error bars. The horizontal line is the time average of $m_q$ over $t=200 \sim 300$. In (a,b,c), it is shown that, due to the effects of the initial condition ($\bm{h}_0$) of the encoder RNN, there is an initial adjustment period, e.g, $t < 80$, before $\bm{z}$ converges to a steady state.} \label{fig:MG_time_latent}
\end{figure}

As discussed in \cite{Yeo19}, the dynamical system of the internal state of an RNN is a relaxation process. Hence, after a characteristic time scale, effects of an impulse vanishes and the internal state of an RNN recovers a stationary dynamics \cite{Yeo19b}. Considering this stationary nature of RNN, VI-RNN first computes $\bm{h}^{enc}_{t+1}$ from a sequence, $\bm{Y}_{0:t}$, and condition the approximate posterior on the code, \emph{i.e.}, $q(\bm{z}|\bm{h}^{enc}_{t+1}) = q(\bm{z}|\bm{Y}_{0:t})$. It is assumed that $q(\bm{z}|\bm{Y}_{0:t})$ will be invariant once $t$ is larger than the characteristic timescale of the process. To test this assumption, in figure \ref{fig:MG_time_latent}, $q(\bm{z}|\bm{Y}_{0:t})$ are computed for $t = 0,\cdots,300$;
\begin{align}
\bm{h}^{enc}_{i+1} &= \bm{\Psi}_h^{enc}(\bm{y}_{i},\bm{h}^{enc}_{i}), \\
\bm{z}_{i} &\sim \widehat{\bm{\eta}}(\bm{h}^{enc}_{i+1}), ~\text{for}~i = 1,\cdots,300.
\end{align}
In figure \ref{fig:MG_time_latent} (a--c), it is shown that indeed $q(\bm{z}|\bm{Y}_{0:t})$ converges to a stable distribution for $t > 100$. It is also shown that $\sigma_q$ of the three non-trivial latent variables ($z_4,z_5,z_9$) are much smaller than the prior, $\sigma_z$. On the other hand, figure \ref{fig:MG_time_latent} (d) shows that the distributions of the other seven trivial latent variables follow that of the prior, $\mathcal{N}(0,1)$, from the begining and remain unchanged.

\begin{figure}
  \centering
  \includegraphics[width=0.49\textwidth]{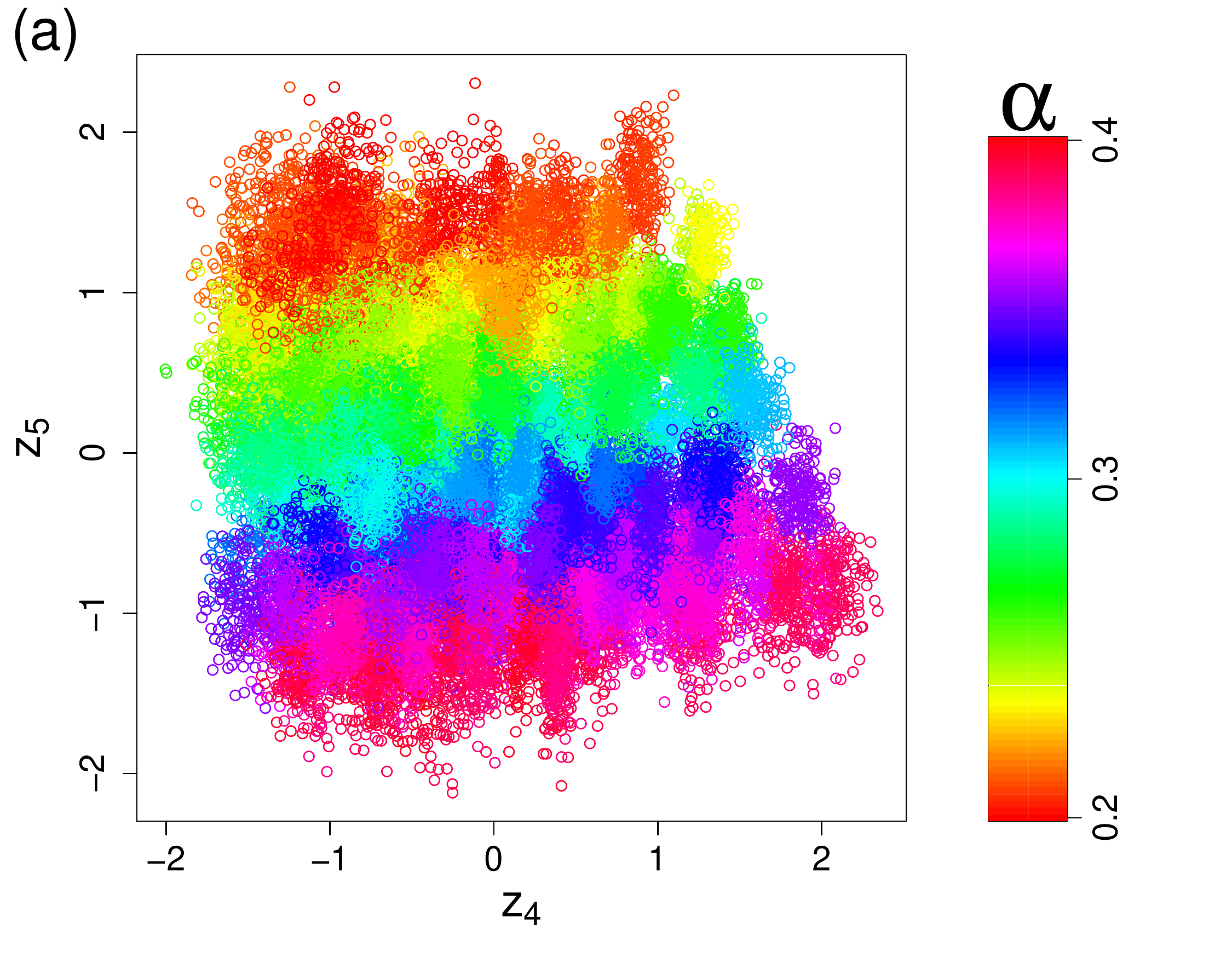}
  \includegraphics[width=0.49\textwidth]{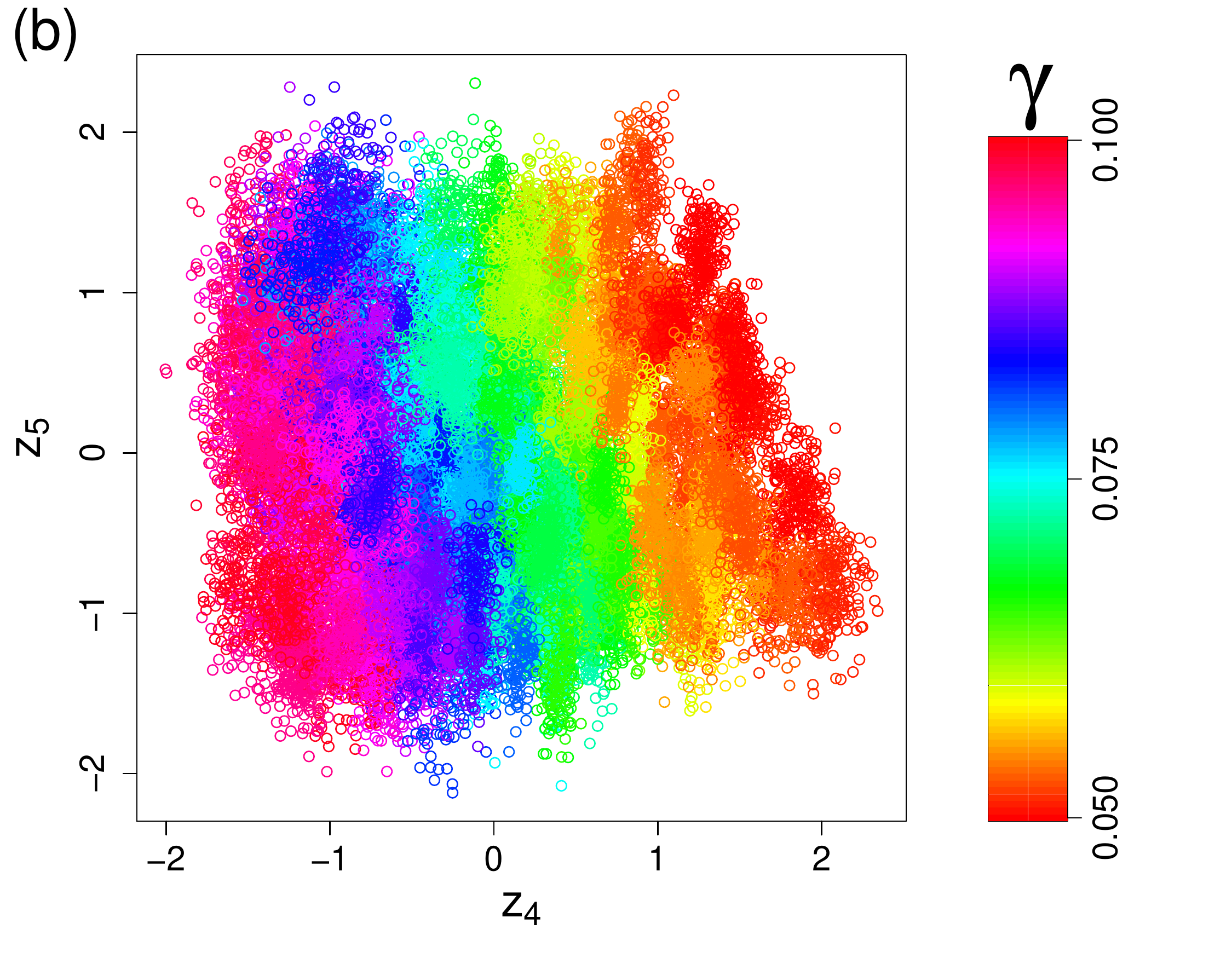}
  \includegraphics[width=0.49\textwidth]{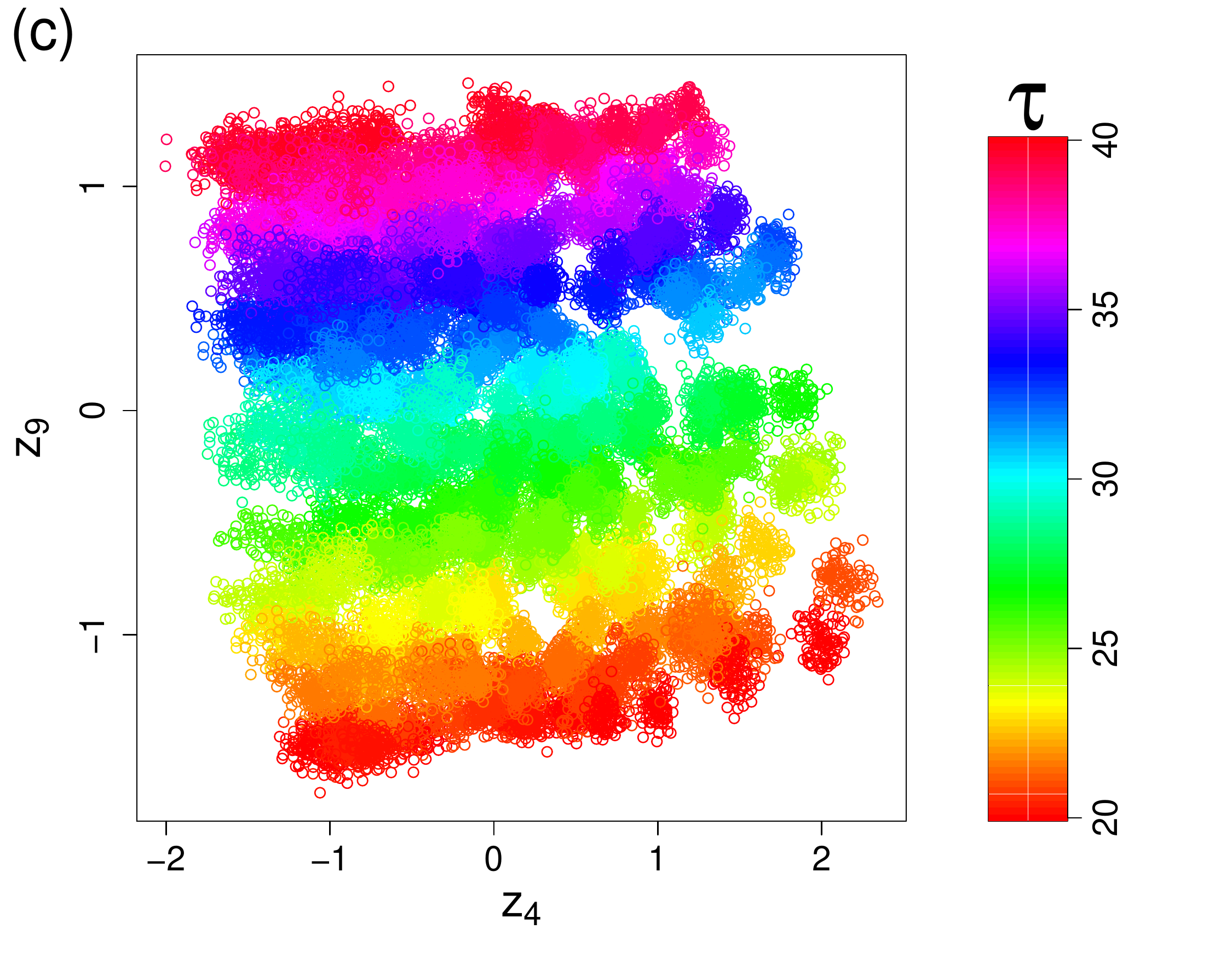}
  \caption{The posterior distribution projected onto the three non-trivial dimensions; (a) $z_4-z_5$, (b) $z_4-z_5$, and (c) $z_4-z_9$ planes. The symbols denote the samples from $q(\bm{z}|\bm{Y}^k)$ for all $\bm{Y}^k \in \mathcal{D}_T$. The symbols are color-coded with (a) $\alpha$, (b) $\gamma$, and (c) $\tau$ values of the corresponding trajectory.} \label{fig:MG_latent_param}
\end{figure}

\begin{table}
\center{
\caption{Correlation between the latent variables and the random parameters} \label{tbl:MG_corr}
\begin{tabular}{c|ccc}
\hline \hline
& $\alpha$ & $\gamma$ & $\tau$\\
\hline
$z_4$ &0.28&$\bm{0.93}$ &-0.22\\
$z_5$ & -$\bm{0.95}$ & -0.14 & -0.04\\
$z_9$ & -0.04 & -0.12 & $\bm{0.98}$\\ 
\hline \hline
\end{tabular}
}
\end{table}

The results so far suggest that VI-RNN is capable of correctly identifying the dimensions of the random parameters from the data in an unsupervised way, \emph{i.e.}, without giving any direct information about the dimensionality of the random parameters. To show a more direct evidence, in figure \ref{fig:MG_latent_param}, we show $\bm{z}$ projected onto the three non-trivial dimensions ($z_4,z_5,z_9$) and color-code with the random parameters. First, $q(\bm{z}|\bm{Y})$  are evaluated at  five different time stamps for all 400 trajectories in $\mathcal{D}_T$. Then, 20 samples are drawn from the each $q(\bm{z}|\bm{Y})$. Those samples are plotted on the $z_4-z_5$ and $z_4-z_9$ planes, and color-coded with the values of $(\alpha,\gamma,\tau)$ of the corresponding trajectories. It is clearly shown that those three non-trivial latent variables indeed capture the variations of the random parameters without having any prior information about the complex nonlinear dynamical system. Table \ref{tbl:MG_corr} shows the correlation between the random parameters and $\bm{z}$. It is shown that the correlation between a random parameter and the corresponding dimension of $\bm{z}$ is larger than 0.9.

\begin{table}
\center{
\caption{Correlation between the one-dimensional latent variable and the random parameters} \label{tbl:MG_corr_1d}
\begin{tabular}{c|ccc}
\hline \hline
& $\alpha$ & $\gamma$ & $\tau$\\
\hline
$z$ &0.02& 0.09 & $-\bm{0.99}$\\
\hline \hline
\end{tabular}
}
\end{table}

We now consider the case where the dimension of $\bm{z}$ is less than the dimension of the random parameters by making the latent variable one-dimensional, \emph{i.e.}, $z \in \mathbb{R}$. The correlation between $z$  and the random parameters is shown in Table \ref{tbl:MG_corr_1d} for VI-RNN trained with $\lambda = 1$. It is shown that $z$ is aligned with the time-delay parameter, $\tau$. The time-delay parameter has a central role in the dynamics of Mackey-Glass system and as, shown in (\ref{eqn:MG_param}), $\tau$ varies widely among the trajectories in $\mathcal{D}$. The result seems to suggest that when the dimension of $\bm{z}$ is less than the dimension of the random parameters, $\bm{z}$ is aligned with the random parameters, which have dominant effects on the dynamics.

\subsection{Forced Van der Pol Oscillator} \label{subsec:VDP}

\begin{figure}
  \centering
  \includegraphics[width=0.48\textwidth]{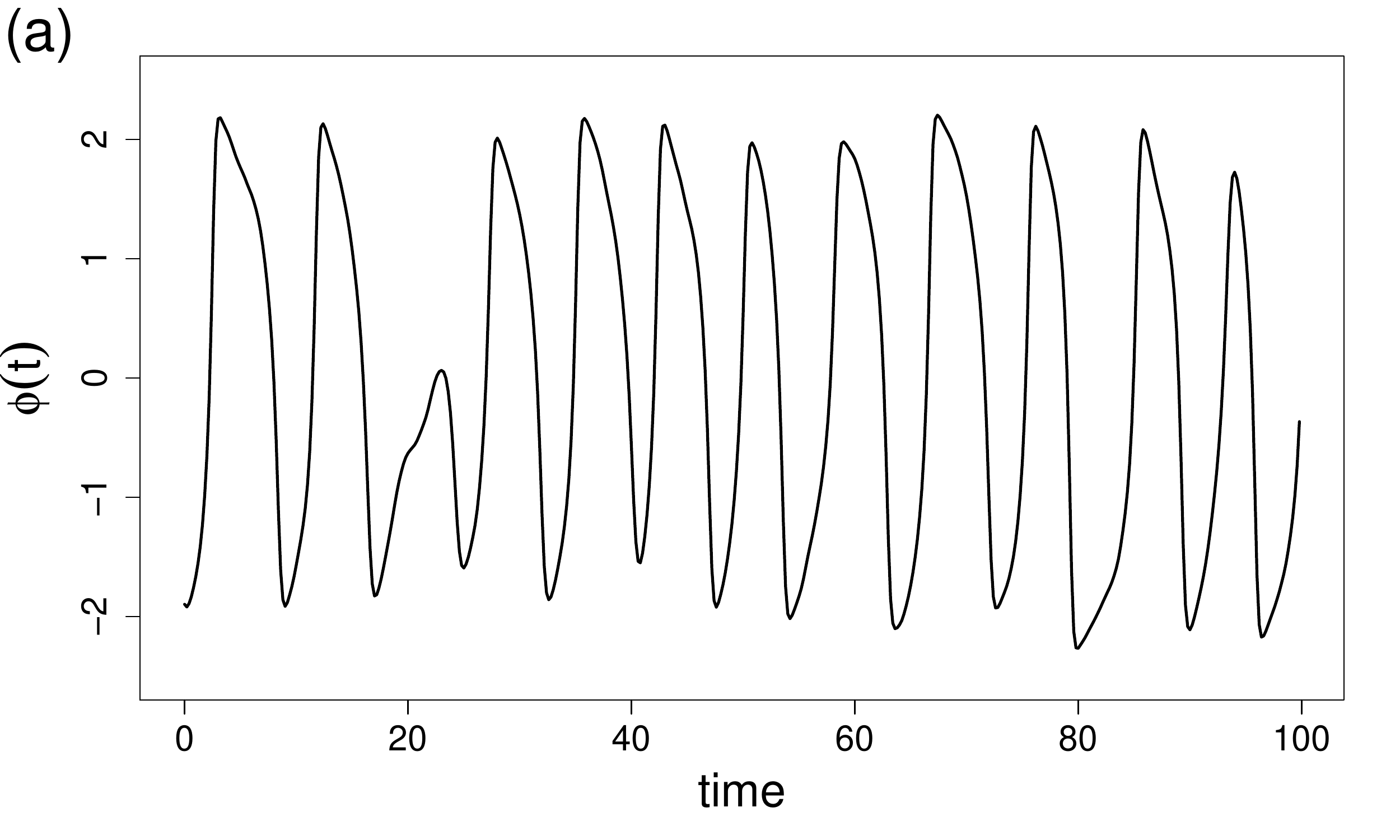}
  \includegraphics[width=0.48\textwidth]{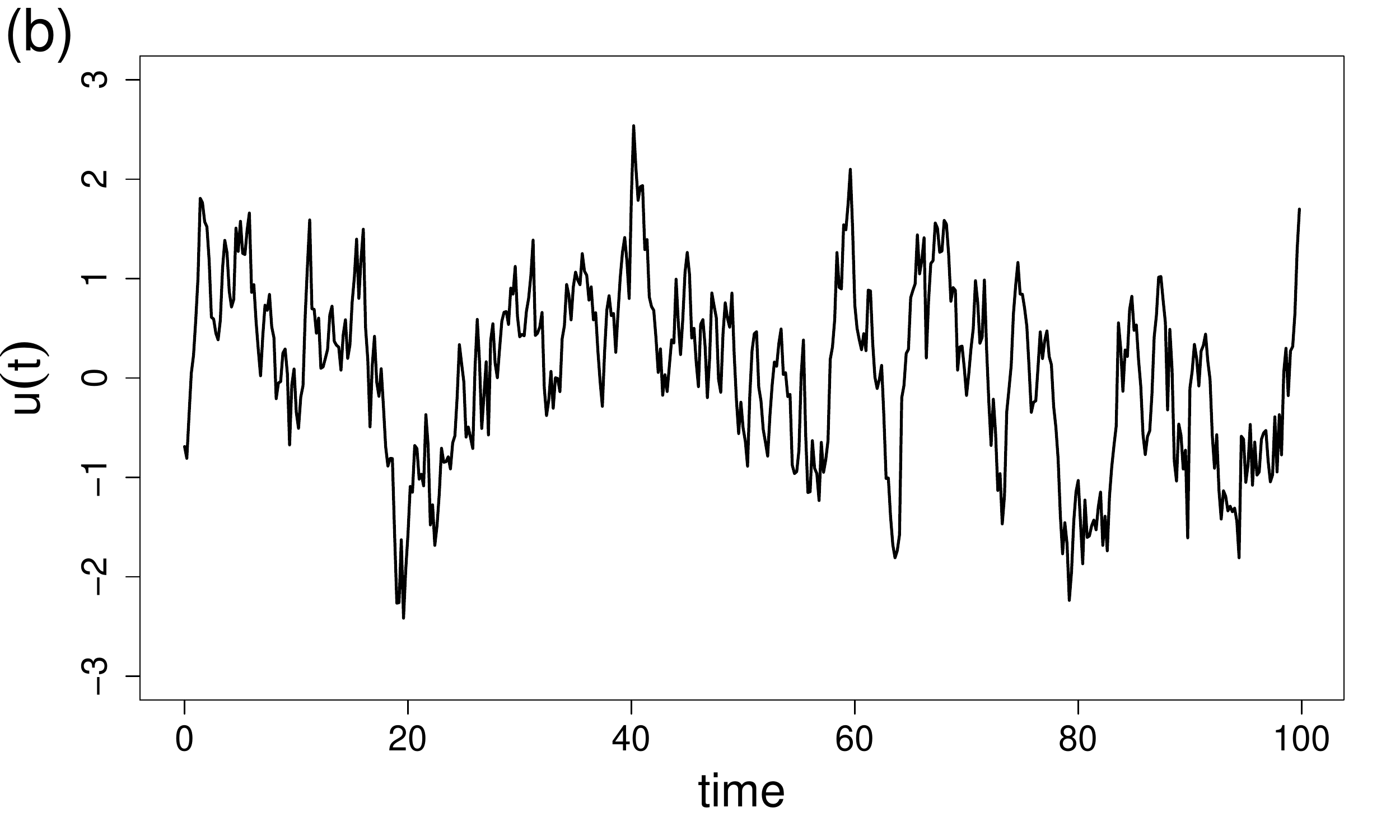}\\
  \includegraphics[width=0.48\textwidth]{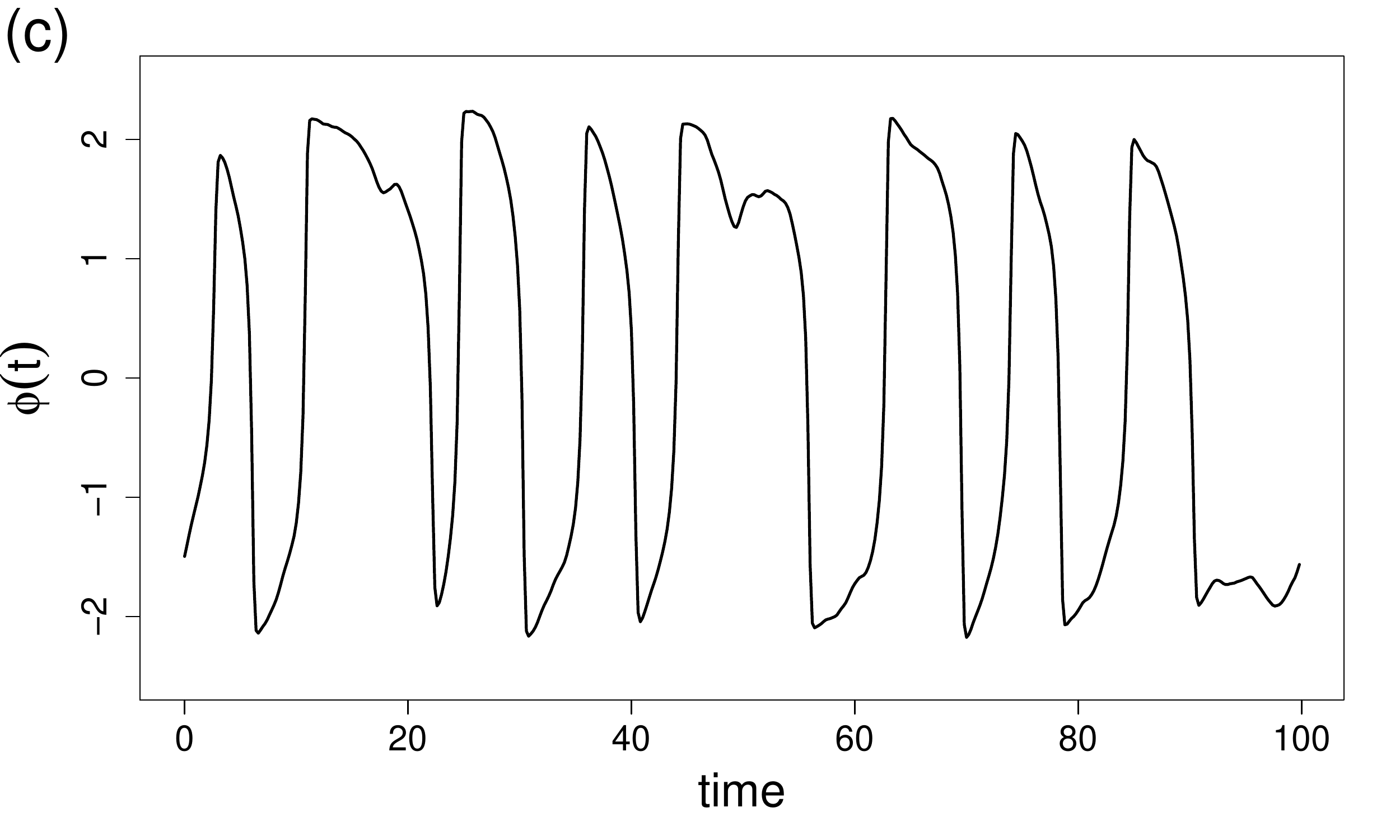}
  \includegraphics[width=0.48\textwidth]{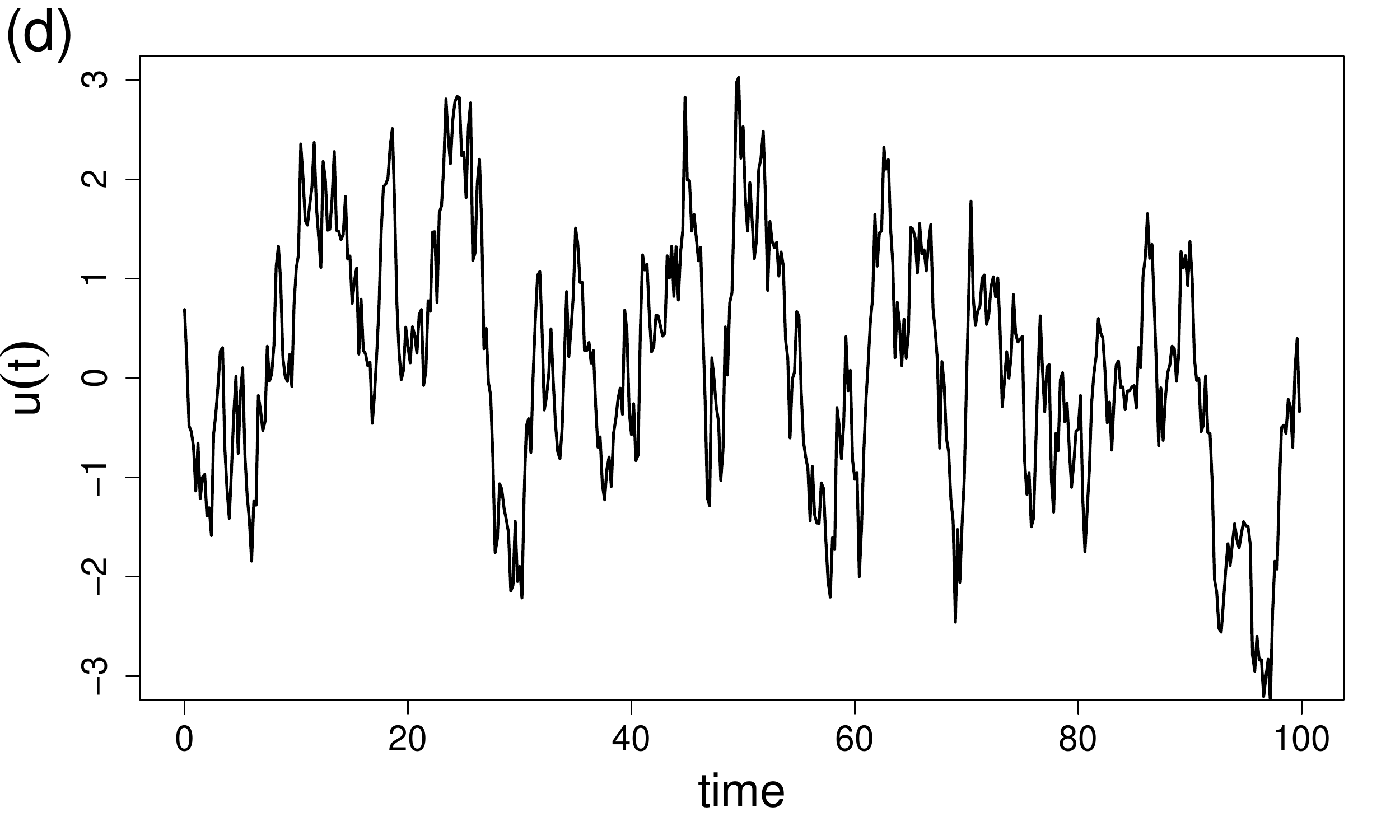}
  \caption{Sample trajectories of the forced Van der Pol oscillator (a,c) and the corresponding exogenous forcing (b,d) for different sets of the parameters; (a) $(\alpha,\gamma) = (0.68,1.77)$, (b) $\theta = 0.42$,  (c) $(\alpha,\gamma) = (0.94,3.63)$, (d) $\theta = 0.66$.} \label{fig:VDP_data}
\end{figure}

For the next example, we consider  a forced Van der Pol oscillator (VDP), which is given by the following equations,
\begin{equation} \label{eqn:VDP}
\frac{d^2 \phi(t)}{dt^2} -\gamma (1-\phi^2(t))\frac{d\phi(t)}{dt}+\phi(t) + \alpha u(t) = 0.
\end{equation} 
An Ornstein-Uhlenbeck process is used as the exogenous forcing, $u(t)$,
\begin{equation} \label{eqn:VDP_OU}
du = -\theta u dt + u_{ref}\sqrt{2\theta} dW,
\end{equation}
in which $W$ is the Wiener process and the stationary standard deviation of $u(t)$ is $u_{ref} = 1$. The parameters in (\ref{eqn:VDP}--\ref{eqn:VDP_OU}) are random variables,
\begin{equation}
\gamma \sim \mathcal{U}(1,4),~~\alpha \sim \mathcal{U}(0.25,1),~\text{and}~~\theta \sim \mathcal{U}(0.25,1).
\end{equation}
Now, the dataset, $\mathcal{D}$, consists of a tuple, $(\bm{Y}_{0:T},\bm{U}_{0:T})$. Here, we are interested in an inference on $\bm{Y}$ given $\bm{U}$. Hence, the random variable, $\theta$, which is used to generate $\bm{U}$, does not play a role in the inference. Because the forcing amplitude, $\alpha$, is a random variable, the effects of $u(t)$ on the dynamics of VDP vary from one trajectory to another, which makes it a challenging inference problem. Figure \ref{fig:VDP_data} shows two sample trajectories of $(\bm{Y},\bm{U})$.

The equations of the forced VDP, (\ref{eqn:VDP} -- \ref{eqn:VDP_OU}), are numerically integrated by using a third-order Adams-Bashforth method with a time step size of 0.001, and a time series is generated by downsampling to make the sampling interval, $\delta t = 0.2$. Then, a zero-mean Gaussian white noise is added as shown in (\ref{eqn:noisy_obs}). The standard deviation of the noise process is, $\sigma_\epsilon = 0.075$, which is about 5\% of the standard deviation of $\bm{Y} \in \mathcal{D}$. Note that the noise is added only to $\bm{Y}$, not $\bm{U}$.

\begin{figure}
  \centering
  \includegraphics[width=0.48\textwidth]{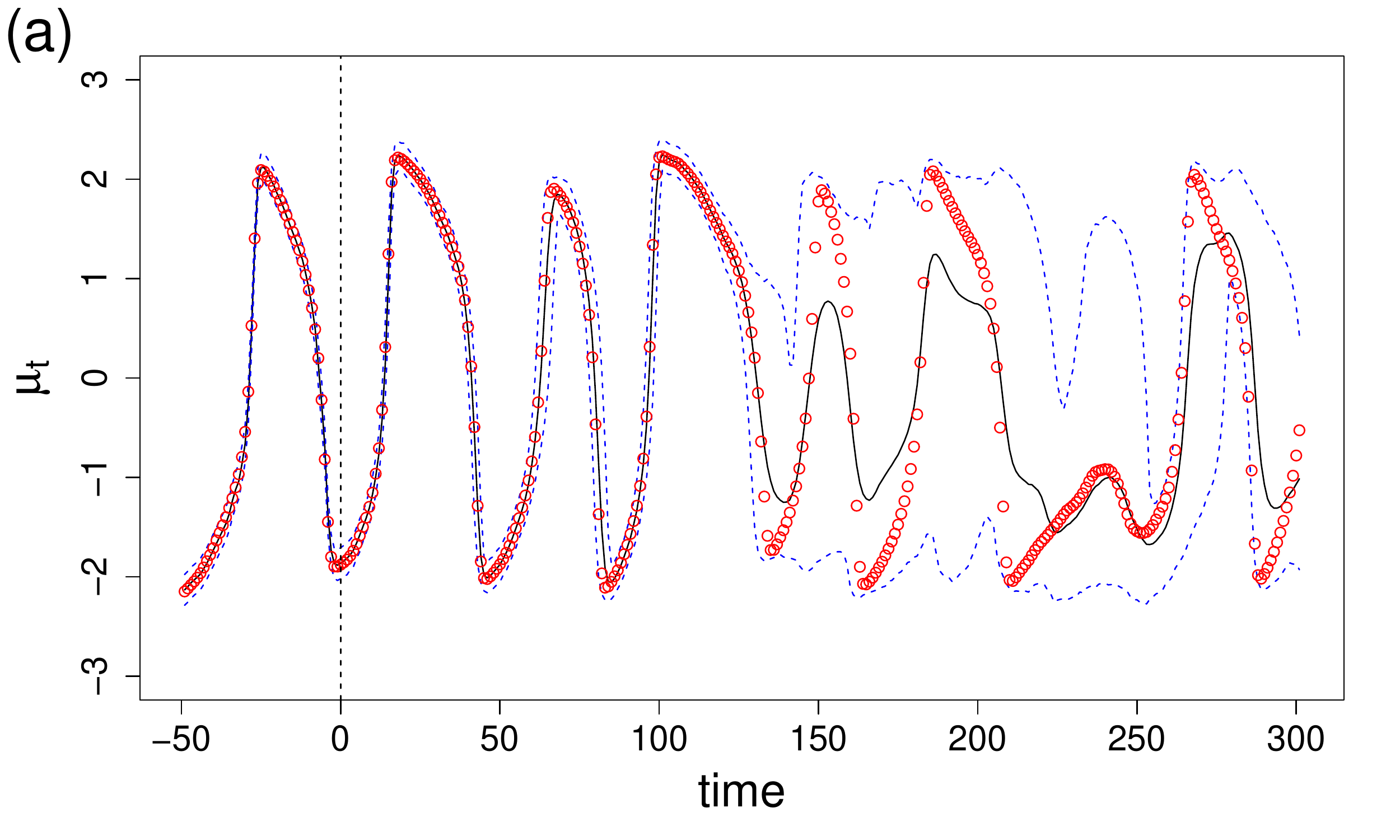}
  \includegraphics[width=0.48\textwidth]{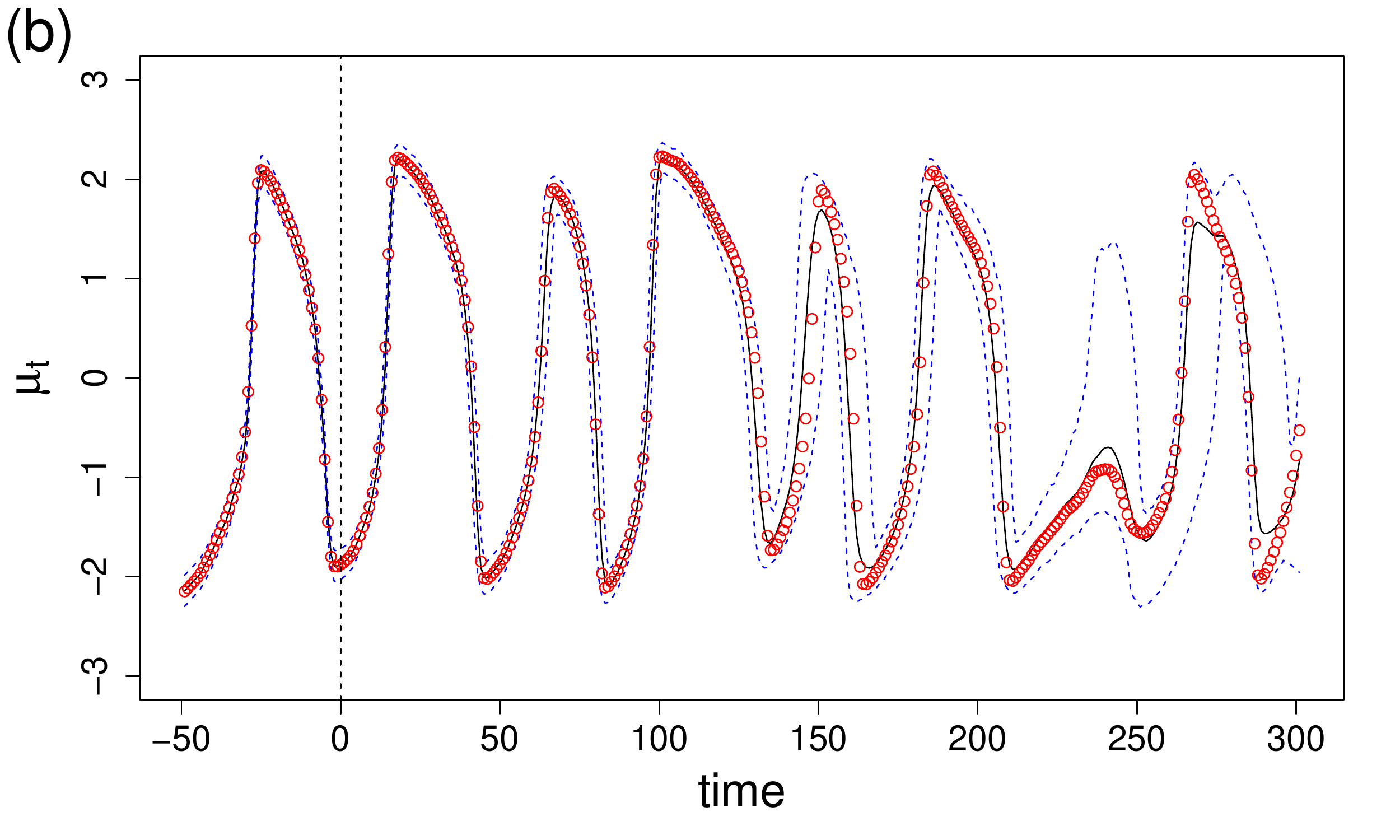}\\
  \includegraphics[width=0.48\textwidth]{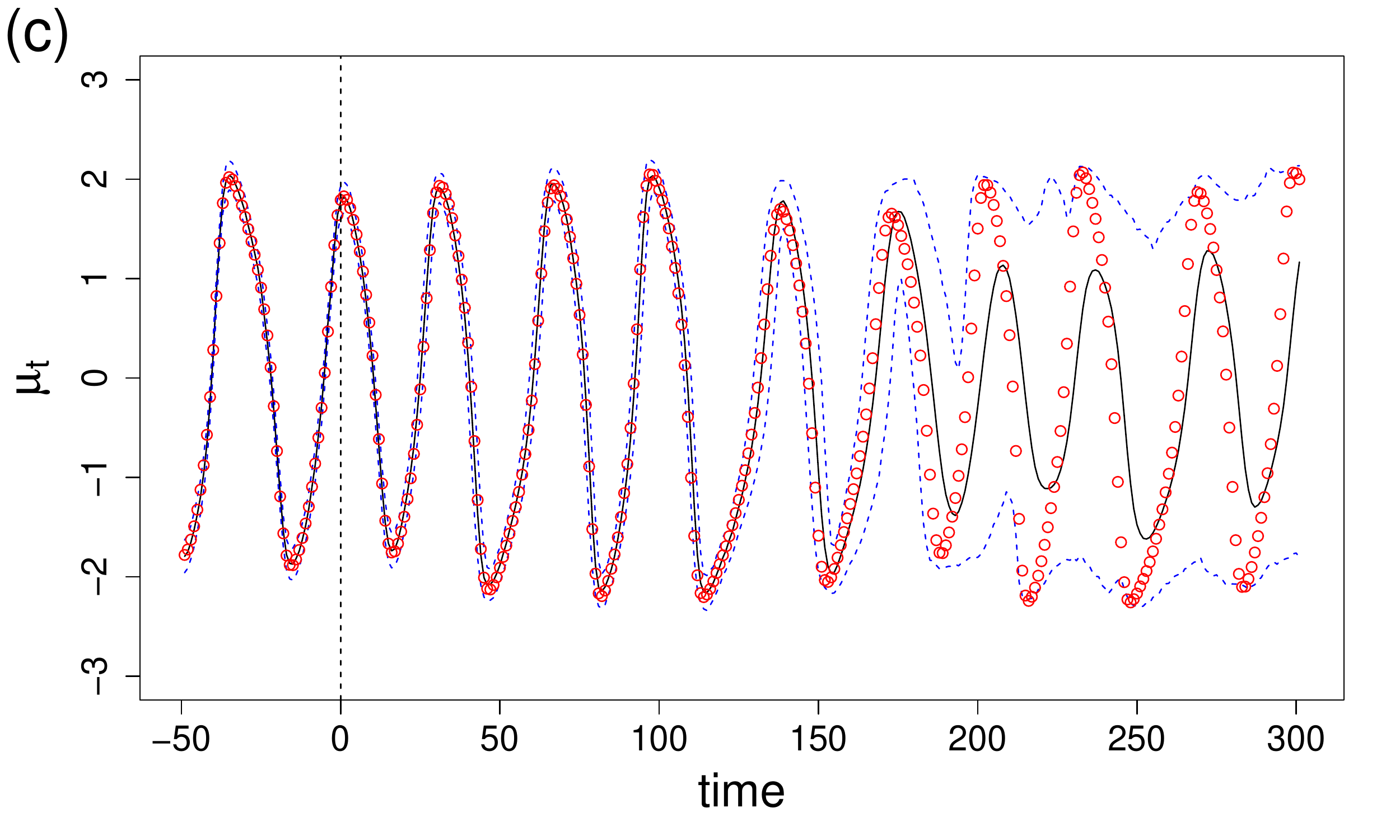}
  \includegraphics[width=0.48\textwidth]{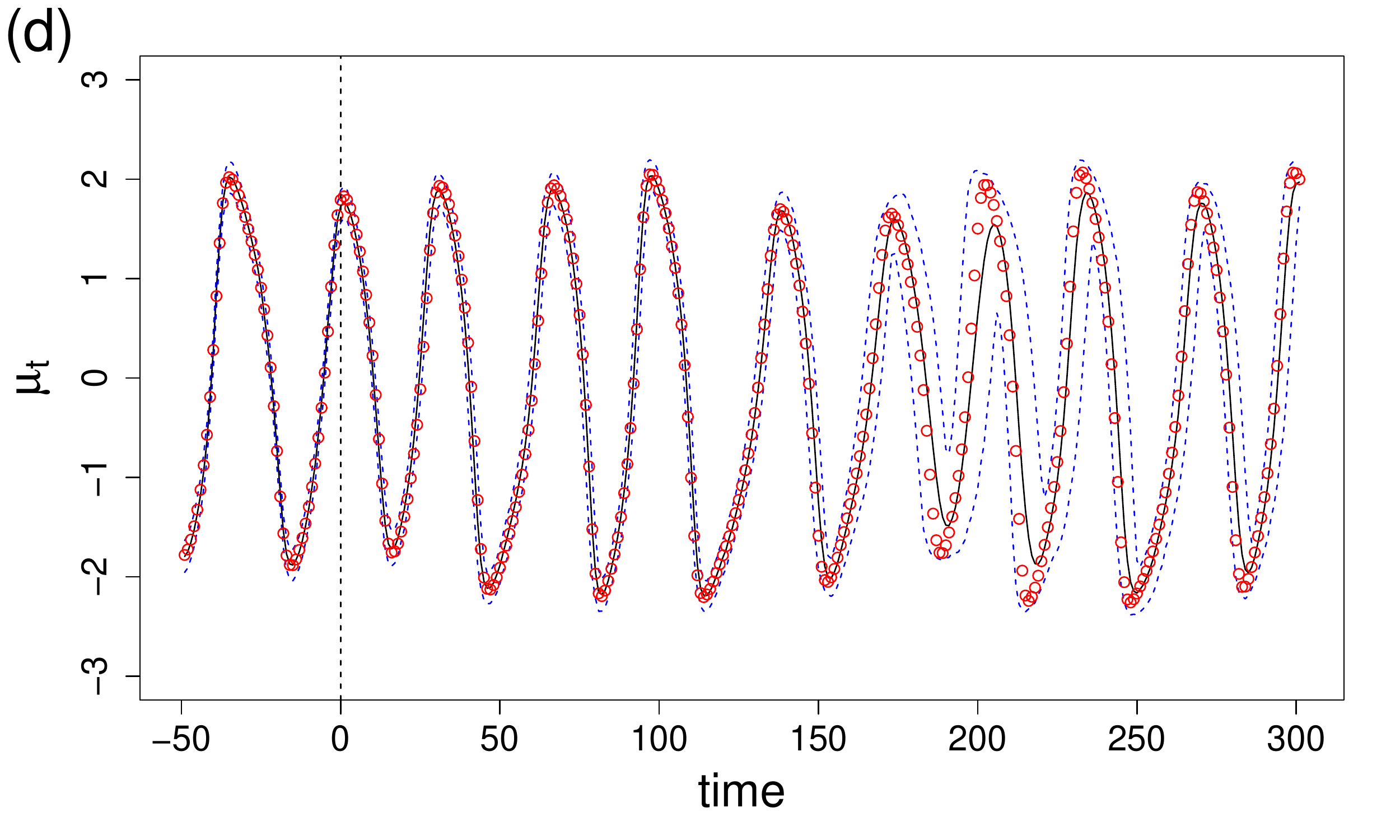}
  \caption{Multiple-step forecast of the Van der Pol oscillator for two sets of the parameters, $(\alpha,\gamma)$; (a,b) (0.668,2.271), and (c,d) (0.378,1.035). (a,c) are computed from RNN, and (b,d) are from VI-RNN with $\lambda = 1$. The solid and dashed lines, respectively, denote the expectation and 95\% prediction interval, and the circles (${\color{red}\circ}$) are the ground truth. The time is normalized by the sampling interval, $\delta t$. The vertical dashed line denotes the starting point of the multiple-step forecast.} \label{fig:VDP_multi_step}
\end{figure}

\begin{table}
\center{
\caption{ Absolute maximum cross-correlation of $\bm{z}$ in terms of $\lambda$}\label{tbl:max_cor_vdl}
\begin{tabular}{c|ccc}
\hline \hline
$\lambda$ & 0.1 & 1.0 & 10.0 \\
\hline
 & 0.89 & 0.04 & 0.01\\
 \hline \hline
\end{tabular}
}
\end{table}

Table \ref{tbl:max_cor_vdl} shows the maximum of the absolute value of cross-correlation coefficients of $\bm{z}$, which is computed by
\[
\| Cor(\bm{z},\bm{z})-\bm{I} \|_{max}.
\]
It is shown that at $\lambda=  0.1$, the latent variable, $\bm{z}$, has a high correlation. At $\lambda = 1$, the latent variable becomes almost linearly independent. Hence, we mainly focus on the model behavior for $\lambda = 1$ in this section. 

\begin{table}
\center{
\caption{Empirical coverage probability, $\text{CP}_p$} \label{tbl:VDP_cover}
\begin{tabular}{c|ccccc}
\hline \hline
 $p$ & 0.6 & 0.7 & 0.8 & 0.9 & 0.95 \\
 \hline
 RNN & 0.64 & 0.74 & 0.84 & 0.93 & 0.97 \\
 VI-RNN & 0.65 & 0.75 & 0.84 & 0.93 & 0.96 \\
 \hline \hline 
 \end{tabular}
}
\end{table}

Table \ref{tbl:VDP_cover} shows the empirical coverage probabilities of RNN and VI-RNN. Similar to the Mackey-Glass time series, the difference in $\text{CP}_p$ between RNN and VI-RNN is not so noticeable. However, it is shown that $\text{CP}_p$ is about $3 \sim 5\%$ larger than the coverage level of the prediction interval, indicating that both of the models are under-confident, \emph{i.e.}, the uncertainty predicted by the model is higher than the actual. The time series data of the Van der Pol Oscillator is generated by a downsampling with a sampling rate of 1 in 20. When the model is trained with the downsampled data, it learns to make a prediction without knowing the stochastic forcing between the sampling points. It seems that this additional uncertainty associated with the ``filtered'' forcing makes the model under-confident.

In figure \ref{fig:VDP_multi_step}, the multiple-step forecasts of RNN and VI-RNN are compared. VI-RNN is trained with $\lambda=1$. In the multiple-step forecast, the exogenous forcing, $\bm{u}$, is given for the entire forecast horizon and the observation is given only up to $t= 0$, \emph{i.e.}, the dataset consists of $(\bm{Y}_{-200:0},\bm{U}_{-200:300})$.  Similar to the Mackey-Glass time series, VI-RNN is able to make a much more accurate long-term forecast compared to RNN.

\begin{figure}
  \centering
  \includegraphics[width=0.48\textwidth]{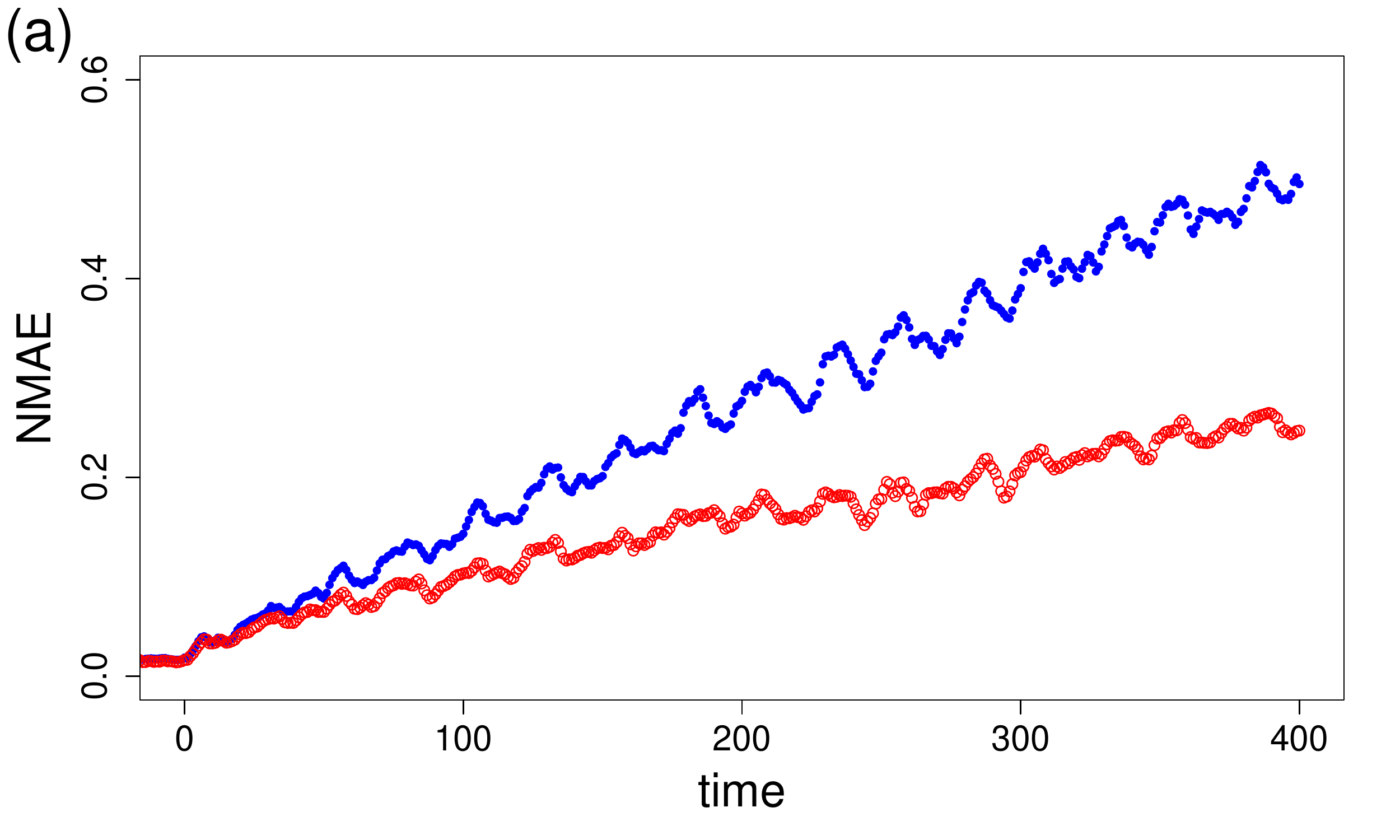}
  \includegraphics[width=0.48\textwidth]{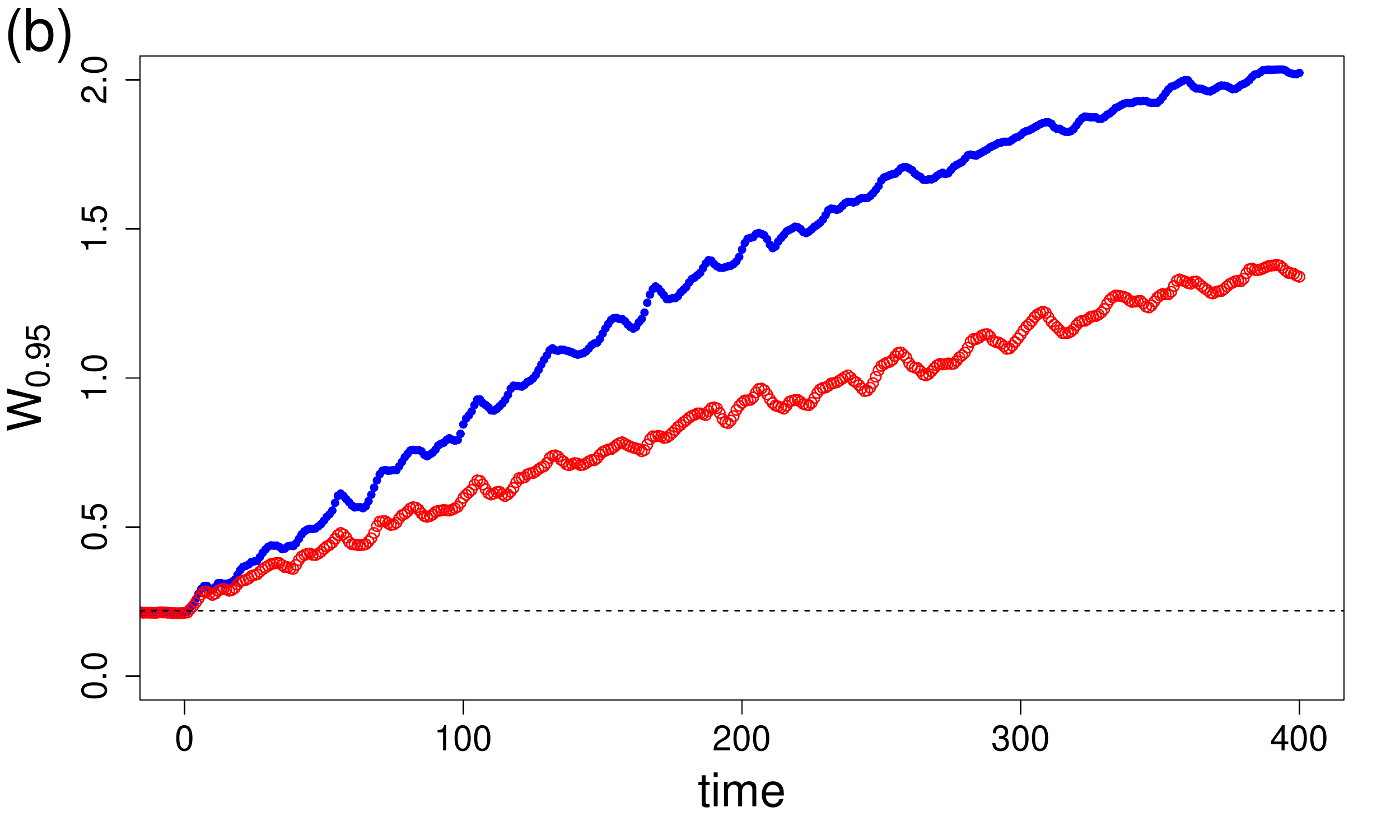}
  \caption{Temporal growth of (a) normalized mean absolute error and (b) width of 95\% prediction interval for RNN (${\color{blue}\bullet}$) and VI-RNN (${\color{red}\circ}$) for the multiple-step forecast. The dashed line in (b) denotes the noise level. The time is normalized by the sampling interval, $\delta t$. } \label{fig:VDP_multistep_error}
\end{figure}

The multiple-step forecast accuracies and the uncertainty ranges are shown in figure \ref{fig:VDP_multistep_error}. As expected, both NMAE and $W_{0.95}$ of VI-RNN show much slower growths in time compared to those of RNN. In a short-range forecast, $t < 50 \delta t$, the difference between VI-RNN and RNN is not noticeable. However, as  the forecasting horizon, $t$, increases, the advantage of VI-RNN becomes more pronounced. 

In \cite{Yeo19}, it is shown that a RNN, trained on one long trajectory, can make an accurate long-term forecast of VDP, where the prediction uncertainty remains stable even for 3000-step ahead forecasts. However, in this problem, the RNN is trained with an ensemble of trajectories with different parameters, which makes it difficult for the RNN to identify the dynamics of the trajectory. For $t=-200 \delta t \sim 0$, the observations, $y_t$, are provided to constrain the behavior of the RNN. In the short-term forecast, $0<t<100 \delta t$, the RNN simulates the dynamics close to what it learned from $\bm{Y}_{-200:0}$ due to an inertia. Eventually, the internal states of the RNN start to diverge, resulting in the increase in the prediction uncertainty. On the other hand, VI-RNN constrains the behavior of its decoder RNN around the one identified by $q(\bm{z}|\bm{Y}_{-200:0})$, which makes it possible to make a stable simulation for a much longer time horizon.

\begin{figure}
  \centering
  \includegraphics[width=0.48\textwidth]{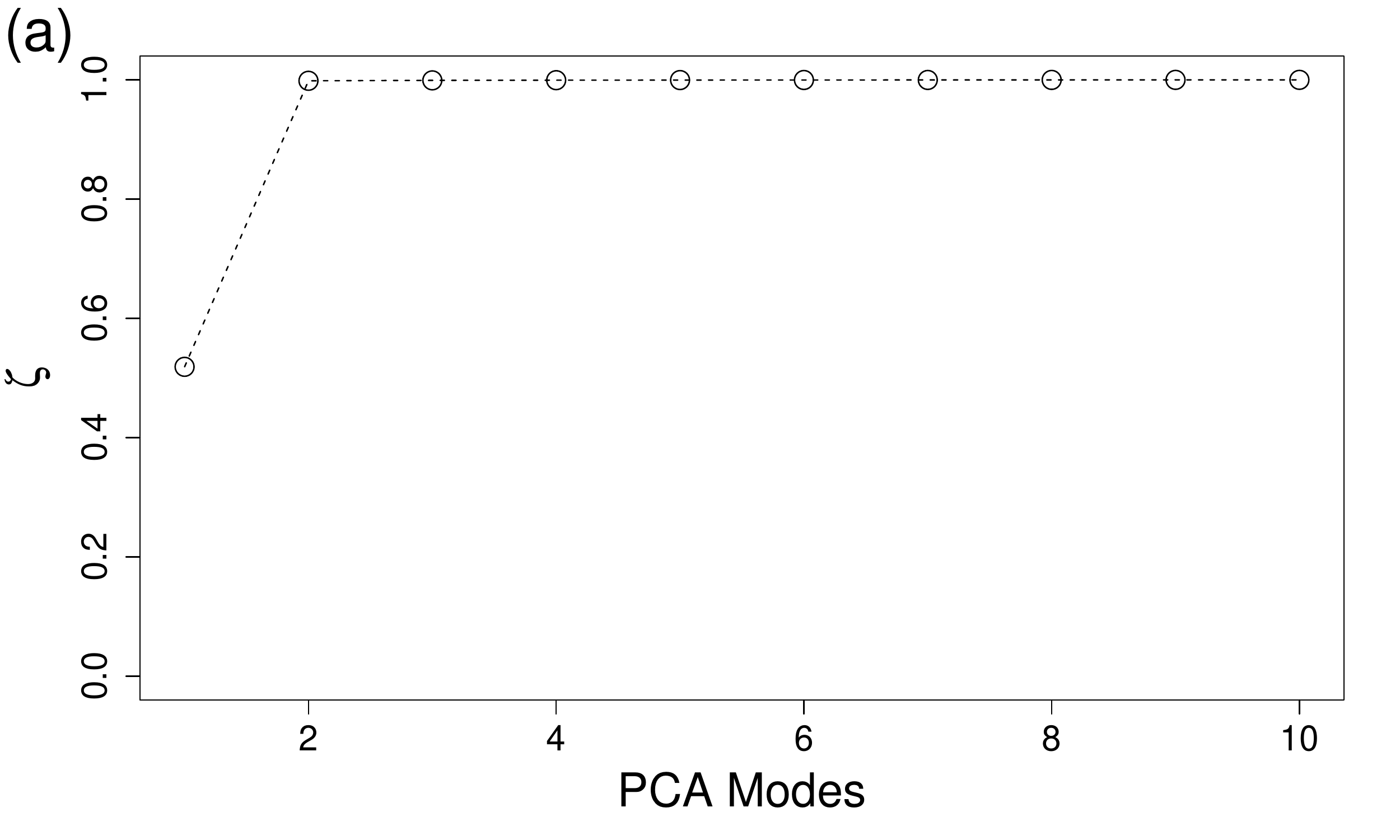}
  \includegraphics[width=0.48\textwidth]{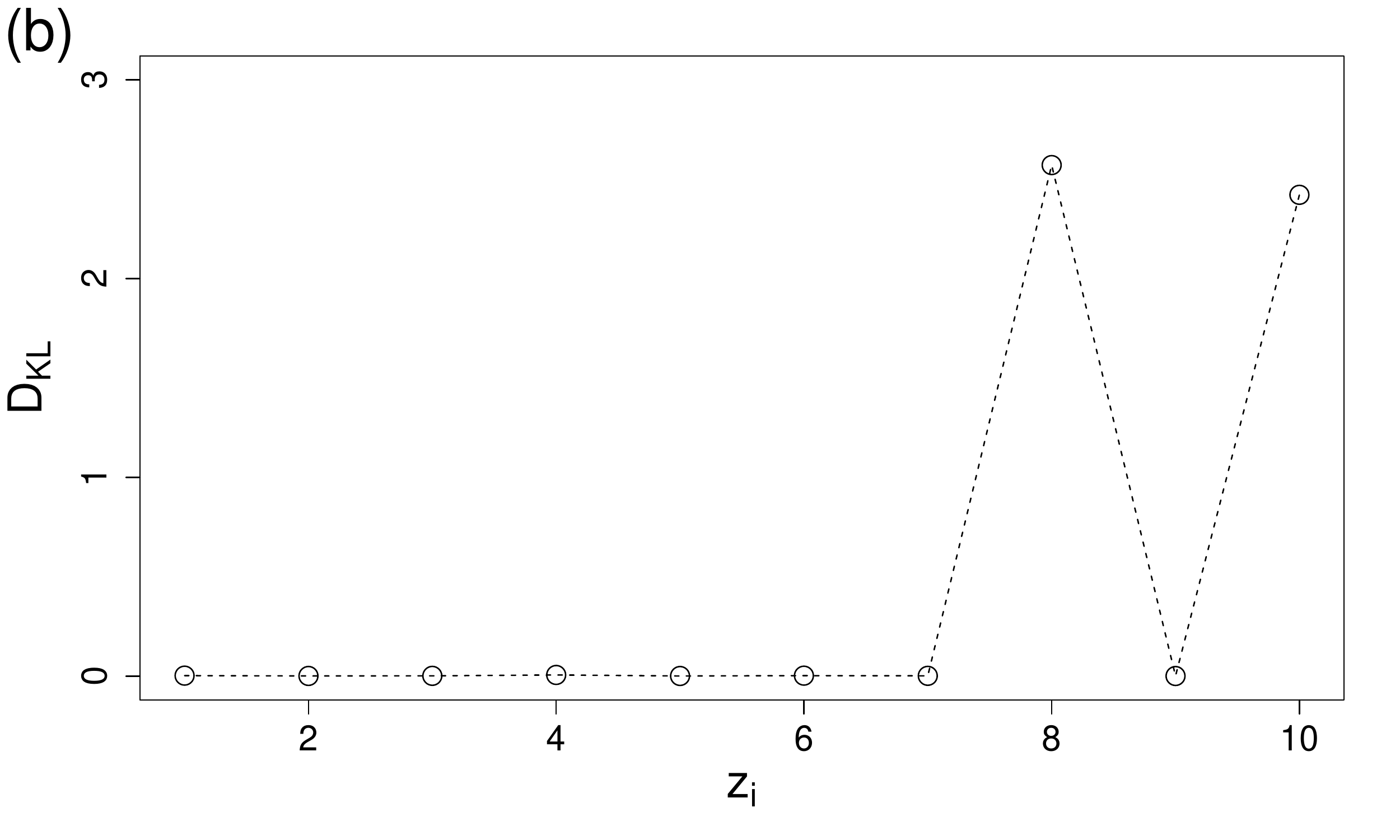}
  \caption{(a) Cumulative eigenvalues of PCA modes of $\bm{m}_q$ and (b) element-wise Kullback-Leibler divergence. } \label{fig:VDP_PCA_KL}
\end{figure}

Figure \ref{fig:VDP_PCA_KL} shows inferences from $q(\bm{z}|\bm{Y})$ for the training dataset, $\mathcal{D}_T$. It is shown in figure \ref{fig:VDP_PCA_KL} (a) that the cumulative eigenvalues becomes larger than 0.99 from the second PCA modes, $\zeta_2 = 0.998$, indicating the variations in $\bm{z}$ for $\mathcal{D}_T$ can be almost totally explained with only two variables. In fact, figure \ref{fig:VDP_PCA_KL} (b) shows that only two dimensions ($z_8$, $z_{10}$) are significantly different from the prior distribution.

\begin{figure}
  \centering
  \includegraphics[width=0.49\textwidth]{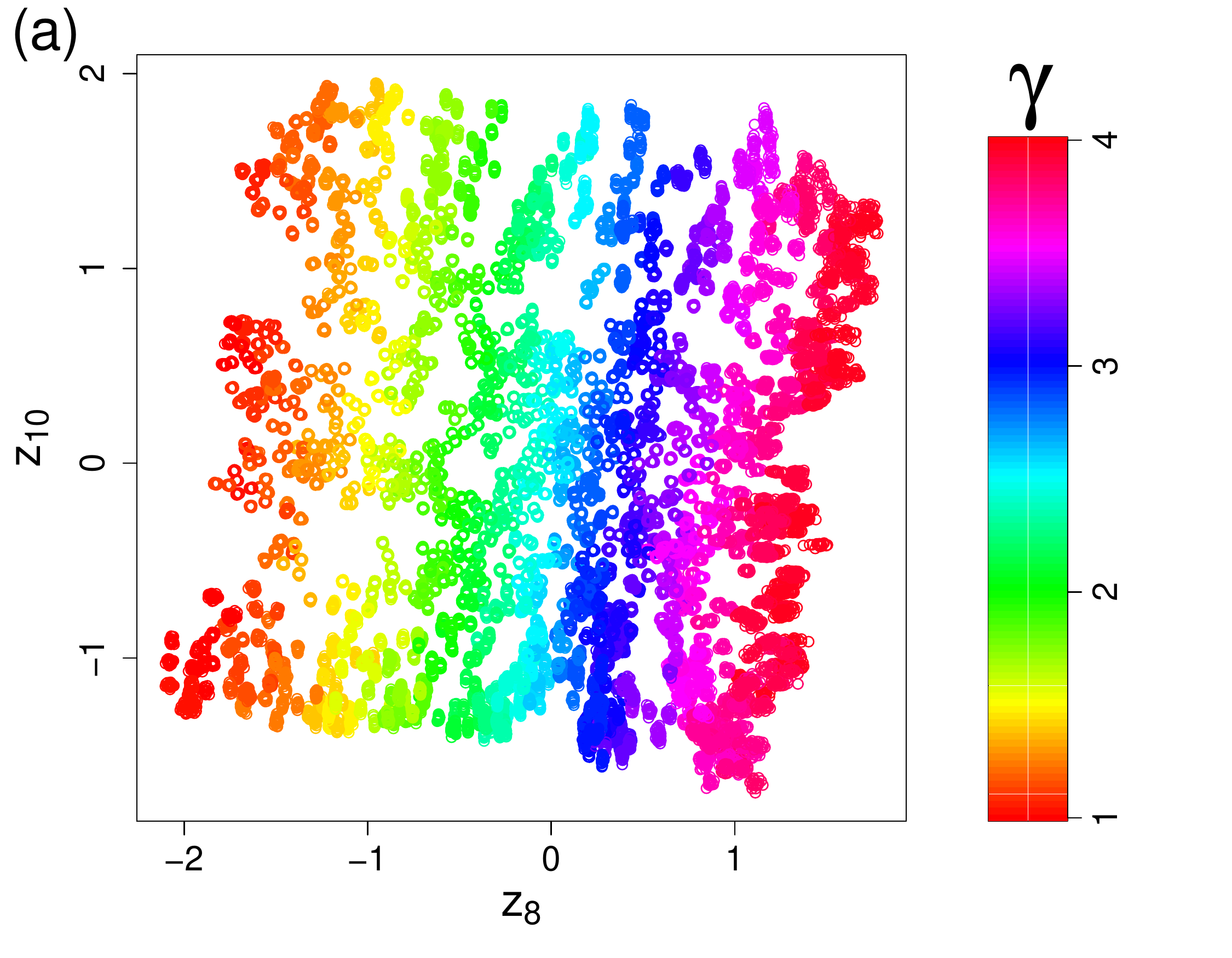}
  \includegraphics[width=0.49\textwidth]{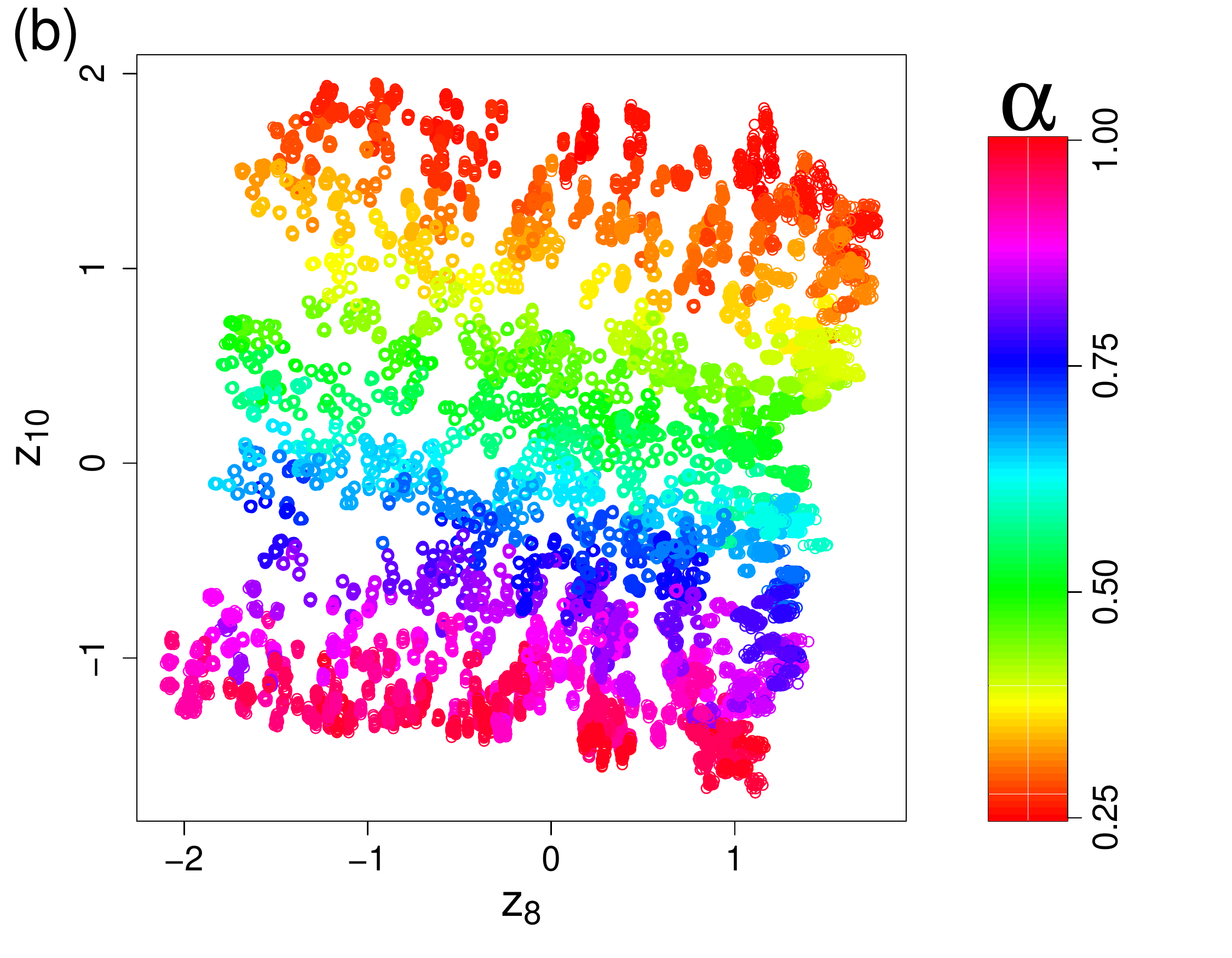}
  \caption{The posterior distribution projected onto the two non-trivial dimensions; $z_8-z_{10}$ plane. The symbols denote the samples from $q(\bm{z}|\bm{Y}^k)$ for all $\bm{Y}^k \in \mathcal{D}_T$. The color coding denotes (a) $\gamma$ and (b) $\alpha$.} \label{fig:VDP_latent_param}
\end{figure}

The two meaningful dimensions of $\bm{z}$ color-coded with the random parameters are shown in figure \ref{fig:VDP_latent_param}. Similar to the Mackey-Glass time series, the random parameters show almost linear variations in the $z_8-z_{10}$ plane. 

\section{Summary}  \label{sec:summary}
A deep learning approach is proposed to learn the dynamics from a time series dataset of a dynamical system with unknown parameters without any prior knowledge on the dynamical system or the dimensionality of the parameters. The learning problem is formulated as a variational inference problem, in which the effects of the parameters are approximated by introducing a latent variable, $\bm{z}$. In the variational inference framework, two artificial neural networks are jointly trained, one feedforward network to compute the approximate posterior distribution, $q(\bm{z}|\bm{Y})$, and a recurrent neural network to approximate the dynamics given the latent state, \emph{e.g.}, $\bm{y}_{t+1} = \bm{\Psi}^{VI}(\bm{y}_t,\bm{z})$. 
In this study, a two-level GRU is used to model the dynamics, because empirically it shows a better accuracy in learning complex stochastic processes, compared to the standard single-layer GRU. However, we also found that there is no noticeable difference in accuracy between single-layer and two-layer GRUs for the dynamics systems considered in this study.
To make the inference model flexible, the conditioning variable for $q(\bm{z}|\bm{Y})$ is obtained from an encoder recurrent neural network, which is trained in a standard way without considering the randomness in the parameters. The loss function of the variational inference problem is derived from the evidence lower bound of the marginal data likelihood function. The loss function consists of two terms, the Kullback-Leibler divergence (KLD) between $q(\bm{z}|\bm{Y})$ and a prior distribution, $p(\bm{z})$, and a reconstruction error, which is similar to the negative log likelihood of a recurrent neural network. The KLD regularizes $q(\bm{z}|\bm{Y})$ to be similar to $p(\bm{z})$, while the other term pushes $q(\bm{z}|\bm{Y})$ away from $p(\bm{z})$ in a direction to minimize the negative log likelihood (NLL). It is discussed that, in a time series problem, the relative contribution between KLD and NLL to the loss function can be controlled by changing the sequence length of NLL, or, alternatively, by introducing a penalty parameter, $\lambda$.

The proposed variational-inference model, VI-RNN, is tested by using two nonlinear dynamical systems with random parameters. It is shown that, although VI-RNN does not show advantages over RNN in a simple one-step prediction task, VI-RNN outperforms RNN in the multiple-step predictions as the prediction horizon increases. In the beginning of the stochastic simulation, the RNN prediction shows a good agreement with the ground truth due to an inertia developed in the spin-up period. However, as the simulation time increases, the effects of the inertia vanishes and the RNN prediction quickly diverges from the ground truth, which is accompanied by a rapid increase in the prediction uncertainty. On the other hand, VI-RNN first makes an inference about the effects of the unknown parameters from the data in the spin-up period and use the outcome to drive the simulation. Hence, VI-RNN proves a much more stable and accurate simulation compared to RNN.

In the numerical experiments, we keep the structures of the recurrent neural network used for VI-RNN and the standard RNN approaches exactly the same, except for the input dimensions. The input variable of VI-RNN is appended with the latent state, $\bm{z}$. Hence, any improvement of VI-RNN over RNN can be attributed to how well the approximate posterior, $q(\bm{z}|\bm{Y})$, represents the effects of random parameters. It is shown that VI-RNN is capable of identifying the dimensionality of the random parameters and each meaningful dimension of $\bm{z}$ has a direct correspondence to one of the random parameters. The penalty parameter, $\lambda$, plays a role in the inference of VI-RNN. Increasing $\lambda$ makes $q(\bm{z}|\bm{Y})$ be similar to the prior distribution. Because an independent Gaussian distribution is used as the prior distribution, a large $\lambda$ makes $\bm{z}$ linearly independent. Since it was empirically shown that a large value of the penalty parameter, $\lambda \gg 1$, results in a better inference in \cite{Higgins17}, many of the follow-up studies adopt the suggestion. However, as shown in our derivation, the loss function does not have a reference scale, and it also depends on a few parameters, such as the dimensions of $\bm{y}_t$ and $\bm{z}$, and the length of the training sequence. Hence, it is difficult to assert what is the proper range of $\lambda$. From the numerical experiments, we suggest to use a small $\lambda$, which makes $\bm{z}$ almost linearly independent, \emph{e.g.},
\[
\lambda^* = \text{min}~ \lambda~~~\text{s.t.}~~~\|Cor(\bm{z},\bm{z})-\bm{I}\|_{max} < \delta,
\]
where the tolerance level, $\delta$, may be chosen to be 0.1 or smaller. In our numerical experiments, it is found that the prediction results are not highly sensitive to $\lambda$, as long as it remains small. For example, in our experiments on the Mackey-Glass time series, at $\lambda = 10$, the large contribution of the Kullback-Leibler divergence to ELBO makes $\bm{z}$ a random noise, and VI-RNN fails to capture the effects of the parameters. However, for $\lambda \le 1.0$, it is found that the prediction results are not very sensitive to $\lambda$. It is also shown that, by tuning $\lambda$ to make $\bm{z}$ barely linearly independent, VI-RNN can provide meaning information about the dynamical system, such as the dimension of the unknown parameters. Even though $\bm{z}$ becomes linearly dependent at a small $\lambda$, a principal component analysis of $\bm{z}$ may be used to identify an effective dimension of the problem.

In the present study, we focus only on a stationary process, in which the unknown parameters do not change over time. We expect the present method can be applied to identify a quasi-steady process, where the temporal changes of the parameters occur at a much longer timescale than the time required to identify the latent state (see figure \ref{fig:MG_time_latent}). It is a subject of a follow-up study to expand the proposed framework to make a robust inference on the dynamical systems with time-varying random parameters. 

\appendix

\section{Gated Recurrent Unit} \label{sec:gru}
The gated recurrent unit (GRU) is proposed by \cite{Cho14} as a simplified version of the Long Short-Term Memory Network (LSTM) \cite{Hochreiter97,Gers00}. Let $\bm{x}_t \in \mathbb{R}^{N_x}$ be an input vector to a GRU and $\bm{h}_t \in \mathbb{R}^{N_h}$ be the state vector of the GRU at time step $t$. In GRU, first auxiliary variables are computed as
\begin{align}
\bm{p}_t &= \sigma_g(\bm{W}_{px} \bm{x}_t + \bm{W}_{ph}\bm{h}_t+\bm{B}_p), \nonumber \\
\bm{q}_t &= \sigma_g(\bm{W}_{qx} \bm{x}_t + \bm{W}_{qh}\bm{h}_t+\bm{B}_q), \nonumber \\
\bm{r}_t &= \tanh(\bm{W}_{rx}\bm{x}_t + \bm{W}_{rh}(\bm{q}_t \odot \bm{h}_t) + \bm{B}_r). \nonumber \\ 
\end{align}
Here, $\sigma_g$ is the Sigmoid function, $\bm{W}_k$ for $k\in(px,ph,qx,qh,rx,rh)$ denotes the weight matrix and $\bm{B}_k$ for $k\in(p,q,r)$ is the bias vector, and $\odot$ is the element-wise multiplication operator, \emph{i.e.}, the Hadamard product. The dimensions of the auxiliary variables are the same with the state vector, \emph{i.e.}, $\dim(\bm{p}_t)=\dim(\bm{q}_t)=\dim(\bm{r}_t)=N_h$. Once the auxiliary variables are computed, the state vector is updated as,
\begin{equation}
\bm{h}_{t+1} = (1-\bm{p}_t) \odot \bm{h}_t + \bm{p}_t \odot \bm{r}_t.
\end{equation}
In the numerical experiments for the dynamical systems considered in this study, the difference between LSTM and GRU is not noticeable, while GRU has a smaller number of parameters. For more complex problems, however, LSTM may outperform our RNN model due to the large number of parameters \cite{Ferrandis19}. 

\section{Joint optimization of prior and posterior} \label{sec:prior_network}

The negative evidence lower bound (ELBO) of a VI problem is
\begin{equation} 
\mathcal{L} = D_{KL}(q(\bm{z}|\bm{X})||p(\bm{z})) - E_{\bm{z}\sim q(\bm{z}|\bm{X})}[ \log p(\bm{X}|\bm{z})],
\end{equation}
in which $\bm{X}$ is the data, $p(\bm{z})$ is a prior distribution, and  $q(\bm{z}|\bm{X})$ is the approximate posterior distribution. Let assume that the prior distribution, $p(\bm{z})$, is a Gaussian distribution with a diagonal covariance,
\[
p(\bm{z}) = \mathcal{N}(\bm{z};\bm{m}_0,diag(\bm{\gamma_0})),
\]
and similarly the approximate posterior distribution is 
\[
q(\bm{z}|\bm{X}) = \mathcal{N}(\bm{z};\bm{m}_q,diag(\bm{\gamma_q})).
\]
Here, $\gamma_0$ and $\gamma_q$ denote the variances of the respective Gaussian distributions.
We use an artificial neural network ($\bm{\eta}$) to approximate the posterior distribution, such that 
\begin{equation}
(\bm{m}_q,\bm{\gamma}_q) = \bm{\eta}(\bm{X}).
\end{equation}
Following the formulation of \cite{chung15}, let assume that the prior distribution is computed from another artificial neural network,
\begin{equation}
(\bm{m}_0,\bm{\gamma}_0) = \bm{\zeta}(\bm{X}).
\end{equation}
Without loss of generality, we assume $N_z = 1$. Then, the Kullback-Leibler divergence in (\ref{eqn:elbo}) can be written as
\begin{equation}
D_{KL}(q(z|\bm{X})||p(z)) = \frac{1}{2} \left\{  \frac{\gamma_q+(m_0-m_q)^2}{\gamma_0}  - \log\left( \frac{\gamma_q}{\gamma_0} \right) \right\} - \frac{1}{2}.
\end{equation}

At a (local) minimum of $\mathcal{L}$, the following condition holds
\begin{equation}
\bm{\nabla}_{\bm{\theta}} \mathcal{L} = \bm{0},
\end{equation}
in which $\bm{\theta} =(m_0,m_q,\gamma_0,\gamma_q)$. The gradients with respect to $\bm{\theta}$ are
\begin{align}
\frac{\partial \mathcal{L}}{\partial m_0} &= \frac{1}{\gamma_0}(m_0-m_q) = 0,\\
\frac{\partial \mathcal{L}}{\partial \gamma_0} &= -\frac{1}{2\gamma^2_0}\left[ \gamma_q +(m_0-m_q)^2\right] + \frac{1}{2\gamma_0}= 0,\\
\frac{\partial \mathcal{L}}{\partial m_q} &= \frac{1}{2\gamma_0}(m_q-m_0)-\beta = 0,\\
\frac{\partial \mathcal{L}}{\partial \gamma_q} &= \frac{1}{2\gamma_0}-\frac{1}{2\gamma_q} - \alpha = 0,
\end{align}
in which
\[
\alpha = \frac{\partial}{\partial \gamma_q} E_{z\sim q(z|\bm{X})}[ \log p(\bm{X}|z)],~~\text{and}~\beta = \frac{\partial}{\partial m_q} E_{z\sim q(z|\bm{X})}[ \log p(\bm{X}|z)].
\]
Hence, at a minimum, the solution should satisfy
\begin{eqnarray}
m_0 - m_q &= 0,\\
(\gamma_0 - \gamma_q) - (m_0 - m_q)^2 &= 0,\\
(m_q - m_0) - 2\beta \gamma_0 &= 0,\\
(\gamma_q-\gamma_0) - 2 \alpha \gamma_0\gamma_q &= 0,
\end{eqnarray}
which results in a trivial solution,
\[
m_0 = m_q,~\gamma_0 = \gamma_q,~\alpha = 0,~\beta = 0.
\]
Because both $(m_0,\gamma_0)$ and $(m_q,\gamma_q)$ are unknown variables, \emph{i.e.}, outputs of artificial neural networks, the probabilistic model is unidentifiable. 
It is also important to note that $\alpha = \beta = 0$ implies that the reconstruction error becomes independent from the distribution of $\bm{z}$. In other words, the latent variable does not play a role in the inference.

\section{Training of VI-RNN} \label{sec:training}

\begin{figure}
  \centering
  \includegraphics[width=0.8\textwidth]{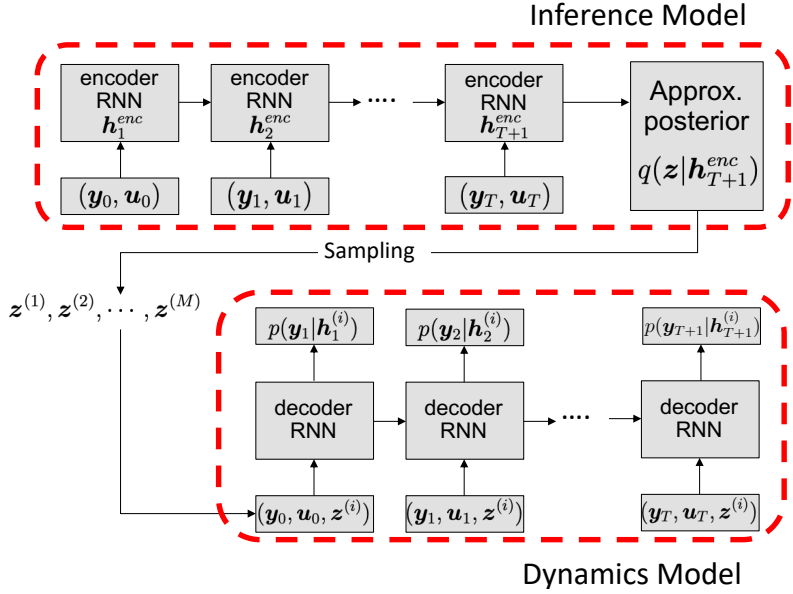}
  \caption{Sketch of VI-RNN} \label{fig:sketch}
\end{figure}

A sketch of VI-RNN is shown in figure \ref{fig:sketch}. The encoder RNN is pre-trained. Here, we consider the training of the posterior network, $\widehat{\bm{\eta}}$, and the decoder RNN, $\bm{\Psi}^{VI}$. Let $\bm{\theta}_{\bm{\eta}}$ and $\bm{\theta}_{\bm{\Psi}}$ be the parameters of $\widehat{\bm{\eta}}$ and $\bm{\Psi}^{VI}$, respectively. In the model training, we aim to find the parameters, which minimize the loss function by solving the following optimization problem,
\begin{equation} \label{eqn:optimization_loss}
\min_{\bm{\theta}_{\bm{\eta}},\bm{\theta}_{\bm{\Psi}}} \sum_{k=1}^K \lambda \mathcal{L}^k_q + \mathcal{L}^k_y.
\end{equation}
It is typical to use a variant of Stochastic Gradient Descent (SGD) methods to solve (\ref{eqn:optimization_loss}). Using a SGD, the parameters are updated as
\begin{equation}
  \begin{bmatrix}
    \bm{\theta}_{\bm{\eta}}^{n+1}\\
    \bm{\theta}_{\bm{\Psi}}^{n+1}
  \end{bmatrix}
  =     
  \begin{bmatrix}
    \bm{\theta}_{\bm{\eta}}^{n}\\
    \bm{\theta}_{\bm{\Psi}}^{n}
  \end{bmatrix}
  -
  \xi
  \begin{bmatrix}
   \bm{\nabla}_{\bm{\theta}^n_{\bm{\eta}}} \left( \lambda \mathcal{L}^{s_n}_q + \mathcal{L}^{s_n}_y \right) \\
   \bm{\nabla}_{\bm{\theta}^n_{\bm{\Psi}}} \mathcal{L}^{s_n}_y
  \end{bmatrix}
  ,~\text{for}~ n = 1,\cdots,N_{max},
\end{equation}
in which $n$ is the iteration count, $N_{max}$ is the maximum number of iterations, $\xi$ is a learning rate, and $s_n \sim \mathcal{U}(\mathcal{D})$ is a uniform random variable in $\mathcal{D}$ to sample a trajectory at each iteration. For now, we consider $N_z = d = 1$ for simplicity. Note that, due to the diagonal covariance assumptions for the inference and generative models, it is trivial to extend the results for a higher dimensional problem.

Computing the gradients of $\mathcal{L}_q$ is straightforward;
\begin{equation} \label{eqn:SGD}
\frac{\partial \mathcal{L}_q}{\partial \bm{\theta}_\eta} = \frac{\partial \mathcal{L}_q}{\partial m_q}\frac{\partial m_q}{\partial \bm{\theta}_\eta} + \frac{\partial \mathcal{L}_q}{\partial \log \sigma_q}\frac{\partial \log \sigma_q}{\partial \bm{\theta}_\eta}.
\end{equation}
Here,
\[
\frac{\partial \mathcal{L}_q}{\partial m_q} = \frac{m_q}{\sigma^2_z}, ~~\text{and}~~
\frac{\partial \mathcal{L}_q}{\partial \log \sigma_q} = \frac{\sigma^2_q}{\sigma^2_z}-1.
\]
The gradients of $m_q$ and $\sigma_q$ with respect to $\bm{\theta}_{\bm{\eta}}$ can be easily computed by using a back propagation method on the computation of $\bm{\eta}$.

The loss function, $\mathcal{L}_y$, involves an expectation over $\bm{z}$, which is approximated by a Monte Carlo sampling
\begin{equation}\label{eqn:Ly-1}
  \mathcal{L}_y = -E_{z \sim q(z|\bm{Y}_{0:T})} \left[ \sum_{t=1}^T \log p(y_t|\bm{h}_t) \right] \simeq -\frac{1}{M} \sum_{m=1}^M  \left\{ \sum_{t=1}^T \log p(y_t|\bm{h}_t) \right\} \Big|_{z^m}.
\end{equation}
Here, $z^m$ is sampled from the posterior distribution, $q(z|\bm{Y}_{0:T})$, as
\begin{equation} \label{eqn:sample}
z^m = \mu_q + \sigma_q \epsilon^m,~\text{for}~l = 1,\cdots,M,
\end{equation}
in which $\epsilon^m \sim \mathcal{N}(0,1)$. Let define
\[
l^m_t =  - \log p(y_t|\bm{h}_t) \Big|_{z^m} = \left( \frac{1}{2}\frac{(y_t - \mu_t)^2}{\sigma^2_t} + \log \sigma_t \right) \Big|_{z^m}.
\]
Note that the constant term is neglected. 
The SGD update requires to compute
\begin{equation}
\frac{\partial l^m_t}{\partial \bm{\theta}_{\eta}} = \frac{\partial l^m_t}{\partial z}\left( \frac{\partial z}{\partial m_q}\frac{\partial m_q}{\partial \bm{\theta}_{\eta}} + \frac{\partial z}{\partial \log \sigma_q}\frac{\partial \log \sigma_q}{\partial \bm{\theta}_{\eta}} \right)\Big|_{z^m}.
\end{equation}
The gradient of $l^m_t$ with respect to the latent variable is
\begin{equation} \label{eqn:grad_lt}
\frac{\partial l^m_t}{\partial z} = \frac{\partial l^m_t}{\partial \mu_t}\frac{\partial \mu_t}{\partial z}+ \frac{\partial l^m_t}{\partial \log \sigma_t}\frac{\partial \log \sigma_t}{\partial z},
\end{equation}
where
\[
 \frac{\partial l^m_t}{\partial \mu_t} = \frac{\mu_t-y_t}{\sigma^2_t} \Big|_{z^m}~~\text{and}~~
 \frac{\partial l^m_t}{\partial \log \sigma_t} = 1 - \frac{(\mu_t - y_t)^2}{\sigma_t^2} \Big|_{z^m}.
\]
A back propagation algorithm can be used to evaluate the gradients, $\frac{\partial \mu_t}{\partial z}|_{z^m}$ and $\frac{\partial \log \sigma_t}{\partial z}|_{z^m}$, in (\ref{eqn:grad_lt}). From  the random sampling procedure (\ref{eqn:sample}), it is clear that
\begin{equation}
\frac{\partial z}{\partial m_q} \Big|_{z^m} = 1,~~\text{and}~~~\frac{\partial z}{\partial \log \sigma_q}\Big|_{z^m} = \sigma_q \epsilon^m.
\end{equation}
Now, the gradient of $\mathcal{L}_y$ with respect to $\bm{\theta}_{\bm{\eta}}$ can be computed as
\begin{equation}
\bm{\nabla}_{\bm{\theta}_{\bm{\eta}}} \mathcal{L}_y =  \left[ \frac{1}{M} \sum_{m=1}^M \sum_{t=1}^T \frac{\partial l_t^m}{ \partial z} \right] \frac{\partial m_q}{\partial \bm{\theta}_\eta} + \left[  \frac{\sigma_q}{M} \sum_{m=1}^M \left( \epsilon^m \sum_{t=1}^T \frac{\partial l_t^m}{ \partial z} \right) \right]\frac{\partial \log \sigma_q}{\partial \bm{\theta}_\eta}.
\end{equation}
Here, the gradients of $m_q$ and $\sigma_q$ with respect to $\bm{\theta}_{\bm{\eta}}$ can be easily computed by using a back propagation method on the computation of $\bm{\eta}$.

In summary, the gradients of the loss function are,
\begin{align}
\frac{\partial \mathcal{L}}{\partial \bm{\theta}_{\bm{\eta}}} &=  \left[ \lambda  \frac{m_q}{\sigma^2_z} + \frac{1}{M} \sum_{m=1}^M \sum_{t=1}^T \frac{\partial l_t^m}{ \partial z} \right] \frac{\partial m_q}{\partial \bm{\theta}_\eta} \nonumber \\
&+ \left[  \lambda \left( \frac{\sigma^2_q}{\sigma^2_z}-1\right)+ \frac{\sigma_q}{M} \sum_{m=1}^M \left( \epsilon^m \sum_{t=1}^T \frac{\partial l_t^m}{ \partial z} \right) \right]\frac{\partial \log \sigma_q}{\partial \bm{\theta}_\eta}, \label{eqn:grad-eta}\\
\frac{\partial \mathcal{L}}{\partial \bm{\theta}_{\bm{\Psi}}} &= \frac{1}{M}\sum_{m=1}^M \left[ \sum_{t=1}^T \left( \frac{\mu_t - y_t}{\sigma^2_t} \right) \frac{\partial \mu_t}{\partial \bm{\theta}_{\bm{\Psi}}} + \left( 1-\frac{(\mu_t-y_t)^2}{\sigma^2_t}\right)  \frac{\partial \log \sigma_t}{\partial \bm{\theta}_{\bm{\Psi}}} \right]_{z^m}. \label{eqn:grad-psi}
\end{align}
For each SGD iteration, first $(m_q, \sigma_q)$ are computed from the training trajectory $(\bm{Y}_{0:T})$, and $M$ samples are drawn from $q(z|\bm{Y}_{0:T})$ as shown in (\ref{eqn:sample}). Then, the RNN simulations are performed forward in time for the samples,
\begin{equation}
(\mu_t,\log \sigma_t)|_{z_m} = \bm{\Psi}^{VI}(y_{t-1},\bm{h}_{t-1},z^m),~\text{for}~t = 1,\cdots,T~\&~m=1,\cdots,M.
\end{equation}
Finally, the gradients of the loss function are computed by (\ref{eqn:grad-eta}--\ref{eqn:grad-psi}) through a back-propagation algorithm and the parameters of the artificial neural networks are updated by (\ref{eqn:SGD}).

\section{Monte Carlo simulation of VI-RNN} \label{sec:MC_algorithm}

\begin{algorithm}[!tbh]
	\caption{Monte Carlo simulation of VI-RNN}
	\label{alg:MC}
	\textbf{Input}: Data $(\bm{Y}_{-t:0},\bm{U}_{-t:0})$, MC sample size ($N_s$), forecast horizon ($T_f$)\\
	\textbf{Output}: $N_s$ samples from $p(\bm{y}_i|\bm{Y}_{-t:0},\bm{U}_{-t:i-1})$ for $i = 1,\cdots,T_f$.
	
	\begin{algorithmic}
		\STATE 1. Compute the approximate posterior distribution:
		\STATE \hskip1.5em Sequential update of $\bm{\Psi}^{enc}$ for $(\bm{Y}_{-t:0},\bm{U}_{-t:0})$. \\
		{}\hspace{2.5em}$\displaystyle
		\bm{h}^{enc}_{i} = \bm{\Psi}^{enc}_h(\bm{y}_{i-1},\bm{u}_{i-1},\bm{h}^{enc}_{i-1})~~\text{for}~~i = -t+1,\cdots,1. 
		$
		\STATE \hskip1.5em Compute the approximate posterior distribution\\
		{}\hspace{2.5em}$\displaystyle
		(\bm{m}_q,\bm{\sigma}_q) = \widehat{\bm{\eta}}(\bm{h}^{enc}_0)
		$
		\STATE \hskip1.5em Draw $N_s$ samples from $q(\bm{z}|\bm{Y}_{-t:0},\bm{U}_{-t:0})$\\
		{}\hspace{2.5em}$\displaystyle
		\bm{z}^{(j)} = \bm{m}_q + \bm{\sigma}_q \odot \bm{\epsilon}~~\text{for}~j=1,\cdots,N_s,
		$
		where $\bm{\epsilon} \sim \mathcal{N}(0,\bm{I})$.
		\STATE 2. Simulate the decoder RNN with the input data  $(\bm{Y}_{-t:0},\bm{U}_{-t:0})$:\\
		{}\hspace{2.5em}$\displaystyle
		\bm{h}^{(j)}_{i} = \bm{\Psi}^{VI}_h(\bm{y}_{i-1},\bm{u}_{i-1},\bm{h}^{(j)}_{i-1},\bm{z}^{(j)}). 
		$\\
		{}\hspace{2.5em} for $i = -t+1,\cdots,1$ and $j = 1,\cdots,N_s$.
		
		\STATE 3. Monte Carlo simulation with the decoder RNN:\\		
		\FOR{$i=1,T_s$}	
		\FOR{$j=1,N_s$}
		\STATE Distribution of $\bm{y}_i$: \hspace{0.2em}$\displaystyle
		(\bm{\mu}^{(j)}_i,\bm{\sigma}^{(j)}_i) = \bm{\Psi}^{VI}_y(\bm{h}^{(j)}_i)
		$	
		\STATE MC Sampling of $\bm{y}_i$: $\displaystyle
		\bm{y}_i^{(j)} = \bm{\mu}^{(j)}_i + \bm{\sigma}^{(j)}_i \odot \bm{\epsilon}
		$
		\STATE Update RNN hidden state: $\displaystyle
		\bm{h}^{(j)}_{i+1} = \bm{\Psi}^{VI}_h(\bm{y}^{(j)}_i,\bm{h}^{(j)}_i,\bm{z}^{(j)},\bm{u}_i). 
		$
		
		\ENDFOR
		\ENDFOR
	\end{algorithmic}
\end{algorithm}

%\begin{thebibliography}{1}
\bibliographystyle{siam}
\bibliography{ref_rnn}
%\end{thebibliography}
	
\end{document}